\documentclass[5p,twocolumn,preprint,times]{elsarticle}

\usepackage{amsmath,amssymb}
\usepackage{graphicx}
\usepackage{booktabs} 
\usepackage{caption}  
\usepackage{subcaption}

\usepackage{amssymb}
\usepackage{tabularx}
\usepackage{array} 
\usepackage{amsmath}
\usepackage{placeins}

\usepackage{lineno}

\journal{Ocean Engineering} 

\begin{document}

\begin{frontmatter}

\title{Predicting Wind Loads on Container Ships in Harbor Environments through Multi-Fidelity Modeling}

\author[vki]{Matilde Fiore\corref{cor1}}
\ead{matilde.fiore@vki.ac.be} 
\author[vki]{Andrea Bresciani}
\author[vki,atm,aerg]{Miguel A. Mendez}
\author[vki]{Jeroen van Beeck}

\affiliation[vki]{organization={Environmental and Applied Fluid Dynamics Department, von Karman Institute for Fluid Dynamics},
            addressline={Waterloosesteenweg 72}, 
            city={Sint-Genesius-Rode},
            postcode={B-1640}, 
            country={Belgium}}

\affiliation[atm]{organization={Aero-Thermo-Mechanics Laboratory, École Polytechnique de Bruxelles, Université Libre de Bruxelles},
            addressline={Av. Franklin Roosevelt 50}, 
            city={Brussels},
            postcode={1050}, 
            country={Belgium}} 

\affiliation[aerg]{organization={Aerospace Engineering Research Group, Universidad Carlos III de Madrid},
            addressline={Av. de la Universidad 30}, 
            city={Leganés},
            postcode={28911}, 
            country={Spain}}

\begin{abstract}
Modern container ships face higher wind loads due to increased windage areas, making accurate predictions of wind loads essential for mooring design. Existing empirical models, largely developed for container ships with smaller windage areas and simpler geometrical configurations than those of modern large-scale vessels, often lack accuracy and do not account for the influence of nearby structures. This study proposes a multi-fidelity surrogate modelling framework for the prediction of wind-load coefficients, combining empirical correlations with simplified and detailed CFD models for ships in open-sea and harbor environments. The approach relies on recursive co-kriging to consistently fuse information across fidelity levels, enabling accurate predictions at a reduced computational cost. A sensitivity analysis is used to identify the most influential geometric parameters, and the resulting reduced parameter space is explored through sequential sampling to efficiently construct the training database.

The surrogate models are validated over a wide range of loading configurations and for two distinct harbor environments. The results demonstrate that the multi-fidelity approach significantly improves prediction accuracy compared to single-fidelity models, while substantially reducing the reliance on high-fidelity simulations. In particular, the proposed framework captures the dependence of wind loads on key geometric parameters and consistently outperforms traditional empirical correlations, providing a robust and efficient tool for engineering applications.
\end{abstract}

\begin{highlights}
\item Multi-fidelity surrogate modeling of wind loads on container ships minimizes the number of expensive CFD simulations.
\item The surrogates explain the dependency of wind loads on the geometrical parameters of the ship.
\item The surrogate models integrate and outperform empirical correlations for wind loads developed in the past.
\end{highlights}


\end{frontmatter}


\section{Introduction}

Container ships account for approximately 90\% of global merchandise trade. In recent decades, vessel size has increased dramatically \cite{park2019tendency,gomez2015use,martin2015container}, with some ships exceeding 20,000 TEU, accompanied by a significant growth in windage area. As a result, modern vessels are increasingly susceptible to wind loads and extreme weather conditions. This issue is particularly critical in harbour environments, where ships are manoeuvred in confined spaces and are frequently subjected to mooring operations. Complex port layouts can induce channelling and sheltering effects, leading to increased uncertainty in wind force coefficients. Such uncertainties may have severe consequences, including loss of control during manoeuvres or failure of mooring lines, potentially resulting in high-risk accidents and significant economic losses. Several incidents reported in \cite{torre2021wind} highlight the role of unexpected wind loads during storms in collision events and unmooring operations. These challenges have motivated the development of automated mooring systems within so-called smart ports \cite{yan2023computational}, which require accurate and real-time prediction of wind loads.

Wind loads are traditionally obtained from wind tunnel experiments performed on scaled ship models, for which extensive databases are available in the literature \cite{andersen2013wind, aage1971wind, blendermann2011wind}. Based on these data, empirical correlations have been developed over the past decades \cite{blendermann1994parameter,isherwood1973wind,fujiwara2009experimental} to estimate wind loads for different ship geometries, loading configurations, and angles of attack. However, these correlations are typically derived from simplified representations of ship superstructures and are not fully representative of modern large-scale container vessels. Moreover, they generally neglect the influence of the surrounding harbour environment, which can significantly alter the effective wind loads.

Computational Fluid Dynamics (CFD) provides an alternative approach to predict wind loads at full scale by resolving the flow around detailed ship geometries and their environment. A substantial body of literature addresses Reynolds-averaged Navier–Stokes (RANS) simulations of ships in open sea \cite{saydam2018evaluation,wnkek2015cfd,janssen2017cfd}, with different container arrangements \cite{grlj2023effect}, in harbour configurations \cite{ricci2020cfd,torre2021wind}, and in side-by-side scenarios \cite{koop2020using}. Several studies \cite{ricci2020cfd,janssen2017cfd} have also investigated the impact of geometrical simplifications on predicted wind loads. While simplified geometries may introduce distortions in the aerodynamic response \cite{janssen2017cfd}, they significantly reduce computational cost and enable broader exploration of loading configurations and operating conditions. Nevertheless, high-fidelity CFD simulations remain computationally demanding, often requiring hours to days on standard workstations, which limits their applicability in real-time or large-scale parametric analyses.

To overcome these limitations, surrogate models can be employed to emulate wind load predictions based on available data. Recent studies have explored machine learning approaches, including Artificial Neural Networks (ANNs) \cite{degiuli2025machine,pache2022data,prpic2020hybrid,loft2025data} and Gaussian Process regression \cite{wang2026efficient}. These models are capable of handling complex, high-dimensional relationships between input parameters and wind load coefficients. For instance, \cite{degiuli2025machine} developed ANN-based models using container stack information and wind angle as inputs for multiple vessel types. Alternative parametrizations of loading configurations have been proposed using Fourier-based descriptors \cite{prpic2020hybrid,wang2026efficient}, enabling more compact representations of ship geometry. Other works \cite{pache2022data} have employed convolutional neural networks to predict pressure fields, combined with dimensionality reduction techniques such as Proper Orthogonal Decomposition and autoencoders.

Despite these advances, existing surrogate modelling approaches remain limited by the availability of high-fidelity data. Generating accurate CFD datasets is computationally expensive, restricting the coverage of the design space and the generalization capabilities of the resulting models. On the other hand, simplified models and empirical correlations, while less accurate, provide inexpensive but informative approximations of the underlying physics. This suggests that combining information from multiple fidelity levels could provide an effective strategy to balance accuracy and computational cost.

In this work, a multi-fidelity surrogate modelling framework is proposed for the prediction of wind loads on container ships. The approach combines empirical correlations, simplified CFD simulations, and high-fidelity CFD models within a unified Gaussian Process regression framework \cite{le2013multi}. By exploiting the correlation between data sources of different fidelity, the method enables accurate predictions while significantly reducing the number of expensive high-fidelity simulations required. A sensitivity analysis is used to identify the most influential geometric parameters, and a sequential sampling strategy is employed to efficiently explore the reduced parameter space. The predictive capability of the proposed approach is assessed across a wide range of loading configurations and extended to different harbour environments.

The remainder of the paper is organized as follows. The mathematical formulation of the multi-fidelity surrogate model for wind loads and its feature selection are described in section \ref{data-AHFM}. Section \ref{data_sources} describes the data-sources of various fidelities employed to train the surrogate, while section \ref{training} explains the training algorithm. The results of the training are provided in section \ref{res}. Conclusions and outlooks are given in the last section \ref{concl}.

\begin{figure}[t!] 
	\centering
	\includegraphics[scale=0.75]{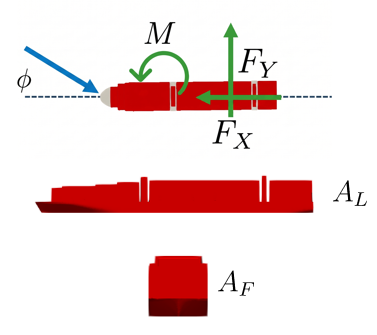}
	\caption{Definition of the reference wind direction and of the local coordinate system.}
	\label{ship_system}
\end{figure} 

\section{Multi-fidelity surrogate model}\label{data-AHFM}

This work addresses the modelling of wind-load coefficients, i.e. dimensionless force and moment coefficients acting on a ship under wind action. With reference to the local coordinate system indicated in Figure \ref{ship_system}, the longitudinal force coefficient is defined as:
\begin{equation}
C_X = \frac{F_X}{\frac{1}{2}\rho U^2 A_T},
\end{equation}
where $F_X$ is the longitudinal force, $U$ is the reference wind velocity, $\rho$ is the air density, and $A_T$ is the transverse projected area of the ship. The lateral wind-load coefficient is defined as:
\begin{equation}
C_Y = \frac{F_Y}{\frac{1}{2}\rho U^2 A_L},
\end{equation}
where $A_L$ is the lateral projected area. The yaw moment coefficient reads as:
\begin{equation}
C_M = \frac{M}{\frac{1}{2}\rho U^2 A_L L_{OA}},
\end{equation}
where $M$ is the yaw moment and $L_{OA}$ is the overall ship length.

The objective of the present work is to develop a fast and accurate predictive model for wind loads, which can be generally expressed as:
\begin{equation}
C_i(\mathbf{x}) = \mathcal{M}_i(\mathbf{x},\boldsymbol{\theta}),
\label{surrogate_model}
\end{equation}
where $i \in \{X,Y,M\}$ and the vector $\mathbf{x} \in \mathbb{R}^d$ denotes the set of input parameters describing the wind conditions, the geometrical properties of the ship, and, potentially, characteristics of the surrounding environment. In this work, the model inputs are linearly scaled using the minimum and maximum values defined by the validity range of the empirical correlation that will be introduced in section \ref{isherwood}. The vector $\boldsymbol{\theta}$ represents the model parameters to be calibrated from data.
In practice, the surrogate models are trained on a reduced set of input parameters, obtained through the sensitivity analysis described in Section \ref{active_subspace}. The resulting reduced input vector is denoted as $\mathbf{z}$, and the surrogate models can therefore be interpreted as mappings $\mathcal{M}_i(\mathbf{z},\boldsymbol{\theta})$. 

The database used to train the models is heterogeneous, i.e. it combines data from multiple fidelity levels (detailed CFD, simplified CFD, and empirical correlations), which will be described in Section \ref{data_sources}. For each fidelity level $t$, we define a dataset $\mathcal{D}^t = \{\mathbf{X}^t, \mathbf{C}^t\}$, where $\mathbf{X}^t \in \mathbb{R}^{N_t \times d}$ collects the input parameters, $\mathbf{C}^t \in \mathbb{R}^{N_t \times 3}$ collects the corresponding wind-load coefficients, and $N_t$ is the number of samples at fidelity level $t$. In the following, independent surrogate models are constructed for each coefficient $C_i$.

The following subsections illustrate the mathematical formulation of the surrogate model introduced in Eq.~\eqref{surrogate_model} and its training strategy to exploit the multi-fidelity nature of the data. Specifically, single- and multi-fidelity Gaussian Process regression are introduced in Sections \ref{GP} and \ref{MFGP}, respectively. Section \ref{active_subspace} presents the sensitivity-based approach used to define the reduced input space, while Section \ref{sobol_indices} introduces the global sensitivity indices employed to analyse the dependence of the wind-load coefficients on the model inputs.

\subsection{Single-fidelity Gaussian Process Regression}\label{GP}
In the present work, each wind-load coefficient $C_i(\mathbf{x})$ is modelled independently using Gaussian Process (GP) regression \cite{williams2006gaussian}. A GP defines a stochastic process such that any finite collection of function values follows a multivariate normal distribution. For each coefficient, the surrogate model $\mathcal{M}_i(\mathbf{x})$ is thus defined as
\begin{equation}
\mathcal{M}_i(\cdot) \sim \mathcal{GP}\big(m_i(\cdot), k_i(\cdot,\cdot)\big),
\end{equation}
where $m_i(\mathbf{x})$ is the mean function and $k_i(\mathbf{x},\mathbf{x}')$ is the covariance (kernel) function.

In this work, a constant mean function $m_i(\mathbf{x}) = \hat{\mu}_i$ is adopted and the covariance between two input points is modelled using a squared exponential kernel with anisotropic length scales:
\begin{equation}
k_i(\mathbf{x},\mathbf{x}') = \sigma_i^2 \exp\left(-\frac{1}{2} (\mathbf{x}-\mathbf{x}')^T \mathbf{L}_i^{-1} (\mathbf{x}-\mathbf{x}') \right),
\end{equation}
where $\sigma_i^2$ is the signal variance and $\mathbf{L}_i = \mathrm{diag}(l_{i,1}^2, \dots, l_{i,d}^2)$ is a diagonal matrix collecting the characteristic length scales along each input dimension. Since the three wind-load coefficients are modelled independently, the kernel functions and means are independent for each coefficient, thereby producing three separate Gaussian Process models.

Given a training dataset $\mathcal{D} = \{\mathbf{X}, \mathbf{c}_i\}$, where $\mathbf{c}_i \in \mathbb{R}^N$ denotes the observations of the $i$-th coefficient, conditioning the joint Gaussian distribution yields the following expressions for the predictive mean and variance at a new input location $\mathbf{x}^*$:

\begin{equation}
\mu_i(\mathbf{x}^*) = \hat{\mu}_i + \mathbf{K}_i(\mathbf{x}^*,\mathbf{X})\,\boldsymbol{\alpha}_i,
\end{equation}

with
\begin{equation}
\boldsymbol{\alpha}_i =
\left(\mathbf{K}_i(\mathbf{X},\mathbf{X}) + \sigma_\varepsilon^2 \mathbf{I}\right)^{-1}
\left(\mathbf{c}_i - \hat{\mu}_i \mathbf{1}\right),
\label{vector_alpha_sf}
\end{equation}

and

\begin{equation}
\sigma_i^2(\mathbf{x}^*) =
k_i(\mathbf{x}^*,\mathbf{x}^*) -
\mathbf{K}_i(\mathbf{x}^*,\mathbf{X})\,\mathbf{v}_i(\mathbf{x}^*),
\label{sigma_GP}
\end{equation}

with
\begin{equation}
\mathbf{v}_i(\mathbf{x}^*) =
\left(\mathbf{K}_i(\mathbf{X},\mathbf{X}) + \sigma_\varepsilon^2 \mathbf{I}\right)^{-1}
\mathbf{K}_i(\mathbf{x}^*,\mathbf{X})^T.
\label{vector_v_sf}
\end{equation}

In both \eqref{vector_alpha_sf} and \eqref{vector_v_sf}, $\sigma_\varepsilon^2$ is a nugget parameter introduced for numerical regularization.

The surrogate model introduced in Eq.~\eqref{surrogate_model} depends on a set of parameters $\boldsymbol{\theta}_i$, which in the present case collect the mean, kernel parameters, and regularization term $
\boldsymbol{\theta}_i = [\hat{\mu}_i, \sigma_i^2, l_{i,1}, \dots, l_{i,d}, \sigma_\varepsilon^2].$

These hyperparameters are determined by maximizing the log-marginal likelihood of the observed data:
\begin{equation}
\boldsymbol{\theta}_i = \arg\max_{\boldsymbol{\theta}_i} \ \mathcal{L}(\boldsymbol{\theta}_i; \mathbf{X}, \mathbf{c}_i),
\end{equation}
where the log-marginal likelihood reads:
\begin{align}
\mathcal{L} =\; & -\frac{1}{2} \log|\mathbf{K}_i + \sigma_\varepsilon^2 \mathbf{I}| \nonumber \\
& - \frac{1}{2} (\mathbf{c}_i - \hat{\mu}_i \mathbf{1})^T 
(\mathbf{K}_i + \sigma_\varepsilon^2 \mathbf{I})^{-1} 
(\mathbf{c}_i - \hat{\mu}_i \mathbf{1}) \nonumber \\
& - \frac{N}{2} \log(2 \pi).
\end{align}

\subsection{Recursive formulation}\label{MFGP}

The multi-fidelity formulation extends the single-fidelity surrogate model $\mathcal{M}_i(\mathbf{x},\boldsymbol{\theta})$ introduced in Section~\ref{GP} by considering a hierarchy of models $\mathcal{M}_i^t(\mathbf{x})$, $t=1,\dots,s$, corresponding to increasing levels of fidelity. In this framework, $C_i^t(\mathbf{x}) = \mathcal{M}_i^t(\mathbf{x})$ denotes the prediction of the $i$-th wind-load coefficient at fidelity level $t$, with the highest level $t=s$ defining the surrogate of interest.

To combine information across fidelity levels, we adopt the autoregressive multi-fidelity model of Kennedy and O’Hagan~\cite{kennedy2000predicting}. In practice, we employ its recursive training formulation, which allows the model to be constructed sequentially across fidelity levels as proposed by Le Gratiet~\cite{le2013multi,le2014recursive}.. For each coefficient $i$, the datasets $\mathcal{D}_i^t = \{\mathbf{X}^t, \mathbf{c}_i^t\}$ are assumed to be hierarchically nested, such that higher-fidelity observations are available at a subset of the lower-fidelity input locations:
$\mathbf{X}^s \subseteq \mathbf{X}^{s-1} \subseteq \dots \subseteq \mathbf{X}^1.$

The relation between successive fidelity levels is described by the autoregressive model:
\begin{equation}
C_i^t(\mathbf{x}) = \rho_{i,t-1} \, C_i^{t-1}(\mathbf{x}) + r_{i,t}(\mathbf{x}),
\label{MF_eq}
\end{equation}
where $\rho_{i,t-1}$ is an autoregressive scaling factor (or inter-level scaling coefficient) relating the predictions at fidelity levels $t-1$ and $t$, and $r_{i,t}(\mathbf{x})$ is a Gaussian Process $r_{i,t}(\cdot) \sim \mathcal{GP}\big(m_{i,t}(\cdot), k_{i,t}(\cdot,\cdot)\big)$ modelling the discrepancy between the two levels, with mean and covariance functions parametrized as in the single-fidelity case.

Following the notation introduced in Section~\ref{GP}, the set of hyperparameters at fidelity level $t$ is denoted by $\boldsymbol{\theta}_{i,t}$ and collects the parameters of the residual Gaussian Process together with the inter-level scaling coefficient:
\begin{equation}
\boldsymbol{\theta}_{i,t} = \left[\hat{\mu}_{i,t}, \sigma_{i,t}^2, l_{i,t,1}, \dots, l_{i,t,d}, \sigma_{\varepsilon,t}^2, \rho_{i,t-1} \right].
\end{equation}

This formulation implies that the prediction at level $t$ is obtained as a corrected version of the lower-fidelity prediction, where the correction is provided by the residual Gaussian Process. When $\rho_{i,t-1}$ is small, the contribution of the lower-fidelity model is reduced, and the prediction relies primarily on the residual model at level $t$.

After training, the predictive mean and variance at a new location $\mathbf{x}^*$ and fidelity level $t$ read as (see \cite{le2013multi} for the derivation):
\begin{equation}
C_i^t (\mathbf{x}^*) = \hat{\mu}_{i,t} + \rho_{i,t-1} C_{i}^{t-1}(\mathbf{x}^*)
+ \mathbf{K}_t (\mathbf{x}^*,\mathbf{X}^t)\,\boldsymbol{\alpha}_{i,t}\, , 
\end{equation} with 

\begin{equation}
\label{vector_a}
\boldsymbol{\alpha}_{i,t} = 
\left(\mathbf{K}_t(\mathbf{X}^t,\mathbf{X}^t) + \sigma_{\varepsilon,t}^2 \mathbf{I}\right)^{-1}
\left[ \mathbf{c}_i^t - \hat{\mu}_{i,t} \mathbf{1} - \rho_{i,t-1} C_i^{t-1}(\mathbf{X}^t) \right].
\end{equation}

and 

\begin{equation}
\sigma_{i,t}^2 (\mathbf{x}^*) =
\rho_{i,t-1}^2 \sigma_{i,t-1}^2(\mathbf{x}^*)
+ k_t(\mathbf{x}^*,\mathbf{x}^*)
- \mathbf{K}_t(\mathbf{x}^*,\mathbf{X}^t)\,\mathbf{v}_{i,t}(\mathbf{x}^*)\,,
\label{unc_multi_fidelity}
\end{equation} with 

\begin{equation}
\label{vector_v}
\mathbf{v}_{i,t}(\mathbf{x}^*) =
\left(\mathbf{K}_t(\mathbf{X}^t,\mathbf{X}^t) + \sigma_{\varepsilon,t}^2 \mathbf{I}\right)^{-1}
\mathbf{K}_t(\mathbf{x}^*,\mathbf{X}^t)^T.
\end{equation} 

In both \eqref{vector_a} and \eqref{vector_v}, $\sigma_{\varepsilon,t}^2$ is a nugget parameter introduced for numerical regularization at fidelity level $t$.

As for the single-fidelity setting, the surrogate model at each fidelity level depends on a set of hyperparameters $\boldsymbol{\theta}_{i,t}$, defined as
\begin{equation}
\boldsymbol{\theta}_{i,t} = \left[\hat{\mu}_{i,t}, \sigma_{i,t}^2, l_{i,t,1}, \dots, l_{i,t,d}, \sigma_{\varepsilon,t}^2, \rho_{i,t-1} \right].
\end{equation}

These hyperparameters are estimated by maximizing the log-marginal likelihood at each fidelity level, following the same procedure as in the single-fidelity formulation.

\subsection{Active subspace} \label{active_subspace}

This work employs the Active Subspace method~\cite{constantine2014active} to identify the most influential directions in the input space affecting the wind-load coefficients. For each coefficient $C_i$, the method is based on the construction of the covariance matrix of the gradients, estimated from the lowest-fidelity data source ($t=1$) as
\begin{equation}
\boldsymbol{\Sigma}_{AS,i} \approx \frac{1}{N_1} \sum_{k=1}^{N_1} \nabla C_i (\mathbf{x}_k) \, \nabla C_i (\mathbf{x}_k)^T,
\end{equation}
where the expectation is taken over the input space and $N_1$ is the number of samples considered in the evaluation.
In practice, gradients are estimated from a smooth surrogate of the empirical correlations (see Section~\ref{isherwood}), obtained by training single-fidelity Gaussian Process models for each coefficient and evaluating gradients at $N_1=500$ randomly selected input locations.

The dominant directions in the parameter space for each coefficient are identified from the leading eigenvectors of $\boldsymbol{\Sigma}_{AS,i}$. A reduced basis is constructed by retaining the first $p$ eigenvectors, collected in the matrix $\mathbf{W}_i^{(p)} \in \mathbb{R}^{d \times p}$. The proportion of variability captured by these directions is quantified by the associated eigenvalues as ${\mathrm{tr}(\boldsymbol{\Lambda}_{AS,i}^{(p)})}/{\mathrm{tr}(\boldsymbol{\Lambda}_{AS,i})}$.The input parameters can be projected onto the reduced subspace spanned by the leading eigenvectors, leading to a lower-dimensional representation $\mathbf{z} = \mathbf{W}_i^{(p)\,T} \mathbf{x}$. 

In this work, this projection was used as a diagnostic tool to assess the relative importance of the input parameters. The analysis revealed that several parameters have a negligible contribution to the active directions, indicating a limited influence on the wind-load coefficients. Based on this observation, a reduced set of input variables is selected for the surrogate modelling (see Section \ref{sec5p1}). In the following, this feature-selected representation $\mathbf{z}$ is used in place of the original input vector $\mathbf{x}$.

\subsection{Sobol sensitivity indices} \label{sobol_indices}

The developed surrogate model is further exploited for a global sensitivity analysis of the wind-load coefficients with respect to the input parameters. In this work, Sobol sensitivity indices~\cite{saltelli2004sensitivity} are used as a post-hoc analysis tool to quantify the relative importance of the selected variables and to interpret the surrogate model.

The Sobol indices rely on the decomposition of the global variance of the model response $C_i(\mathbf{z})$:
\begin{equation}
\mathrm{Var}(C_i(\mathbf{z})) = \sum_{k=1}^p V_{i,k} + \sum_{1 \leq k < j \leq p} V_{i,kj} + \dots + V_{i,1,2,\dots,p},
\label{dec_variance}
\end{equation}
where $p$ denotes the dimension of the reduced input space. The first-order contributions are defined as
\begin{equation}
V_{i,k} = \mathrm{Var}_{z_k} \left( \mathbb{E}_{\mathbf{z}_{\sim k}} \left[ C_i(\mathbf{z}) \mid z_k \right] \right),
\label{vi}
\end{equation}
where $\mathbf{z}_{\sim k}$ denotes all variables except $z_k$. Similarly, the second-order terms
\begin{equation}
V_{i,kj} = \mathrm{Var}_{z_k, z_j} \left( \mathbb{E}_{\mathbf{z}_{\sim k,j}} \left[ C_i(\mathbf{z}) \mid z_k, z_j \right] \right)
\label{vij}
\end{equation}
represent the contribution due to the interaction between $z_k$ and $z_j$, and higher-order terms are defined analogously.

By normalizing Eq.~\eqref{dec_variance} by the total variance, one obtains the Sobol sensitivity indices:
\begin{equation}
1 = \sum_{k=1}^p S_{i,k} + \sum_{1 \leq k < j \leq p} S_{i,kj} + \dots + S_{i,1,2,\dots,p},
\label{sobol}
\end{equation}
where $S_{i,k} = V_{i,k} / \mathrm{Var}(C_i(\mathbf{z}))$ are the first-order Sobol indices, measuring the contribution of each parameter independently.

The total Sobol index, which accounts for both first-order and interaction effects involving $z_k$, is defined as
\begin{equation}
S_{i,Tk} = \frac{\mathbb{E}_{\mathbf{z}_{\sim k}}\left[ \mathrm{Var}_{z_k} \left( C_i(\mathbf{z}) \mid \mathbf{z}_{\sim k} \right) \right]}{\mathrm{Var}(C_i(\mathbf{z}))}.
\end{equation}

In the present work, these indices are used to provide a complementary, variance-based interpretation of the surrogate model and to assess the influence of the selected input variables, in relation with the gradient-based insights obtained from the Active Subspace analysis.

\section{Data-sources}\label{data_sources}

In this work, the surrogate models are trained using data from multiple sources of increasing fidelity. These data sources differ in terms of physical modelling complexity and computational cost, and are combined within the recursive multi-fidelity framework introduced in Eq.~\eqref{MF_eq}.

The lowest-fidelity (LF) data are provided by empirical correlations (Section~\ref{isherwood}), while computational fluid dynamics (CFD) simulations are used to generate higher-fidelity data. A simplified CFD model is employed as a medium-fidelity (MF) source (Section~\ref{simplified}), whereas a detailed CFD model provides high-fidelity (HF) data (Section~\ref{detailed}).

The definition of the fidelity hierarchy depends on the target application. For surrogate models predicting wind loads in open-sea conditions, the detailed CFD model represents the highest-fidelity level. For surrogate models targeting specific harbour environments, the hierarchy is extended by introducing an additional level: the detailed CFD model in open-sea conditions is treated as an intermediate fidelity, while the highest-fidelity data are provided by detailed CFD simulations including harbour structures.

The validity of this hierarchical structure is assessed a posteriori through the consistency of the multi-fidelity model. The different data sources and their corresponding fidelity levels are described in the following subsections.

\subsection{Isherwood's correlation (low-fidelity)} \label{isherwood}

The lowest-fidelity data are provided by the empirical correlation proposed by Isherwood~\cite{isherwood1973wind}, which is based on experimental measurements for conventional merchant ships. This model expresses the wind-load coefficients as algebraic functions of geometrical parameters and the angle of attack $\phi$, defined in Fig.~\ref{ship_system}.

The longitudinal and lateral force coefficients and the yaw moment coefficient are given by:
\begin{align}
C_X &= A_0(\phi) + A_1(\phi) \frac{2 A_L}{L_{OA}^2}
+ A_2(\phi) \frac{A_T}{B^2}
+ A_3(\phi) \frac{L_{OA}}{B} \nonumber \\
&\quad + A_4(\phi) \frac{S}{L_{OA}}
+ A_5(\phi) \frac{C}{L_{OA}}
+ A_6(\phi)\, M,
\label{isherwood_X}
\end{align}

\begin{align}
C_Y &= B_0(\phi) + B_1(\phi) \frac{2 A_L}{L_{OA}^2}
+ B_2(\phi) \frac{A_T}{B^2}
+ B_3(\phi) \frac{L_{OA}}{B} \nonumber \\
&\quad + B_4(\phi) \frac{S}{L_{OA}}
+ B_5(\phi) \frac{C}{L_{OA}}
+ B_6(\phi) \frac{A_{SS}}{A_L},
\label{isherwood_Y}
\end{align}

\begin{align}
C_M &= D_0(\phi) + D_1(\phi) \frac{2 A_L}{L_{OA}^2}
+ D_2(\phi) \frac{A_T}{B^2}
+ D_3(\phi) \frac{L_{OA}}{B} \nonumber \\
&\quad + D_4(\phi) \frac{S}{L_{OA}}
+ D_5(\phi) \frac{C}{L_{OA}}.
\label{isherwood_M}
\end{align}

Here, $L_{OA}$ is the overall ship length and $B$ is the beam, while $A_L$ and $A_T$ denote the lateral and transverse (frontal) projected areas, respectively. The parameter $S$ is the perimeter of the lateral projection, $C$ is the distance of the centroid of the lateral projection from the bow, $A_{SS}$ is the area of the superstructure, and $M$ is the number of distinct container groups. The coefficients $A_i(\phi)$, $B_i(\phi)$, and $D_i(\phi)$ are tabulated functions of the angle of attack $\phi$, as reported in~\cite{isherwood1973wind}.

\begin{figure*}[htbp]
	\centering
	\includegraphics[scale=0.3]{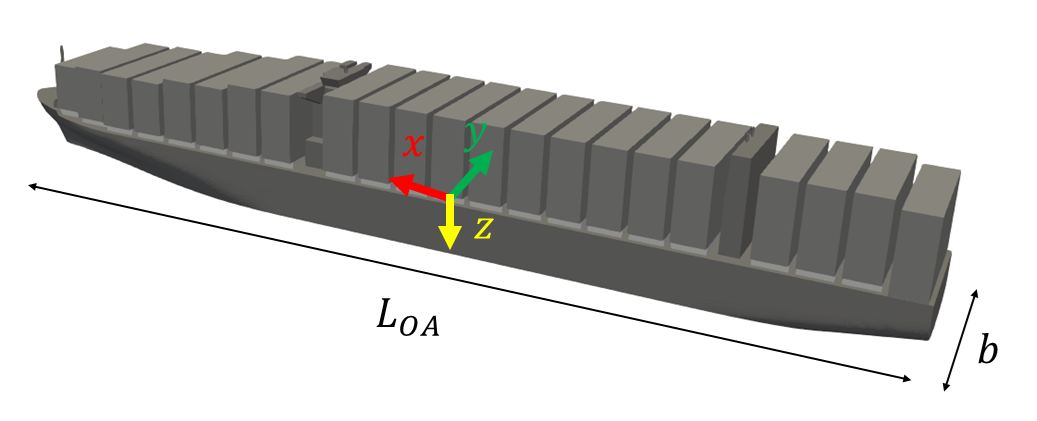}
	\caption{Container ship considered in the present work.}
	\label{container_ship}
\end{figure*}

\begin{figure*}[htbp]
    \centering
    \subfloat[][]{\includegraphics[width=0.33\textwidth]{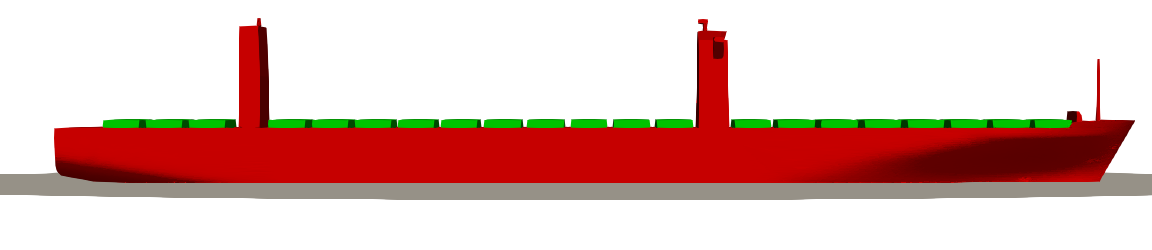}}\!
    \subfloat[][]{\includegraphics[width=0.33\textwidth]{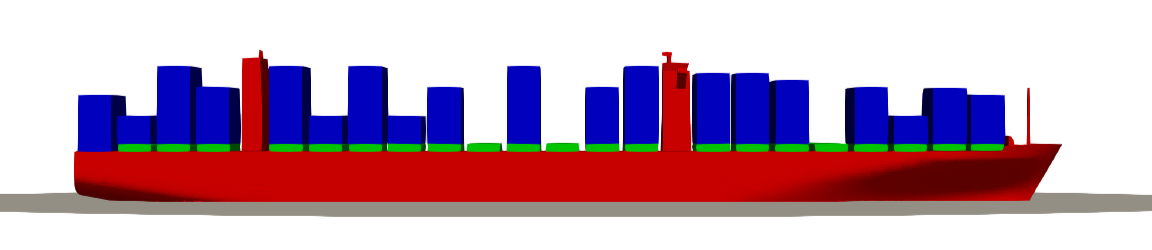}}\!
    \subfloat[][]{\includegraphics[width=0.33\textwidth]{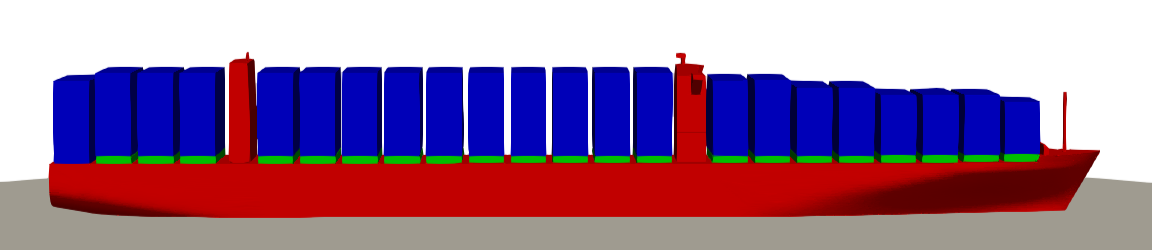}}\\
    \subfloat[][]{\includegraphics[width=0.24\textwidth]{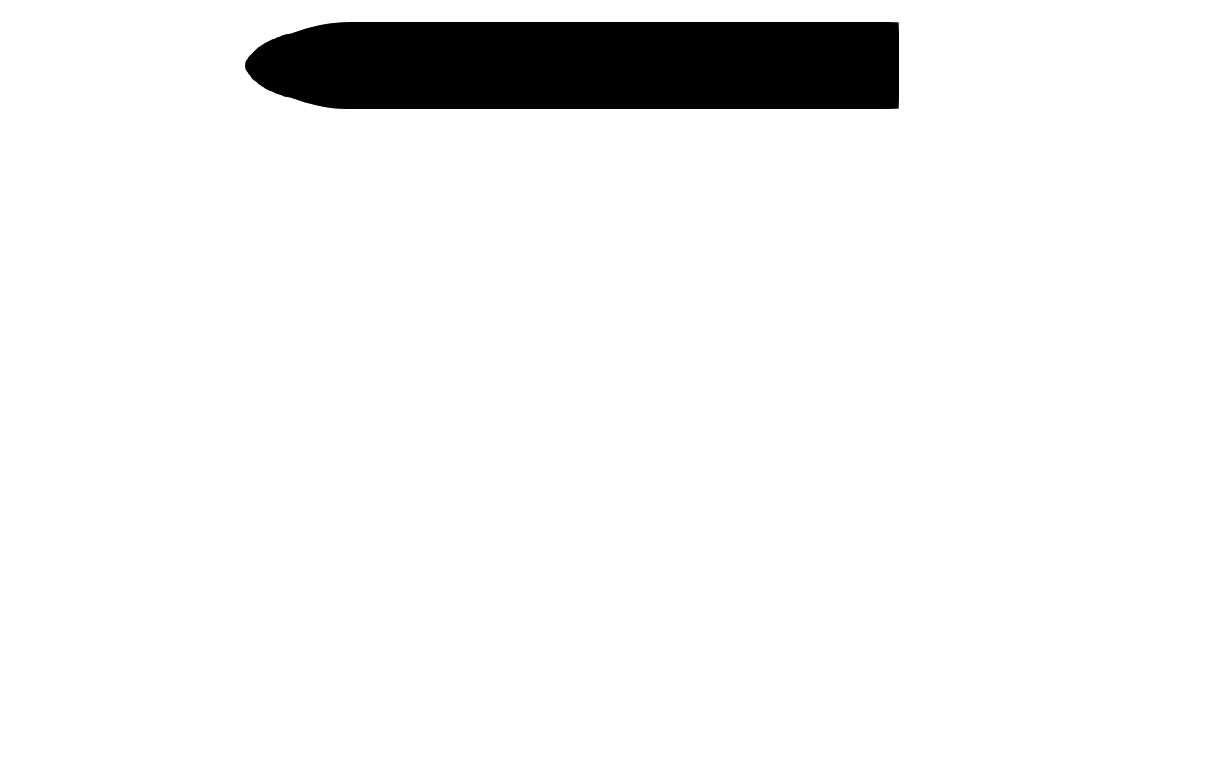}}\!
    \subfloat[][]{\includegraphics[width=0.24\textwidth]{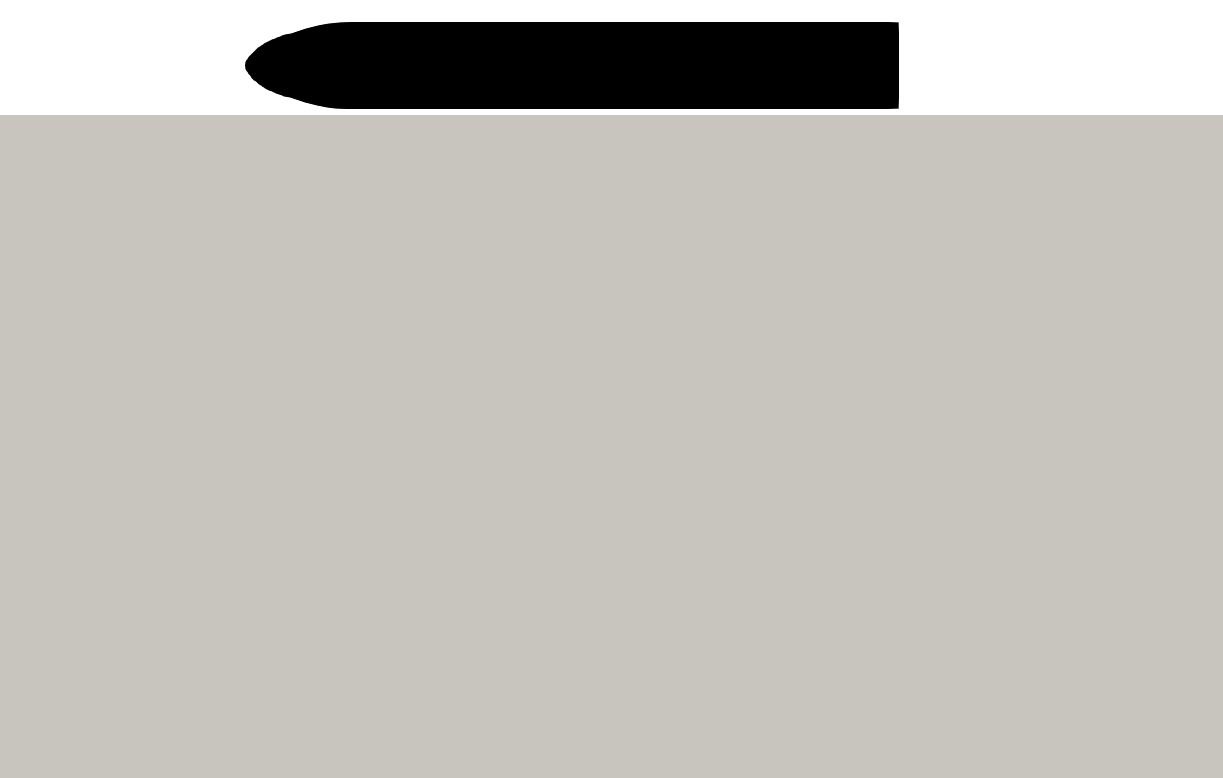}}\!
    \subfloat[][]{\includegraphics[width=0.24\textwidth]{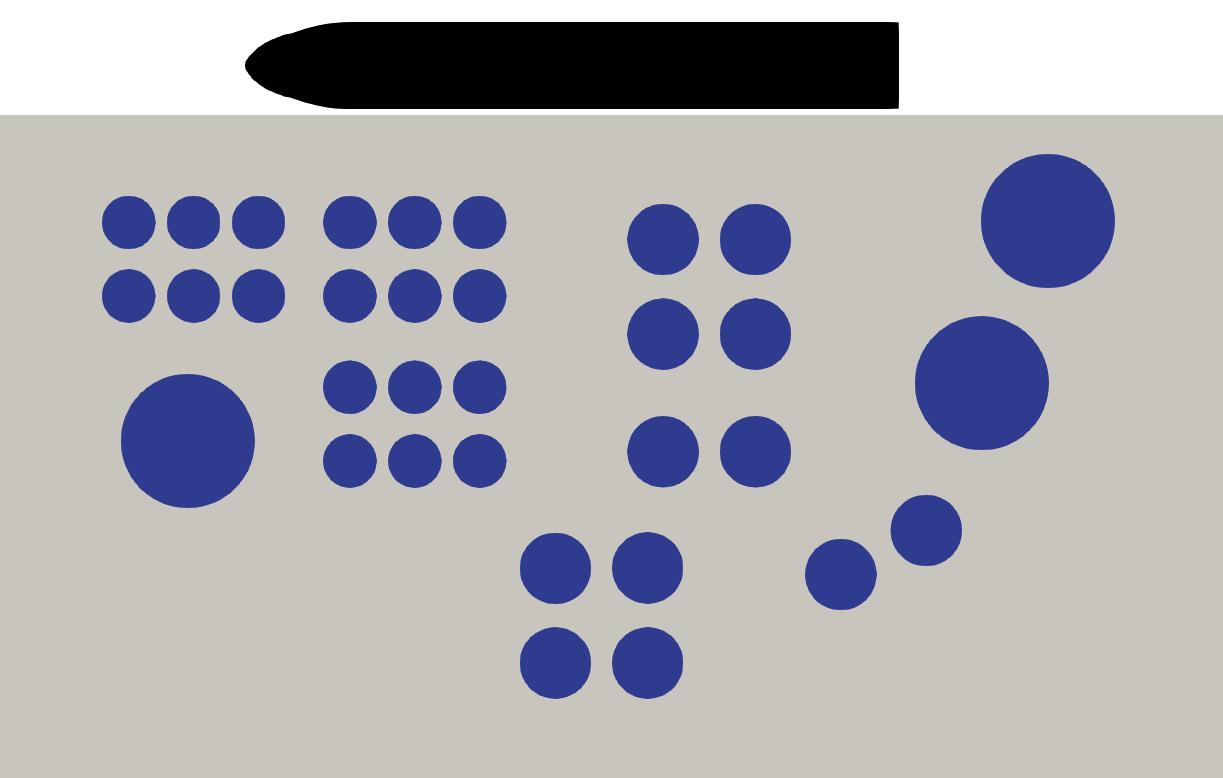}}\!
    \subfloat[][]{\includegraphics[width=0.24\textwidth]{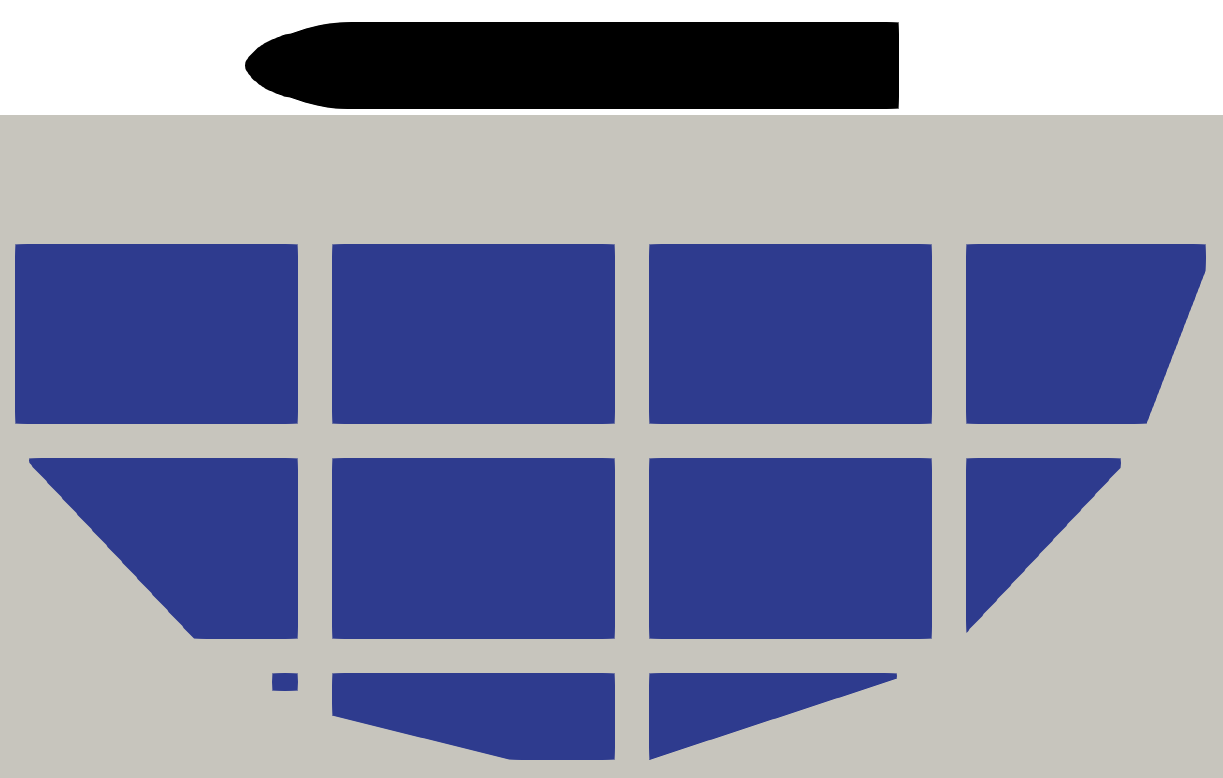}}
    \caption{The three loading configurations and four environments simulated: (a) empty ship, (b) intermediate load, (c) full load, (d) open sea, (e) empty quay, (f) tanks environment, and (g) container environment.}
    \label{fig:loadings_envs}
\end{figure*}

\begin{figure*}[htbp]
    \centering
    \subfloat[][]{\includegraphics[width=0.49\linewidth,trim=0cm 1cm 0cm 0cm, clip]{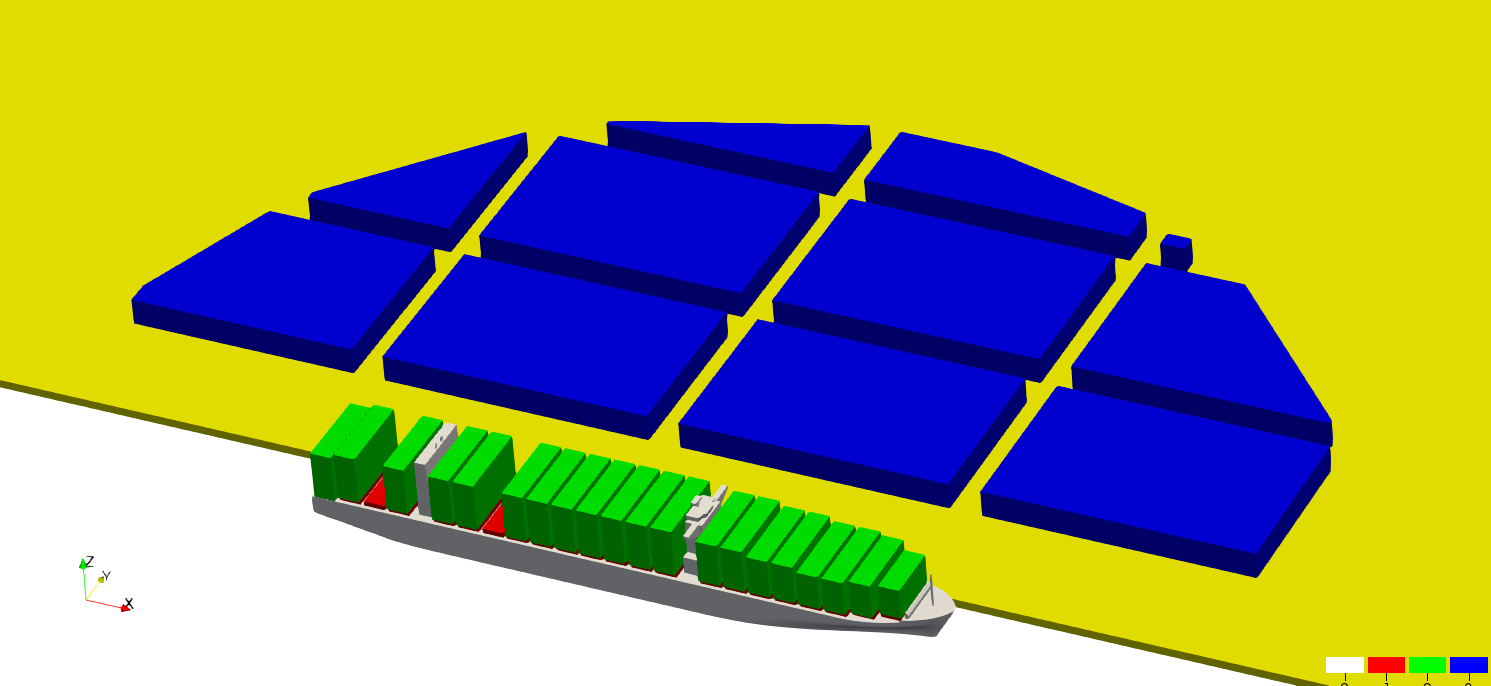}}
    \subfloat[][]{\includegraphics[width=0.49\linewidth,trim=0cm 1cm 0cm 0cm, clip]{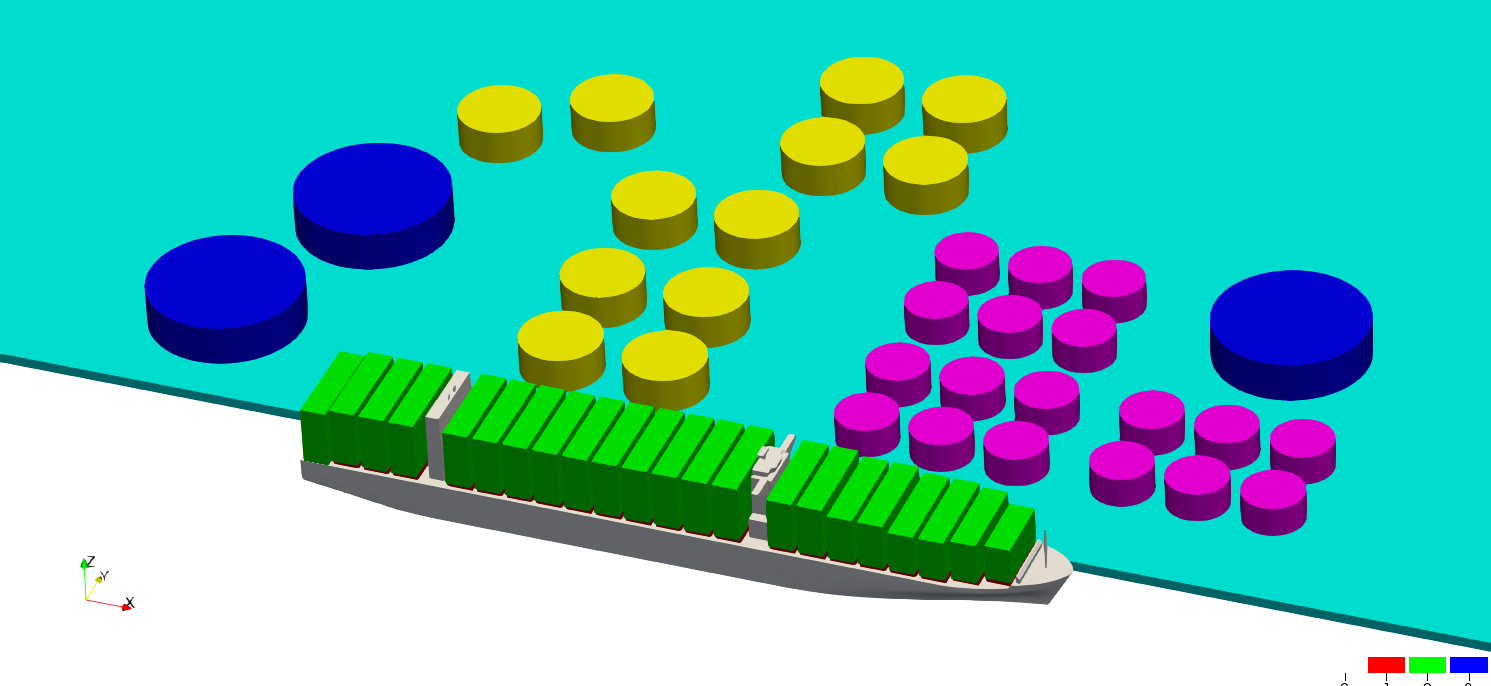}} \\
    \caption{(a) Containers environment and (b) tanks environment.}
    \label{fig:envs_3D}
\end{figure*}

\subsection{Detailed CFD (high-fidelity)} \label{detailed}
The detailed CFD model developed in OpenFoam v.9 reproduces the geometry of the container ship indicated in Figure \ref{container_ship} at its full load configuration. This model was constructed using general arrangement plans provided by the Ships and Marine
Technology Division at Ghent University, yielding a model with a Length On Waterline (LWL) of 1.4 m. The characteristic dimensions of the ship are reported in Table \ref{tab:dimensions}. The geometry depicted in Figure \ref{container_ship} was adopted for the wind tunnel experiments described in Ref. \cite{Arnoult2025_EACWE}. Figure \ref{fig:loadings_envs} depicts three loading configurations and four reference environments considered for wind tunnel tests. Specifically, the empty quay environments consists of a flat surface above the sea level, while the environments with containers and tanks are intended to model two generic harbour environments. A better visualization of these simplified harbour structures is provided by Figure \ref{fig:envs_3D}. 

\begin{figure*}[htbp]
    \centering
    \subfloat[][]{\includegraphics[width=0.33\linewidth]{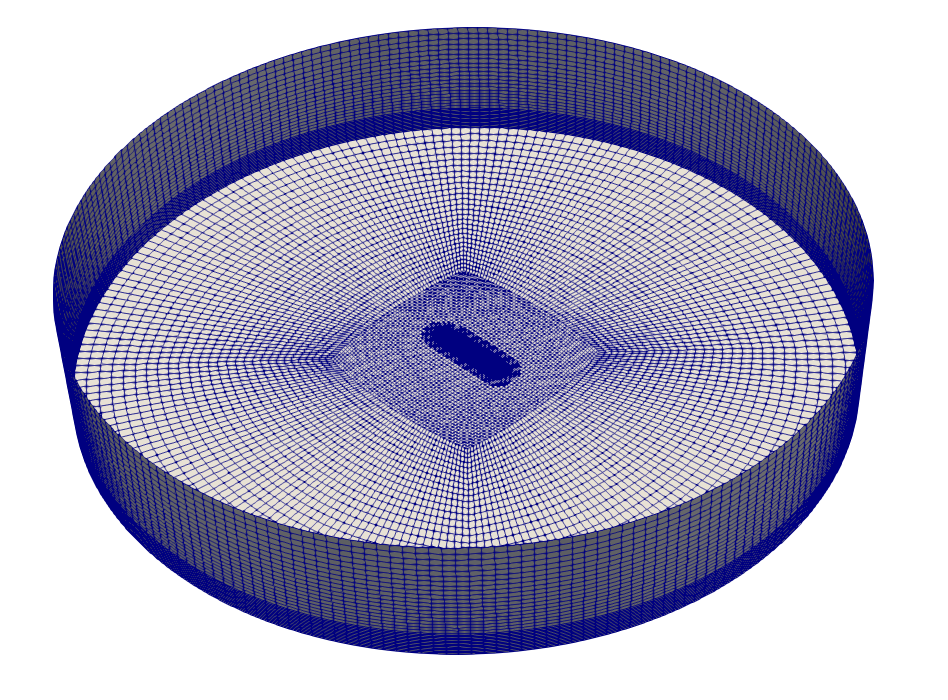}}
    \subfloat[][]{\includegraphics[width=0.33\linewidth]{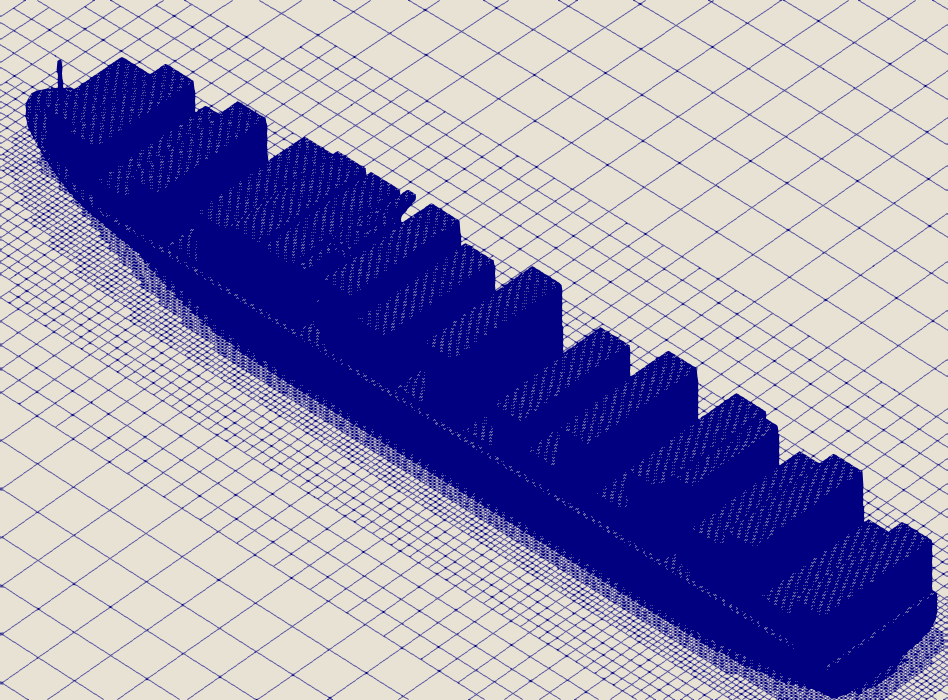}}
    \subfloat[][]{\includegraphics[width=0.33\linewidth]{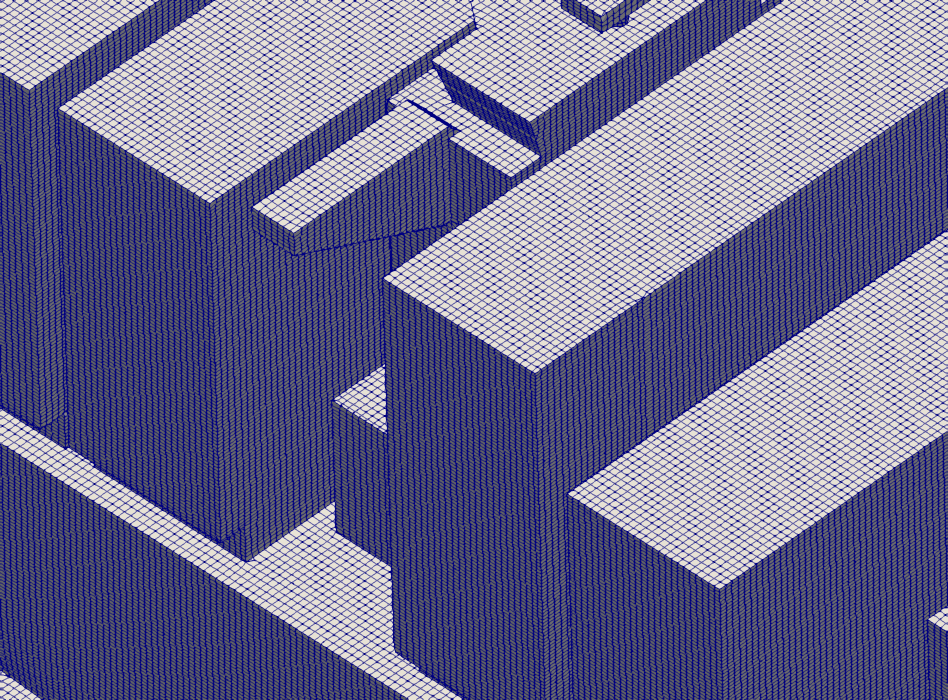}}
    \caption{Mesh used for the high-fidelity CFD approach for the open sea case and intermediate loading configuration. (a) View of the entire cylindrical domain, (b) close-up of the ship, (c) close-up of the ship details.}
    \label{fig:mesh_hf}
\end{figure*}

Aerodynamic loads on the container ship at full scale were computed by solving the incompressible Reynolds‑averaged Navier–Stokes equations with a standard $k$-$\epsilon$ turbulence model in OpenFOAM v9. A cylindrical domain with a diameter of 2.3 km and a height of 560 m was adopted following COST 732 guidelines, ensuring sufficient distance between the ship and the boundaries. At the lateral and top boundaries, a fixed-velocity inlet ($U = 10$~m/s at a reference height of 175~m) or zero pressure-gradient outflow was imposed, depending on the local flow direction. The mesh used for the open sea case with the intermediate loading configuration is shown in Fig.\ref{fig:mesh_hf}.  The inflow profiles were tuned so that the atmospheric boundary layer at the ship location matched wind‑tunnel measurements of offshore and rural terrain profiles when scaled by the 1:250 factor.

The computational meshes were generated with \textit{snappyHexMesh} and feature local refinement around the hull and harbor structures, with minimum cell sizes of 0.25–0.75 m near solid surfaces and up to 25 m in the far field. Depending on the environment (open sea, empty quay, quay with containers, quay with tanks) and loading condition, the resulting meshes range from 2.6 to 15.7 million cells. Pressure–velocity coupling is handled via the SIMPLE algorithm with under-relaxation factors of 0.3 for pressure and 0.7 for velocity, and second-order spatial schemes are used for all convective terms.

\begin{figure*}[htbp]
    \centering
    \subfloat[][]{\includegraphics[width=0.39\linewidth]{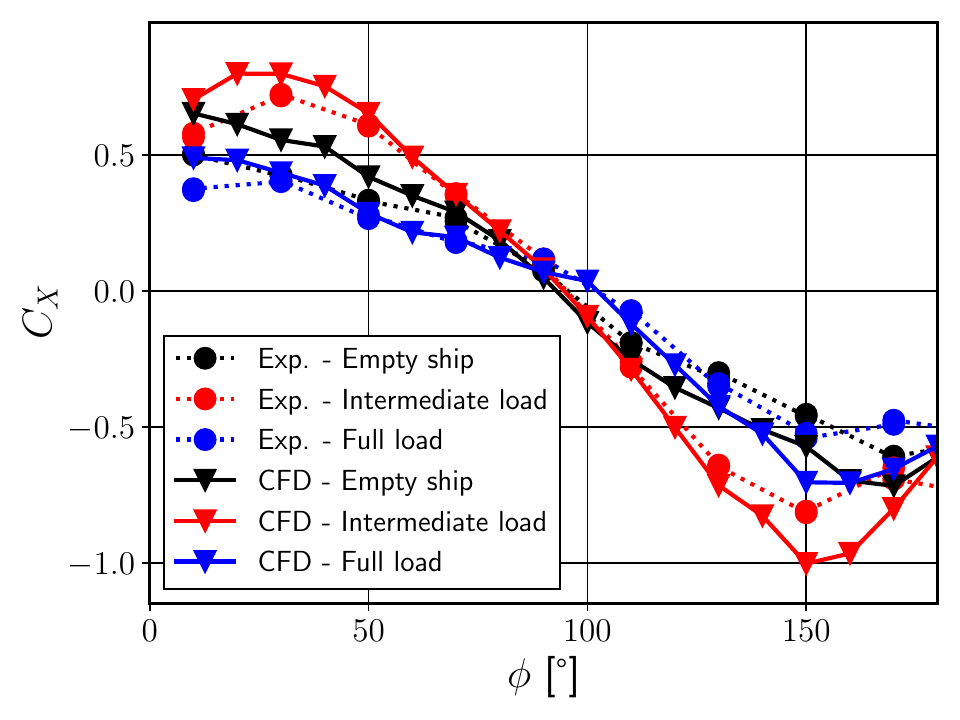}}
    \subfloat[][]{\includegraphics[width=0.39\linewidth]{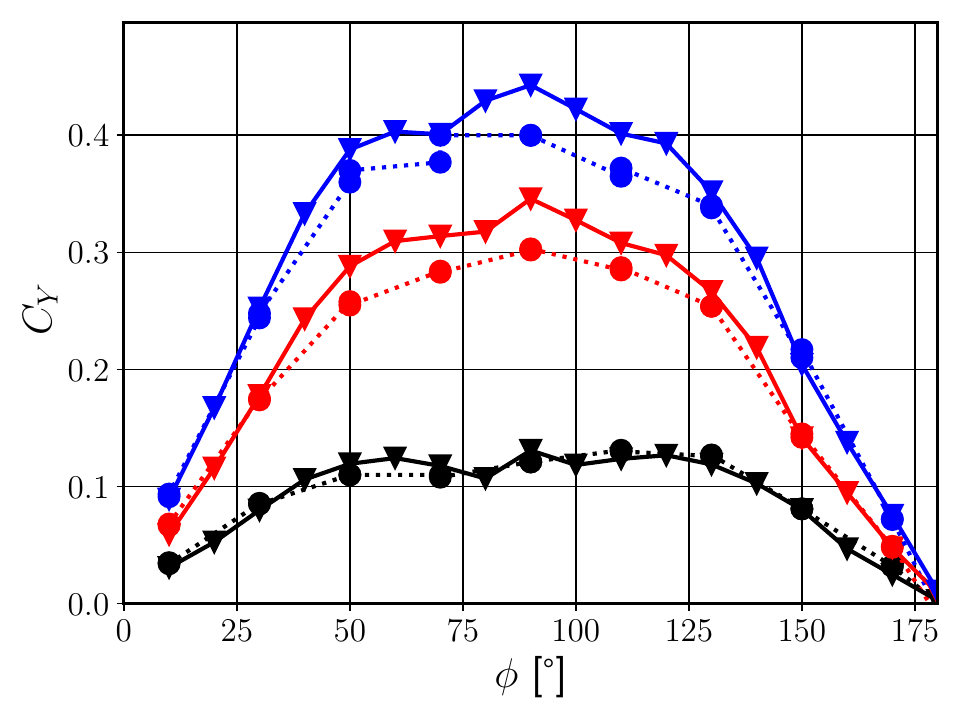}} \\
    \subfloat[][]{\includegraphics[width=0.39\linewidth]{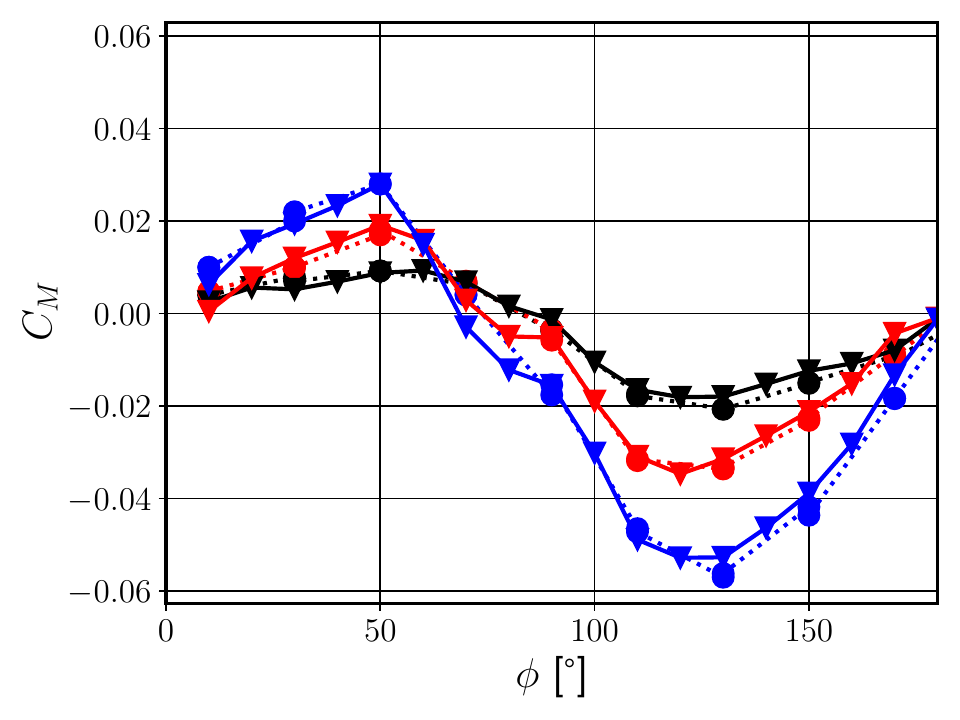}}
    \caption{Comparison between measurements and the high-fidelity CFD approach for the tanks environment. (a) Longitudinal force coefficient, (b) lateral force coefficient, (c) yaw moment coefficient.}
    \label{fig:force_coeff_hf}
\end{figure*}

This numerical setup has been validated at wind‑tunnel scale against force, moment, and surface‑pressure measurements obtained in the VKI L1‑B facility for three loading configurations and four harbor environments, showing generally good agreement over the full range of wind directions relevant for harbor operations. An example of the comparison between the numerical and experimental coefficients for the tank environment is shown in Fig.\ref{fig:force_coeff_hf}. The agreement between experimental and numerical results is satisfactory for all the wind load coefficients examined. This validation justifies the use of this CFD model as a high-fidelity data-source in the present investigation. More details regarding the numerical setup and validation can be found in Ref.\cite{bresciani2026}.

\begin{table}[]
    \centering
    \caption{Characteristic dimensions of the full scale model.}
    \begin{tabular}{ccc}
    \toprule
    Name & Symbol & Value \\
    \midrule
     Length overall & $L_{OA}$ & 365.5 m \\
     Beam & $B$ & 48.25 m \\
     Air draft &$d$ & 56.25 m \\
     Side area & $A_L$ & 15000 m$^2$ \\
     Frontal area & $A_T$ & 2500 m$^2$ \\
     Tanks heights & $h_1, h_2, h_3$ & 15 m, 17.8 m, 20 m \\
     Quay height & $h_4$ & 5 m\\
     Container height & $h_5$ & 16 m\\
    \bottomrule
    \end{tabular}
    \label{tab:dimensions}
\end{table}

\subsection{Simplified CFD (medium fidelity)}\label{simplified}
The simplified setup was constructed in OpenFOAM v9 at wind tunnel scale (1:250), consistent with the experimental configuration presented in \cite{Arnoult2025_EACWE}, rather than at full scale. For the same wind velocity, this results into a lower Reynolds number compared to the full scale configuration. However, preliminary CFD analyses \cite{bresciani2026} showed that scale effects are minimal in the present case and can be neglected. 

\begin{figure*}[!htbp]
	\centering
	\includegraphics[scale=0.6]{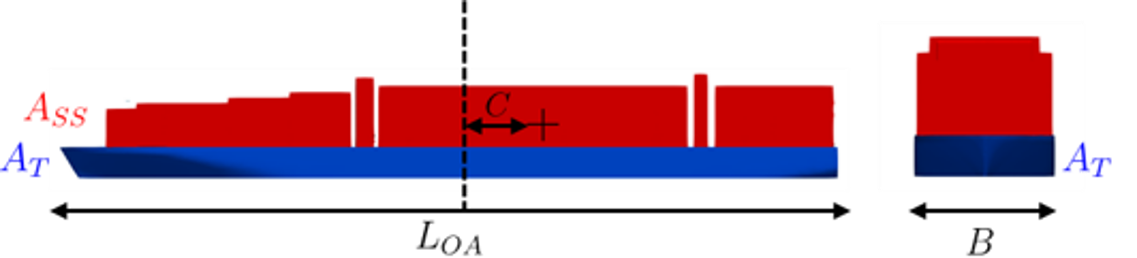}
	\caption{Geometry of the simplified ship employed in the setup of medium fidelity.}
	\label{simplified_ship}
\end{figure*} 

\begin{figure*}[htbp] 
	\centering
	\includegraphics[scale=0.5]{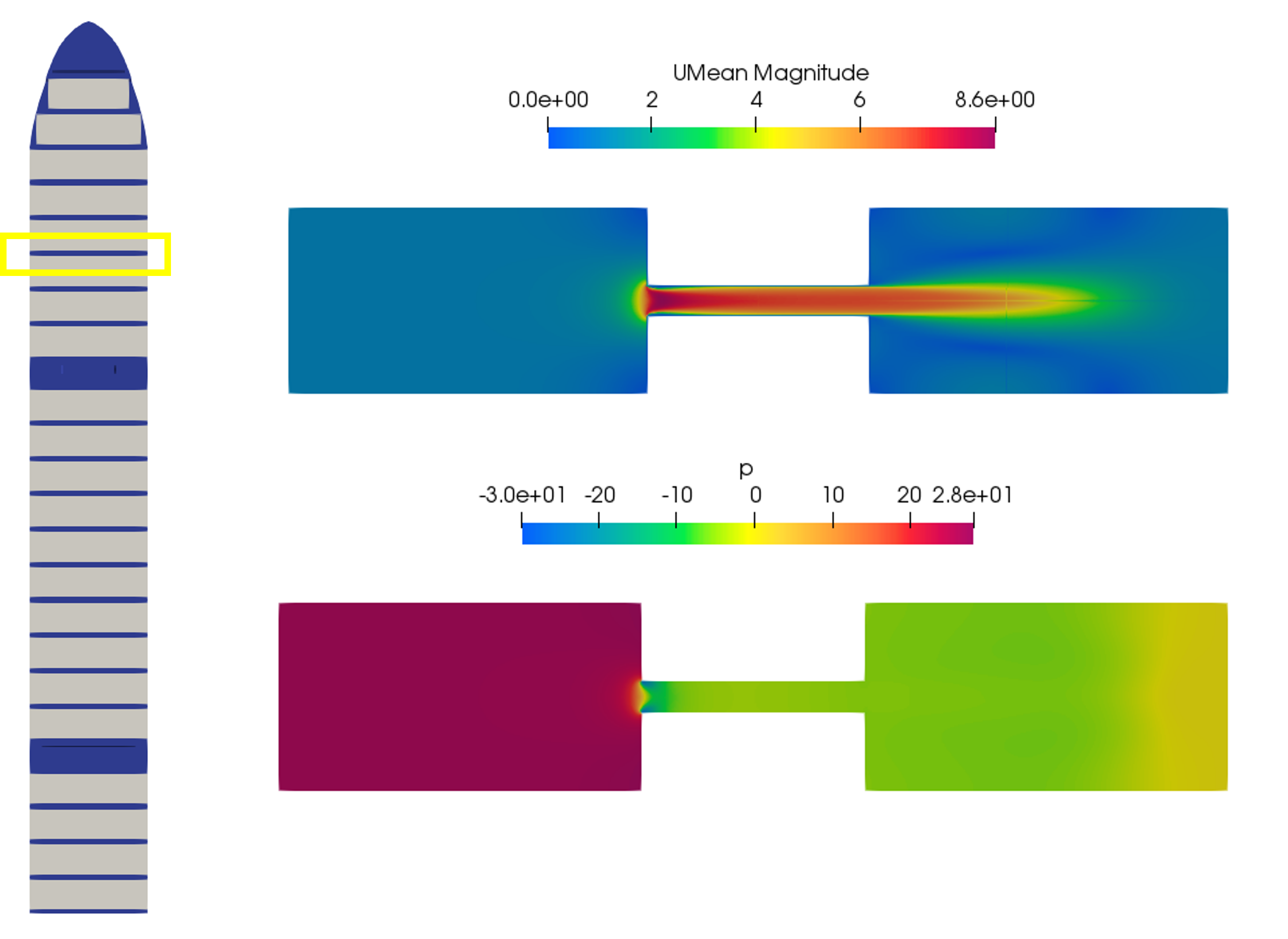}
	\caption{Domain employed to carry out precursor simulations of the flow within the gaps, to estimate the coefficients of the porous model. Velocity is in $m/s$ and the specific pressure is in $m^2/s^2$.}
	\label{precursor}
\end{figure*} 

\begin{figure}[htbp] 
	\centering
	\includegraphics[scale=0.45]{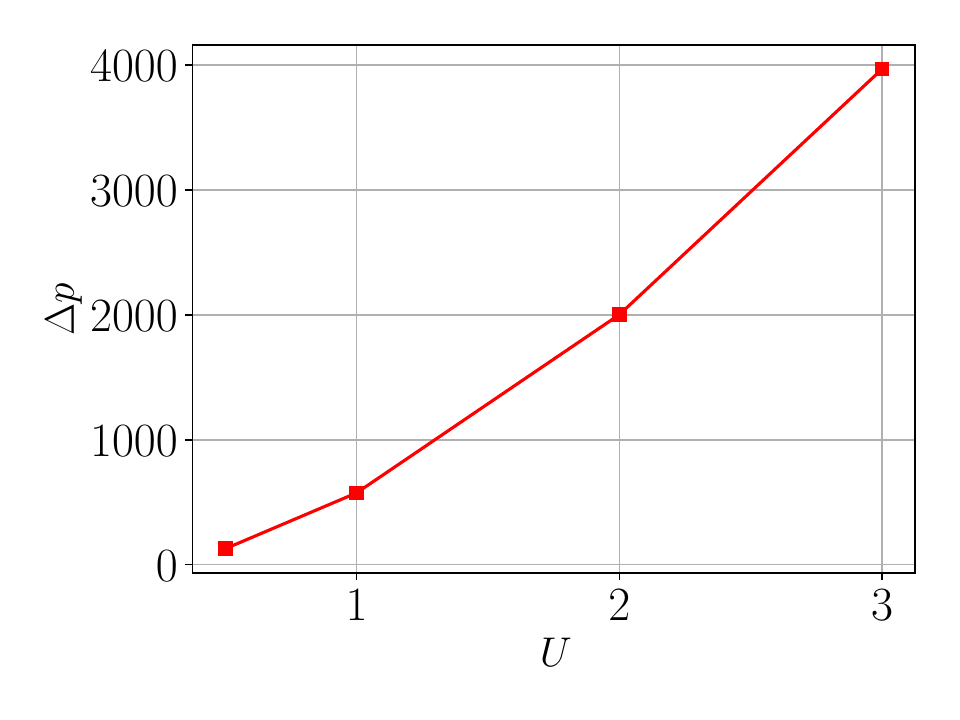}
	\caption{Velocity-pressure drop relationship for the flow within the gaps obtained from the precursor simulations.}
	\label{U_deltaP}
\end{figure}

\begin{figure*}[htbp] 
	\centering
	\includegraphics[scale=0.9]{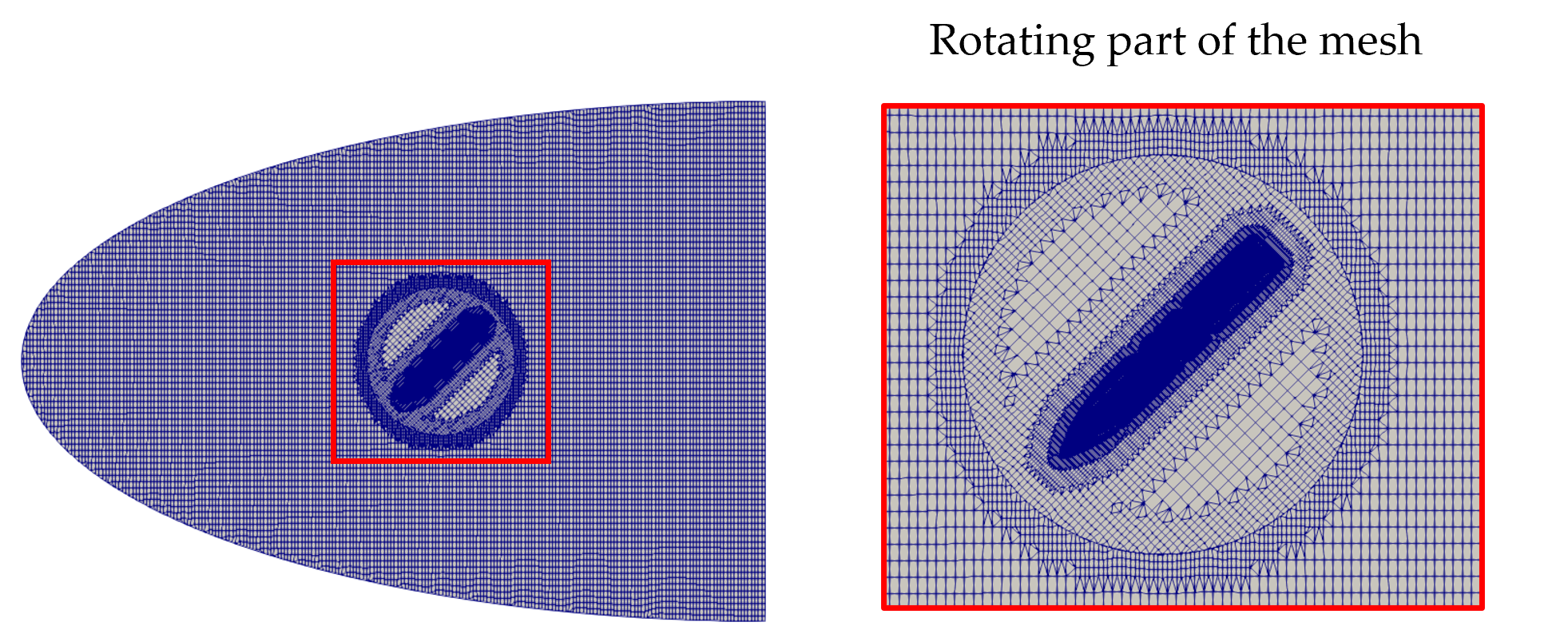}
	\caption{Snapshots of the mesh used in the simplified CFD setup, highlighting the rotating part of the mesh enabling to change the angle of attack.}
	\label{mesh_simplified}
\end{figure*}

The simplified CFD setup neglects minor geometrical details of the container ship, e.g. the gaps between adjacent containers, to reduce the overall mesh requirements. This results into the simplified geometry depicted in Figure \ref{simplified_ship} for the full load configuration. To account for the effect of such gaps on the pressure field, the superstructures are treated as porous regions modelled with the Darcy-Forchheimer law. The law expresses the pressure gradient through the porous medium as:
\begin{equation}
\nabla p = \mu \mathbf{D} \mathbf{U} + \frac{1}{2} \rho \text{ tr}(\mathbf{U} \cdot \mathbf{I}) \mathbf{F} \mathbf{U},
\end{equation}
where $\mathbf{U}$ is the velocity vector, $\mu$ is the air viscosity, and $\mathbf{D}$ and $\mathbf{F}$ are diagonal isotropic tensors, i.e.$\mathbf{D}=D \mathbf{I}$ and $\mathbf{F}=F \mathbf{I}$. 
 The linear and quadratic coefficients $D$ and $F$ are calibrated from precursor CFD simulations of the flow through a single gap at different inlet velocities. An example of those simulations of the gap regions is given in Figure \ref{precursor}. These tests resulted into the pressure drop-velocity curve reported in Figure \ref{U_deltaP}. Based on the obtained results, we imposed $D=2.8 \cdot 10^7$ $Pa \cdot s$ and $F=784.3$ $1/m$ for the modelled porous regions. 

The mesh built with \textit{snappyHexMesh} for the simplified CFD is depicted in 
Figure \ref{mesh_simplified}. The mesh consists of two separate parts that can rotate relative to each other, allowing the simulation of different angles of attack without any mesh or inlet modification. An overview of the different mesh sizes in the detailed and simplified CFD setups is given in Table \ref{tab:mesh_cases}. 

In the simplified setup, the $k-\omega$ turbulence model is used as momentum closure. At the C-type inlet we imposed a uniform velocity of $10$ $m/s$ (same as in the full scale model) and a uniform turbulence intensity of 10\%.

\begin{table*}[h!]
\centering
\caption{Mesh size and fidelity level of the different CFD setups.}
\resizebox{0.7\textwidth}{!}{
\begin{tabular}{l c r}
\hline
\textbf{Setup} & \textbf{Fidelity level} & \textbf{Number of Cells [M]} \\
\hline
Simplified CFD (MF)  & 2  & 2.4 \\
Detailed CFD (open sea, HF) &  3 & 4.7 \\
Detailed CFD (containers environment, HF-CE) &  4 & 15.7 \\
Detailed CFD (tanks environment, HF-TE) & 4 & 11.4 \\
\hline
\end{tabular}
}
\label{tab:mesh_cases}
\end{table*}

\section{Training algorithm} \label{training}

\begin{figure*}[htbp] 
	\includegraphics[scale=0.89]{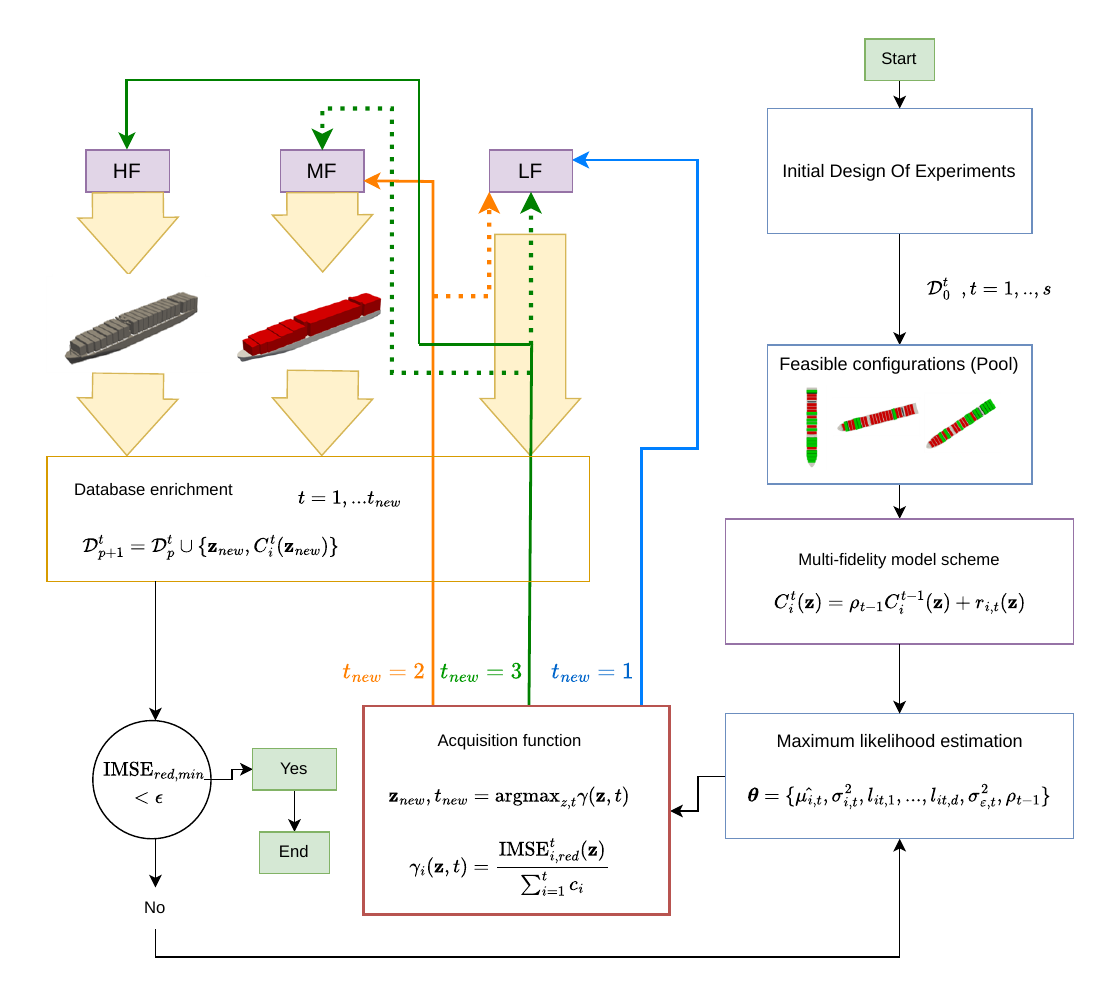}
	\caption{Schematics of the methodology adopted to train the multi-fidelity surrogate model for wind loads with three fidelity levels. The data-sources are here denoted as low-fidelity (empirical correlation, LF), medium fidelity (simplified CFD, MF) and high-fidelity (detailed CFD in open sea, HF).}
	\label{MF_algorithm}
\end{figure*}

The recursive multi-fidelity formulation introduced in Eq.~\eqref{MF_eq} is trained using the heterogeneous datasets described in Section~\ref{data_sources}. For wind-load prediction in open-sea conditions, the surrogate models $\mathcal{M}_i^s(\mathbf{z},\boldsymbol{\theta})$ are constructed using three fidelity levels ($s=3$): low-fidelity (LF) empirical correlations, medium-fidelity (MF) simplified CFD simulations, and high-fidelity (HF) detailed CFD simulations. The overall training strategy is illustrated in Fig.~\ref{MF_algorithm}.

The procedure follows an iterative, cost-aware design space exploration. It is initialized from a design of experiments based on selected loading configurations (intermediate and full load conditions in Fig.~\ref{fig:loadings_envs}), which define the initial training database. A pool of 5000 feasible configurations is generated by uniform sampling in the reduced input space $\mathbf{z}$, providing candidate points for sequential enrichment.

At each iteration, the following steps are performed:

\begin{enumerate}
    \item The multi-fidelity surrogate model $\mathcal{M}_i^s(\mathbf{z},\boldsymbol{\theta})$ is trained based on the current database $\mathcal{D}$, at all its fidelity levels. 

\item A new configuration $\mathbf{z}_{new}$ and its corresponding fidelity level $t_{new}$ are selected by maximizing an acquisition function $\gamma(\mathbf{z},t)$ over the pool:
\begin{equation}
(\mathbf{z}_{new}, t_{new}) = \arg\max_{\mathbf{z},t} \, \gamma(\mathbf{z},t).
\end{equation}
The acquisition function is defined as the Euclidean norm of the individual contributions associated with each coefficient:
\begin{equation}
\gamma(\mathbf{z},t) = \left( \sum_i \gamma_i^2(\mathbf{z},t) \right)^{1/2}.
\end{equation}

In the present implementation, we enforce hierarchically nested datasets (Section~\ref{MFGP}), such that evaluating a configuration at fidelity level $t_{new}$ requires its evaluation at all lower fidelity levels $t=1,\dots,t_{new}$.

Following Le Gratiet~\cite{le2013multi}, each $\gamma_i(\mathbf{z},t)$ is constructed to approximate the expected reduction of the integrated mean squared error (IMSE) at fidelity level $t$. This quantity is approximated as
\begin{equation}
\mathrm{IMSE}^{t}_{i,\mathrm{red}}(\mathbf{z}) 
= \sum_{k=1}^{t} 
\sigma^2_{i,r_k}(\mathbf{z}) 
\prod_{j=k}^{t-1} \rho_{i,j}^{\,2}
\prod_{m=1}^{d} l_{i,k,m},
\label{IMSE_red}
\end{equation}
where $\sigma^2_{i,r_k}(\mathbf{z})$ denotes the posterior variance of the residual Gaussian Process at level $k$, $\rho_{i,j}$ are the inter-level scaling coefficients, and $l_{i,k,m}$ are the characteristic kernel length scales. The last term represents a volume-of-influence associated with the candidate point.

The acquisition function balances uncertainty reduction and computational cost through
\begin{equation}
\gamma_i(\mathbf{z},t) = \frac{\mathrm{IMSE}^{t}_{i,\mathrm{red}}(\mathbf{z})}{\sum^{t}_{i=1}c_t},
\end{equation}
where $c_t$ denotes the computational cost associated with fidelity level $t$.

    \item The selected configuration $\mathbf{z}_{new}$ is analysed with the data-sources corresponding to $1,...,t_{new}$. The new configuration and the corresponding wind-loads are added to the training databases $\mathcal{D}^{1},...,\mathcal{D}^{t_{new}}$. 
    \item The algorithm stops when the maximum $\text{IMSE}_{red}$ at the highest fidelity level falls below a prescribed tolerance $\epsilon=0.1$. This tolerance is defined a priori based on the desired accuracy of the surrogate model and the available computational budget.
\end{enumerate}

\begin{figure*}[htbp]
\centering
	\includegraphics[scale=0.55]{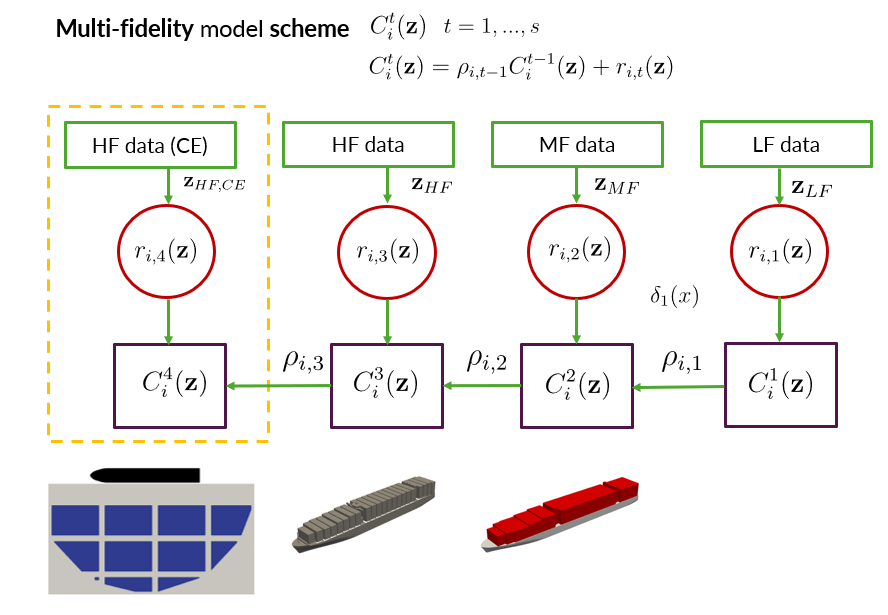}
	\caption{Schematics of the methodology adopted to train the environment-specific multi-fidelity surrogate models for wind loads. The data-sources are here denoted as low-fidelity (LF), medium fidelity (MF), high-fidelity (HF) and high-fidelity with container environment (HF-CE).}
	\label{MF_algorithm2}
\end{figure*}

The above procedure performs an automatic exploration of the design space, resulting in an enriched multi-fidelity database $\mathcal{D}$.

To account for the influence of harbour environments (Fig.~\ref{fig:loadings_envs}), additional surrogate models are trained using four fidelity levels ($s=4$). In this case, the environment-specific CFD simulations are introduced as the highest-fidelity level, while the open-sea CFD model is treated as an intermediate level. This strategy enables the reuse of previously acquired data and leverages the correlations between configurations with and without environmental effects.

\section{Results}\label{res}

\subsection{Parameter space reduction}\label{sec5p1}

\begin{figure*}[h!]
	\centering
	\subfloat[]{%
		\label{eigenvalues_sum}
		\includegraphics[width=0.39\textwidth]{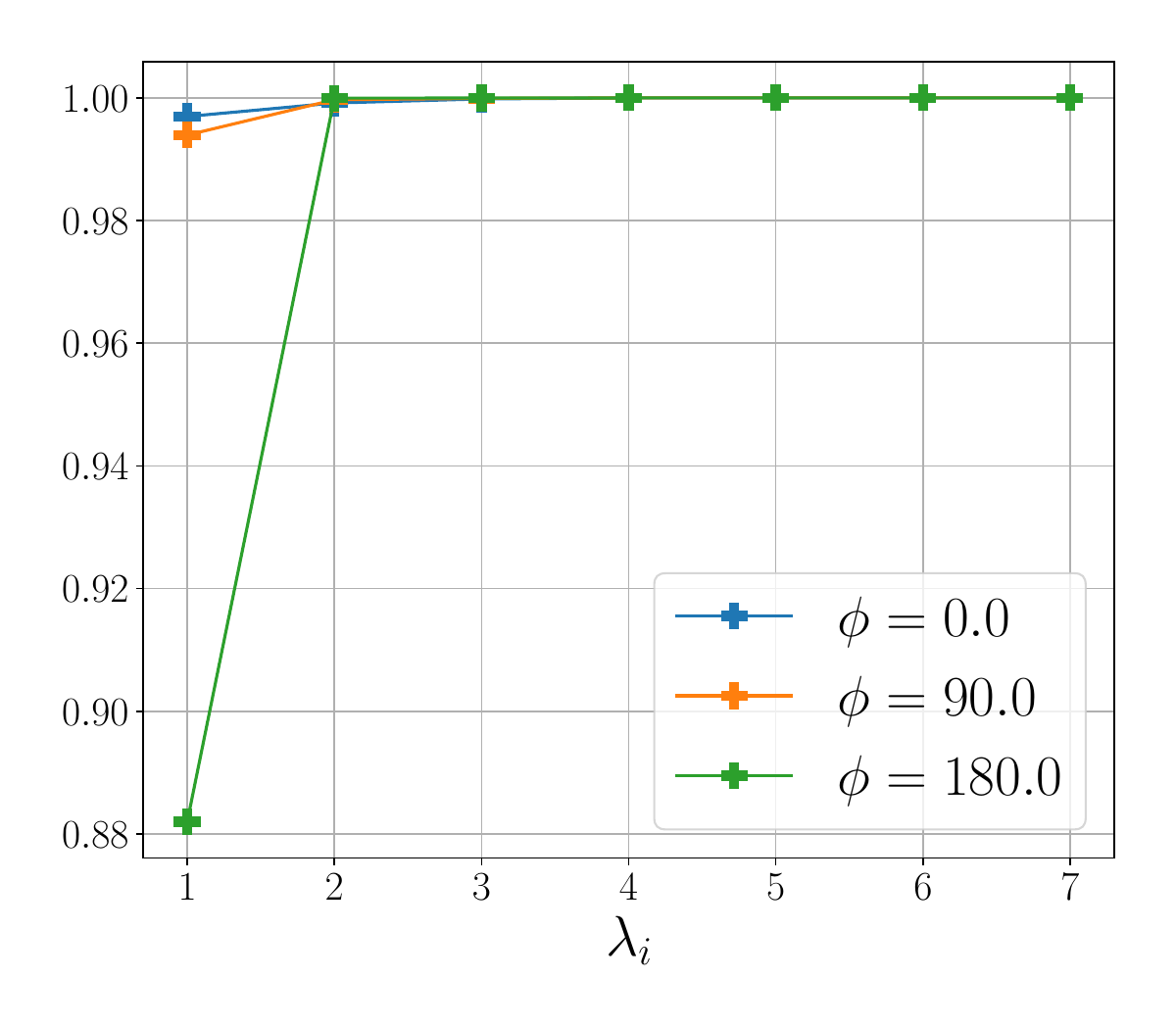}}
        \subfloat[]{
		\hfill
		\includegraphics[width=0.39\textwidth]{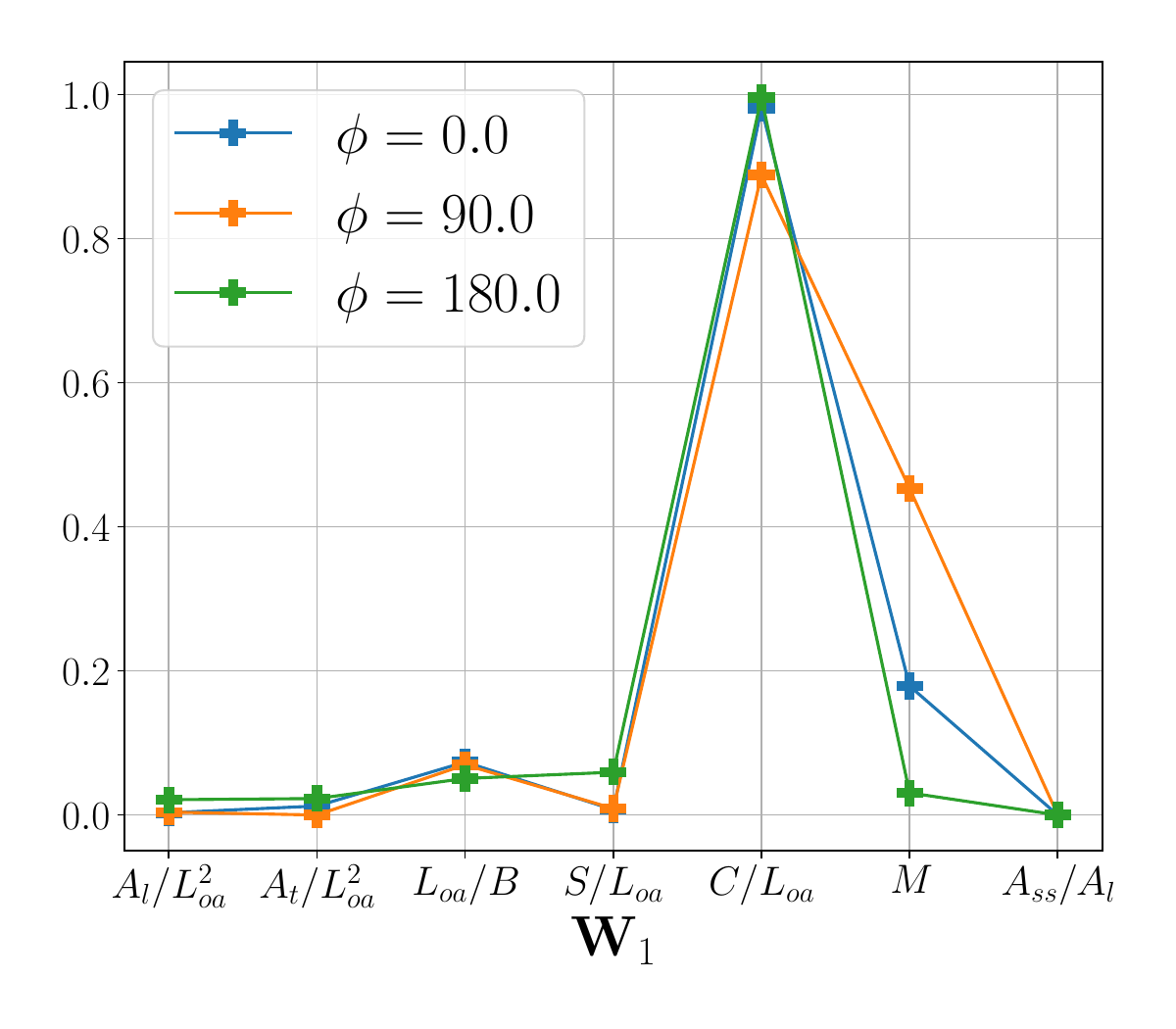}} \\
	\subfloat[]{%
		\label{tr14}
		\includegraphics[width=0.39\textwidth]{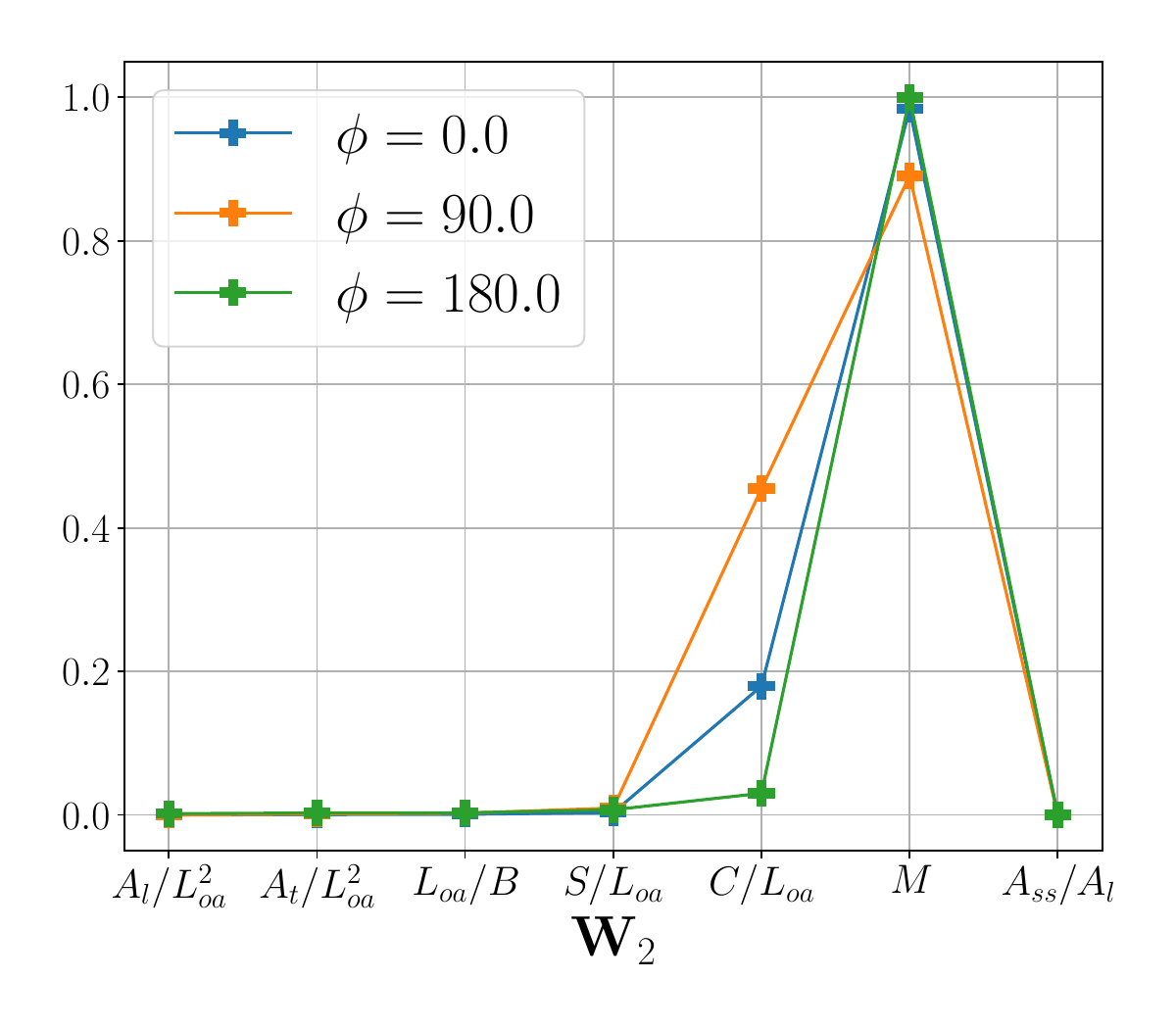}}
        	\subfloat[]{%
		\hfill     
		\includegraphics[width=0.39\textwidth]{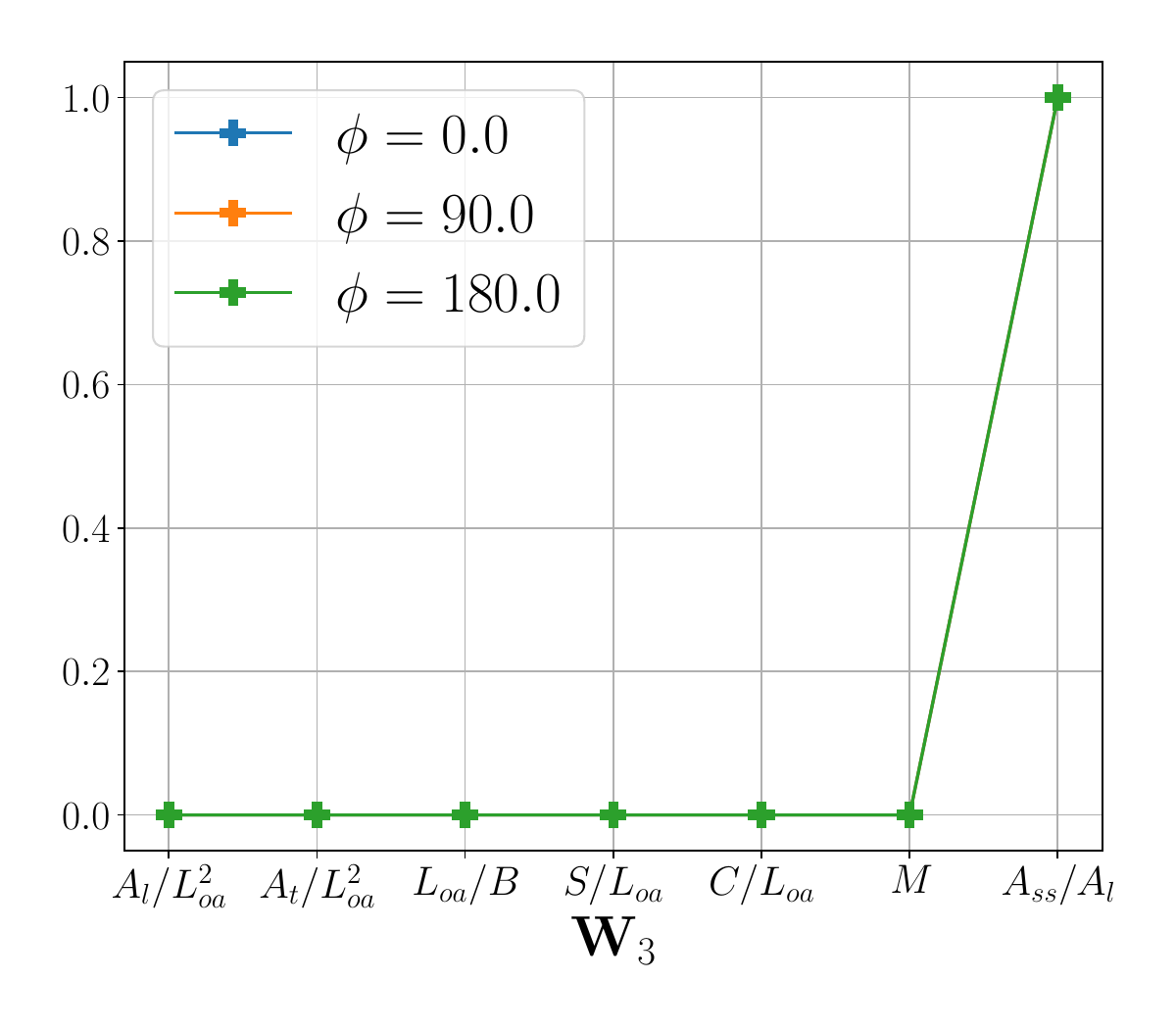}} \\

	\caption{Results of the active subspace analysis: (a) cumulative sum of the eigenvalues of $\Sigma_{AS}$, (b), (c) and (d) illustrate the components of the first three eigenvectors.}
	\label{eigenvalues}
\end{figure*}

The approach described in section \ref{active_subspace} was applied to identify the most important directions in the parameter space defined by the Isherwood correlation, given by eq.\eqref{isherwood_X}, \eqref{isherwood_Y} and \eqref{isherwood_M}. The correlation involve seven dimensionless geometrical parameters of the ship and the angle of attack, which constitute the original input vector:
\begin{equation}
\mathbf{x} = \left[\frac{A_L}{L_{OA}^2},\frac{C}{L_{OA}},\frac{A_T}{B^2}, \frac{L_{OA}}{B}, \frac{S}{L_{OA}}, M, \frac{A_{SS}}{A_L}, \phi\right]
\end{equation}
Three covariance matrices are built from the gradients of $C_X$, $C_Y$ and $C_M$, respectively. The cumulative sum of the eigenvalues of the covariance matrix built from the gradients of $C_X$ is shown in Figure \ref{eigenvalues_sum} (the results obtained for the other wind loads coefficients are very similar). The behaviour of the sum shows that only 3 components out of 7 effectively contribute to the global variance of the empirical law for the given ship. While this result is partly trivial for the parameter $L_{OA}/B$ which remains constant for the baseline ship, it nevertheless highlights that the variability of the model is governed by a reduced number of effective parameters.

The inspection of the leading eigenvectors $\mathbf{w}_1, \mathbf{w}_2$ and $\mathbf{w}_3$ of the covariance matrix provides insight into the relative importance of the input parameters, allowing the identification of those that contribute the most to the output variance.
The components of the first three eigenvectors are represented in Figure \ref{eigenvalues} for 3 different angles of attack. The components highlight that a reduced subspace can be formed from the dimensionless parameters $C/L_{OA}$, $M$ and $A_{SS}/A_L$. This ranking of the geometrical parameters seems reasonable, as it suggests that wind loads are influenced, beside the angle of attack, by the size of the super-structure, its position and the number of subchannels in the container stack. 

Based on the results of the active subspace analysis, the surrogate models developed in the present work considers 4 input parameters:
\begin{equation}
\mathbf{z} = \left[\frac{C}{L_{OA}}, M, \frac{A_{SS}}{A_L}, \phi\right].
\end{equation}

\subsection{Correlation between fidelity levels}
\begin{figure*}[!htbp]
	\centering
	\subfloat[$C_X$]{%
		\label{Cx_fullLoad}
		\includegraphics[width=0.35\textwidth]{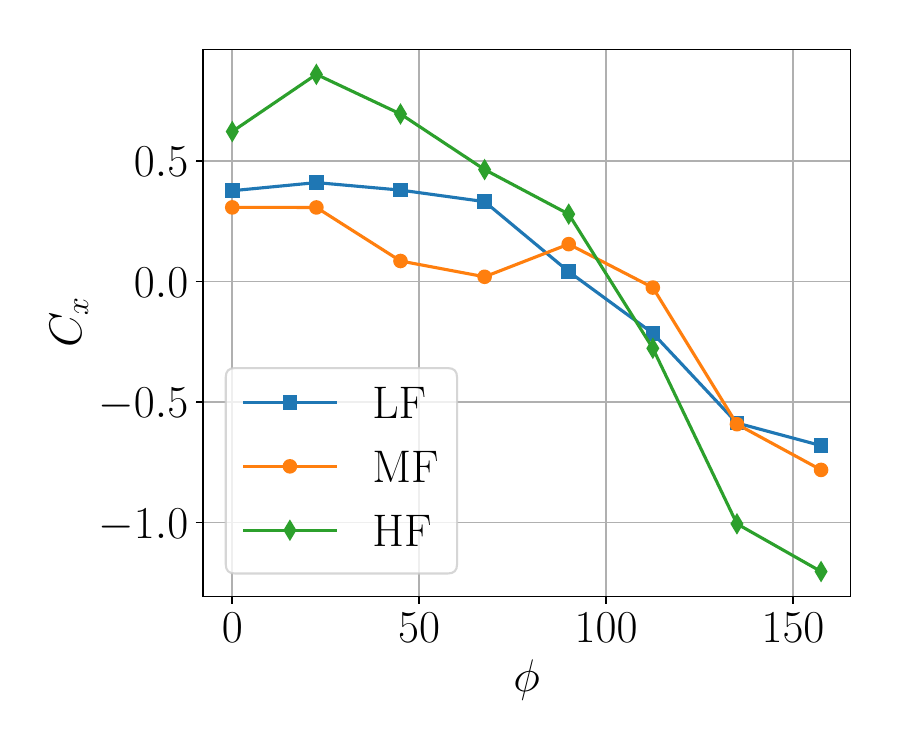}
		\hfill
		\includegraphics[width=0.35\textwidth]{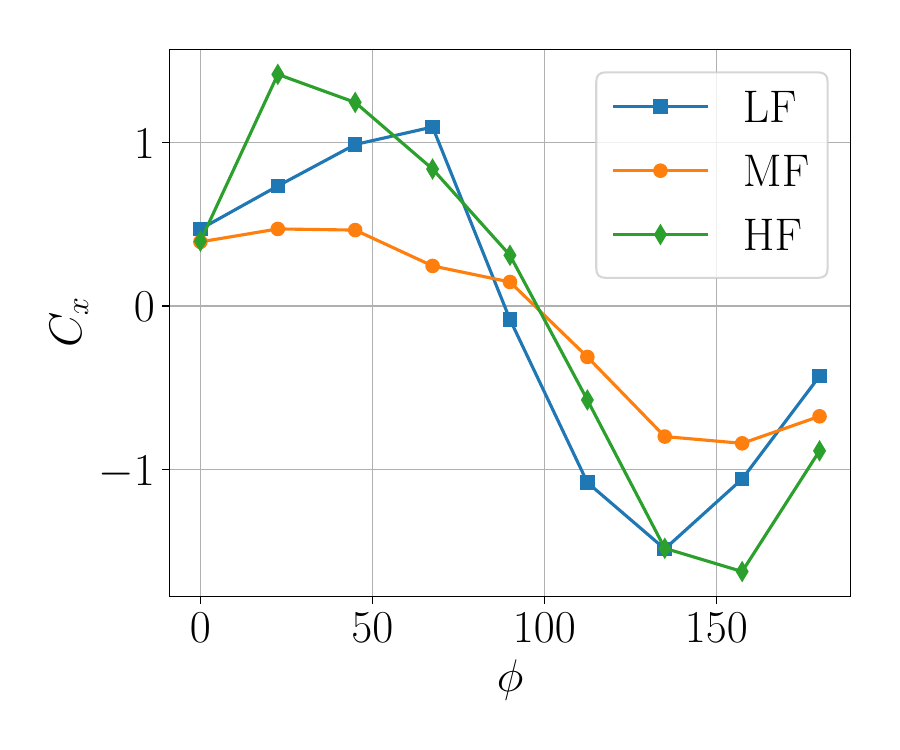}} \\
	\subfloat[$C_Y$]{%
		\label{lateral_comp}
		\includegraphics[width=0.35\textwidth]{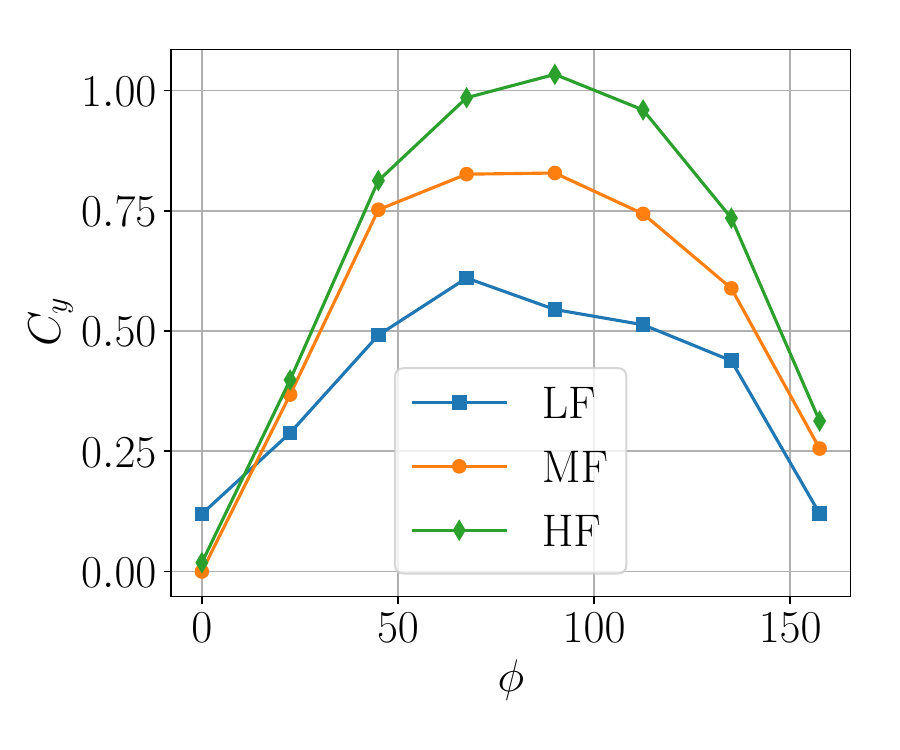}
		\hfill      
		\includegraphics[width=0.35\textwidth]{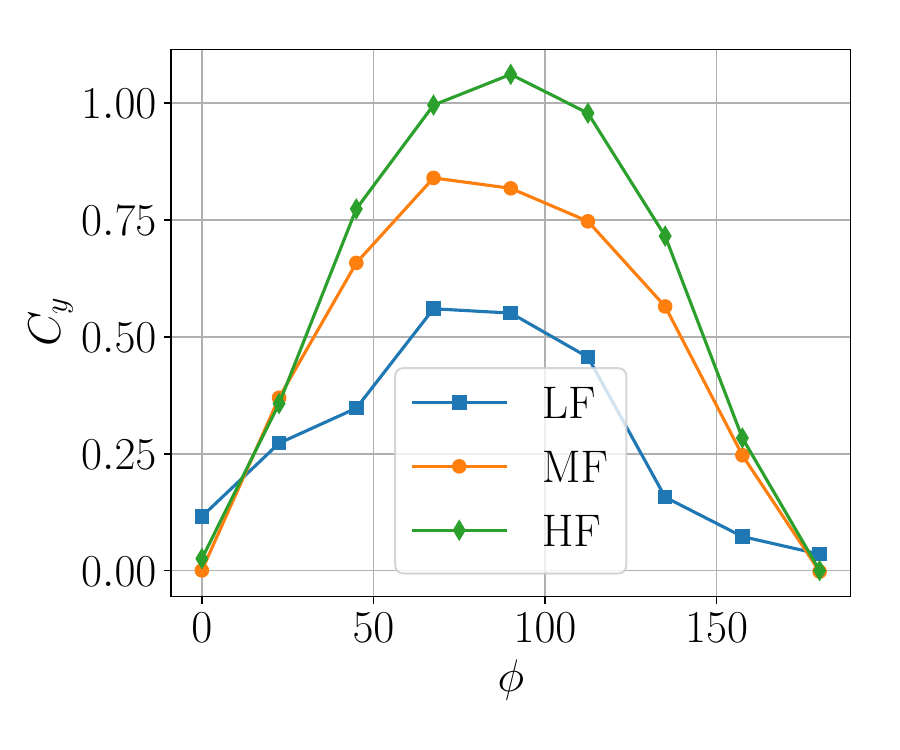}} \\

	\subfloat[$C_M$]{%
		\label{tr14}
		\includegraphics[width=0.35\textwidth]{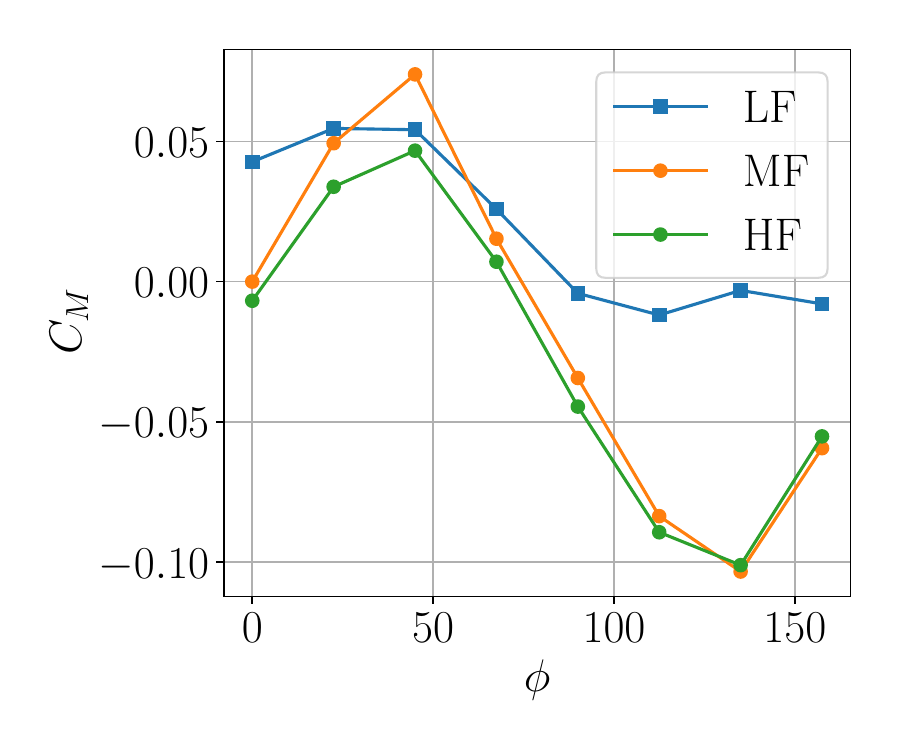}
		\hfill      
		\includegraphics[width=0.35\textwidth]{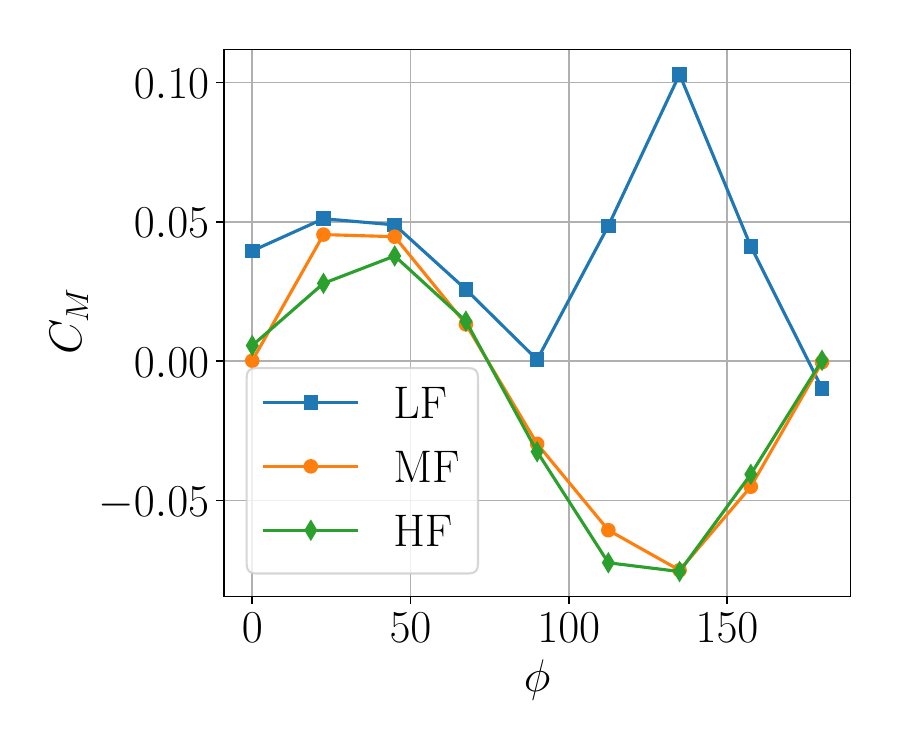}} \\
    
	\caption{Wind load coefficients predicted by the empirical correlation (LF), the simplified CFD (MF) and the detailed CFD (HF) for the full load configuration (left) and the intermediate load configuration (right).}
	\label{comparison_wl}
\end{figure*}

The initial design of experiments considers the full stack and the intermediate stack configurations indicated in Figure \ref{fig:loadings_envs}. The wind load coefficients computed with different data-sources, for angles of attack between $0^o$ and $180^o$ are indicated in Figure \ref{comparison_wl}. The comparison highlights the significant correlation between the data-sources, justifying the application of a multi-fidelity approach. Specifically, LF data are strongly correlated with the MF and HF data for $C_X$ and $C_Y$, especially for the full load configuration. A poor correlation is observed instead between the LF data for $C_M$ and the corresponding MF and HF values for $\phi>90^o$. This lack of correlation in this region of the input space can potentially reduce the performance of the multi-fidelity surrogate model. In principle, this region should be excluded from the LF dataset and from the sequential sampling at this fidelity level. However, it is retained here to avoid violating the assumption of nested training datasets (see section \ref{MFGP}). The poor quantitative agreement between the empirical correlation and the CFD models strengthens the limits of empirical models and the importance of numerical simulations to accurately predict wind loads. 

A good correlation between MF and HF data can be appreciated in Figure \ref{comparison_wl} in both the loading configurations. Figure \ref{Cx_fullLoad} also shows that the medium fidelity data-source (simplified CFD) tends to underpredict the longitudinal force coefficient with respect to the high-fidelity data-source (detailed CFD). This is a consequence of neglecting the gaps between the containers, which leads to a smooth loading surface guiding the flow. Also, the lateral force coefficient depicted in Figure \ref{lateral_comp} is slightly underpredicted by the medium fidelity data-source, suggesting that the porous model implemented in the simplified CFD underestimate the effective resistance. 

\subsection{Exploration of the design space throughout the training}

The training algorithm described in Section~\ref{training} is applied to construct the multi-fidelity surrogate model for wind-load prediction in open-sea conditions. In this case, the database consists of three fidelity levels (LF, MF and HF), as schematized in Fig.~\ref{MF_algorithm}.

\begin{figure*}[!hbtp] 
	\centering
	\includegraphics[scale=0.4]{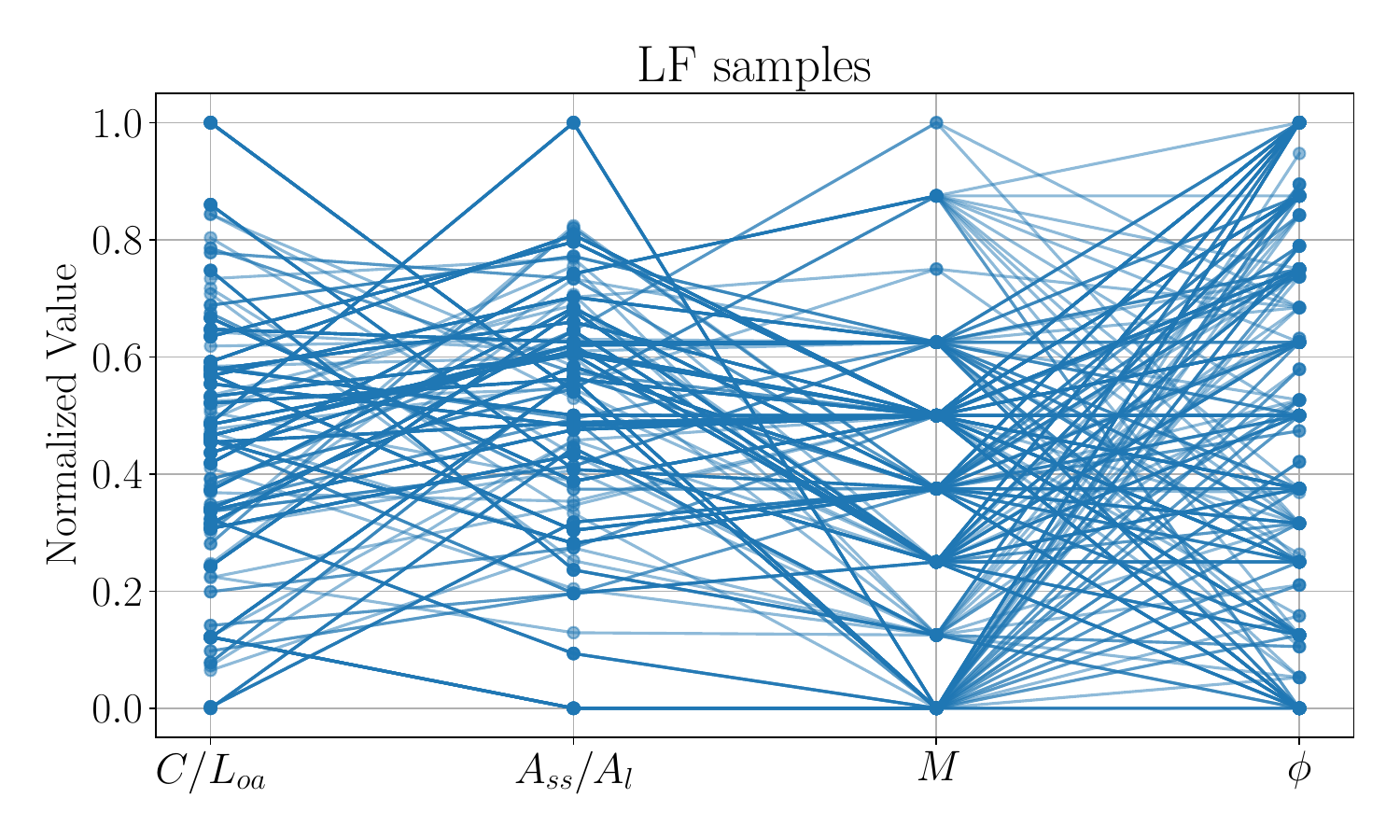}
    	\includegraphics[scale=0.4]{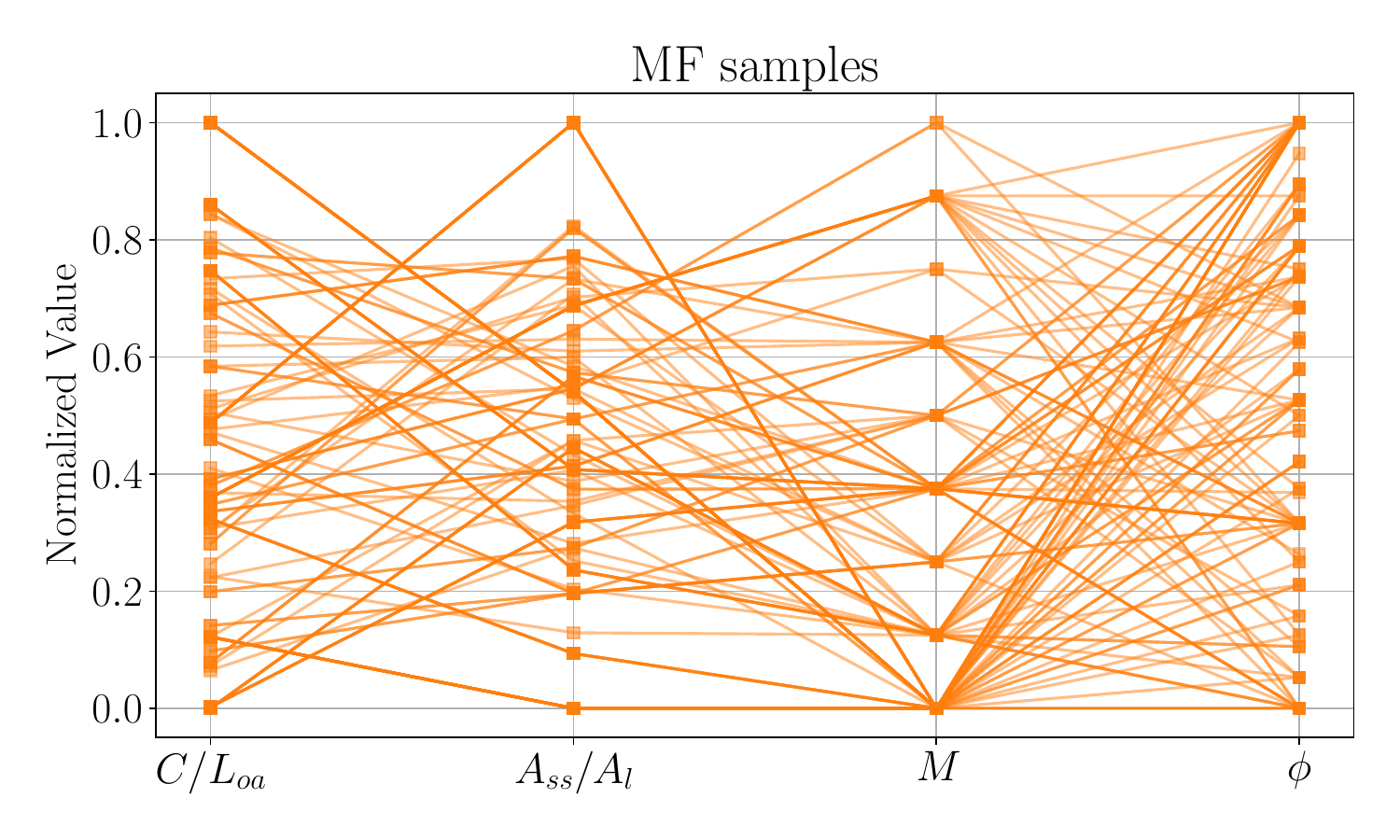}
        	\includegraphics[scale=0.4]{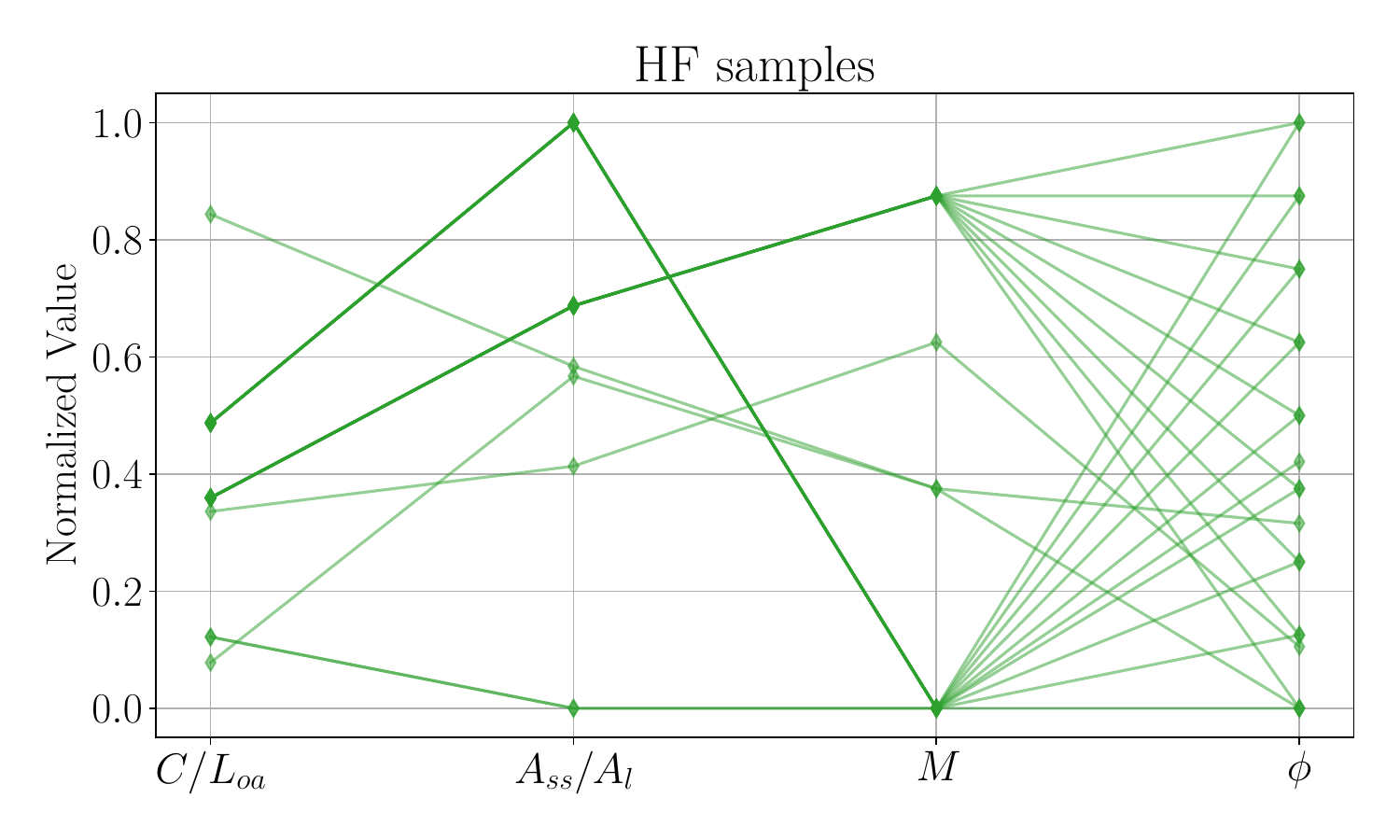}
	\caption{Sample configurations analysed with the data-sources of low, medium and high fidelity throughout the sequential learning process. Each polyline represents one sampled configuration in the reduced input space. The comparison highlights the broader coverage of the MF dataset and the targeted enrichment of the HF samples through the acquisition strategy.}
	\label{samples}
\end{figure*}

The algorithm performs a sequential exploration of the parameter space over 500 iterations. The sample configurations selected at the different fidelity levels are shown in Fig.~\ref{samples}. Each polyline represents a configuration in the reduced input space, enabling a compact visualization of the sampling distribution across all parameters. The figure highlights the distinct roles of the different fidelity levels: the medium-fidelity (MF) data provide a broad coverage of the admissible parameter space, while the high-fidelity (HF) samples are more sparsely distributed and concentrated in selected regions, reflecting the targeted enrichment driven by the acquisition strategy.

The algorithm ensures a uniform coverage of the angle of attack $\phi$ over the interval $[0^\circ,180^\circ]$. In contrast, the geometric parameters are not continuously sampled but restricted to discrete admissible values associated with feasible loading configurations. As a result, the sampling does not densely populate the entire parameter space. In particular, only nine discrete values of the number of container groups ($M$) are considered.

\begin{figure*}[!hbtp] 
	\centering
	\includegraphics[scale=0.45]{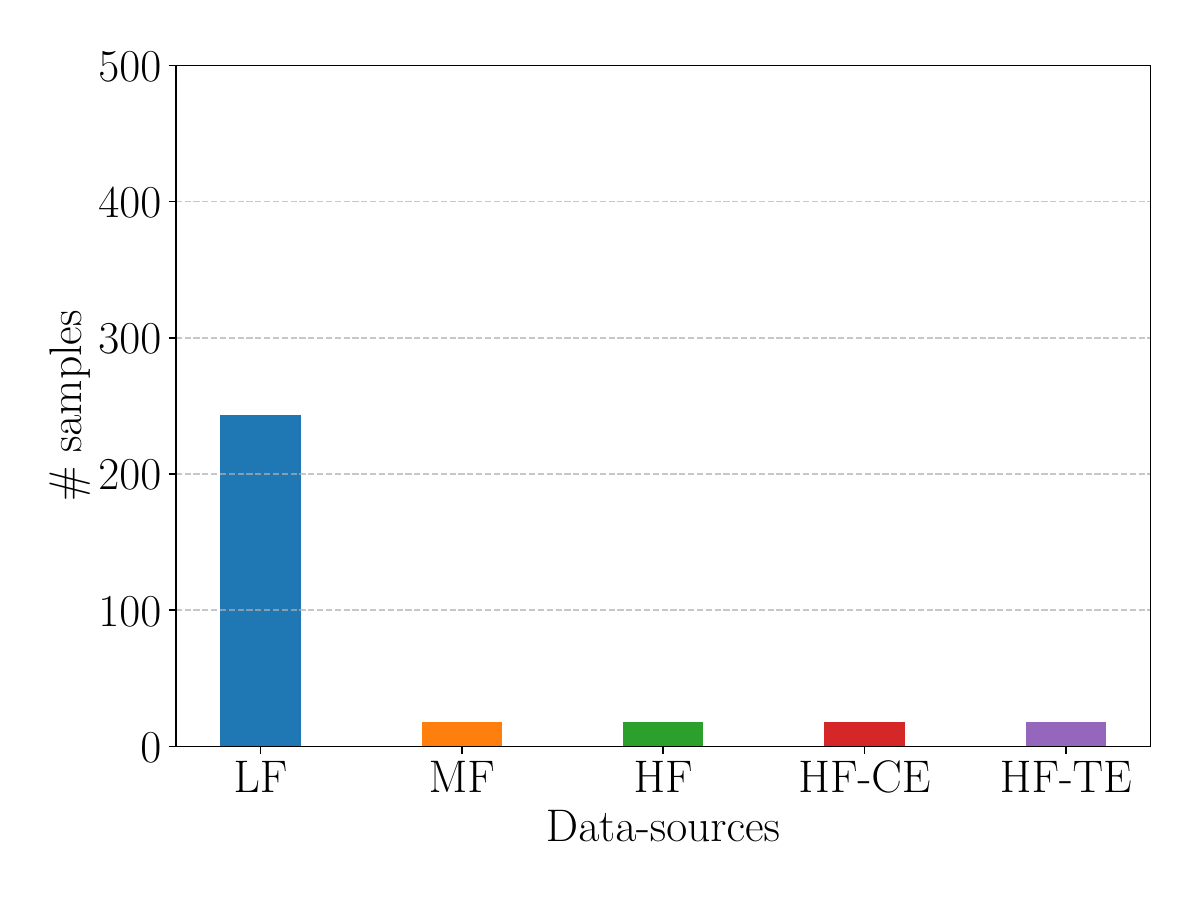}
	\includegraphics[scale=0.45]{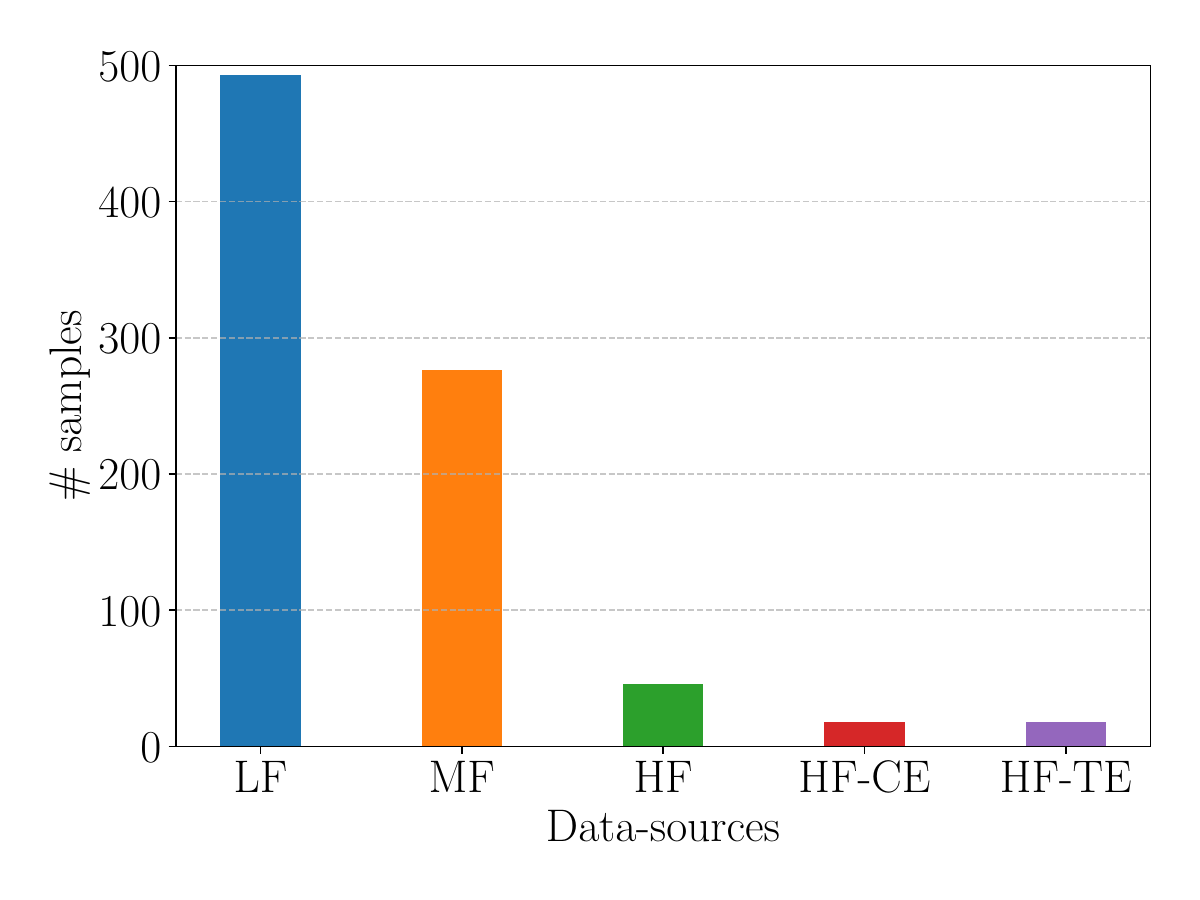}
	\caption{Data distribution at the different fidelity levels at the initialization (top) and at the end of the training process (bottom). Data of lowfidelity (empirical correlation, LF), medium fidelity (simplified CFD, MF), high fidelity (detailed CFD in open sea, HF). The data related to the detailed CFD in container and tanks environments are denoted as HF-CE and HF-TE, respectively.}
	\label{data_dist}
\end{figure*}

Figure~\ref{data_dist} shows the distribution of the training data at the beginning and at the end of the sequential learning process. A total of 490 configurations are evaluated using the low-fidelity empirical model, while only 22 configurations are computed at the highest fidelity level. This imbalance illustrates the cost-aware nature of the multi-fidelity strategy, which leverages inexpensive data sources to explore the design space while reserving high-fidelity evaluations for the most informative configurations.

\begin{figure*}[t!] 
	\centering
	\includegraphics[scale=0.75]{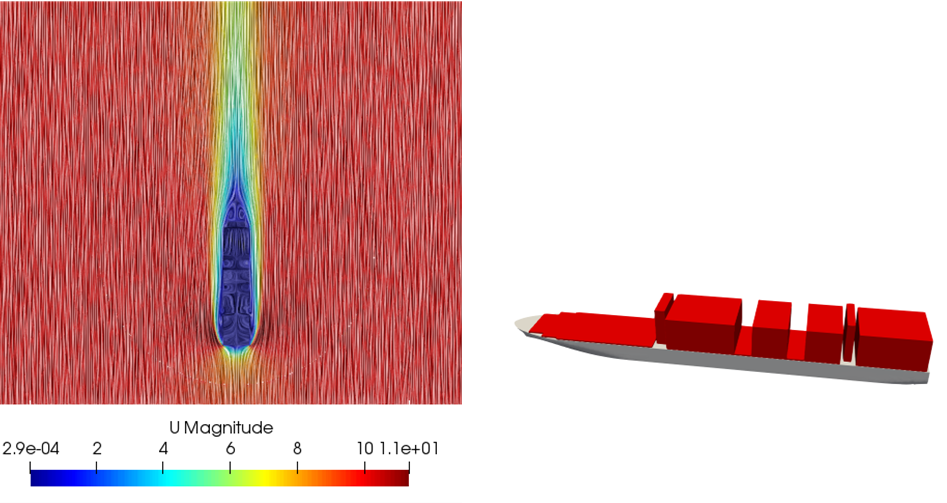}

	\caption{Streamwise velocity contours obtained with sample configurations explored by the training algorithm with the medium fidelity data-source (simplified CFD). The wind velocity $U=10 m/s$ is directed from bottom to top, the angle of attack is $\phi=180^o$. }
	\label{conf_cfd1}
\end{figure*}

\begin{figure*}[t!] 
	\centering
	\includegraphics[scale=0.75]{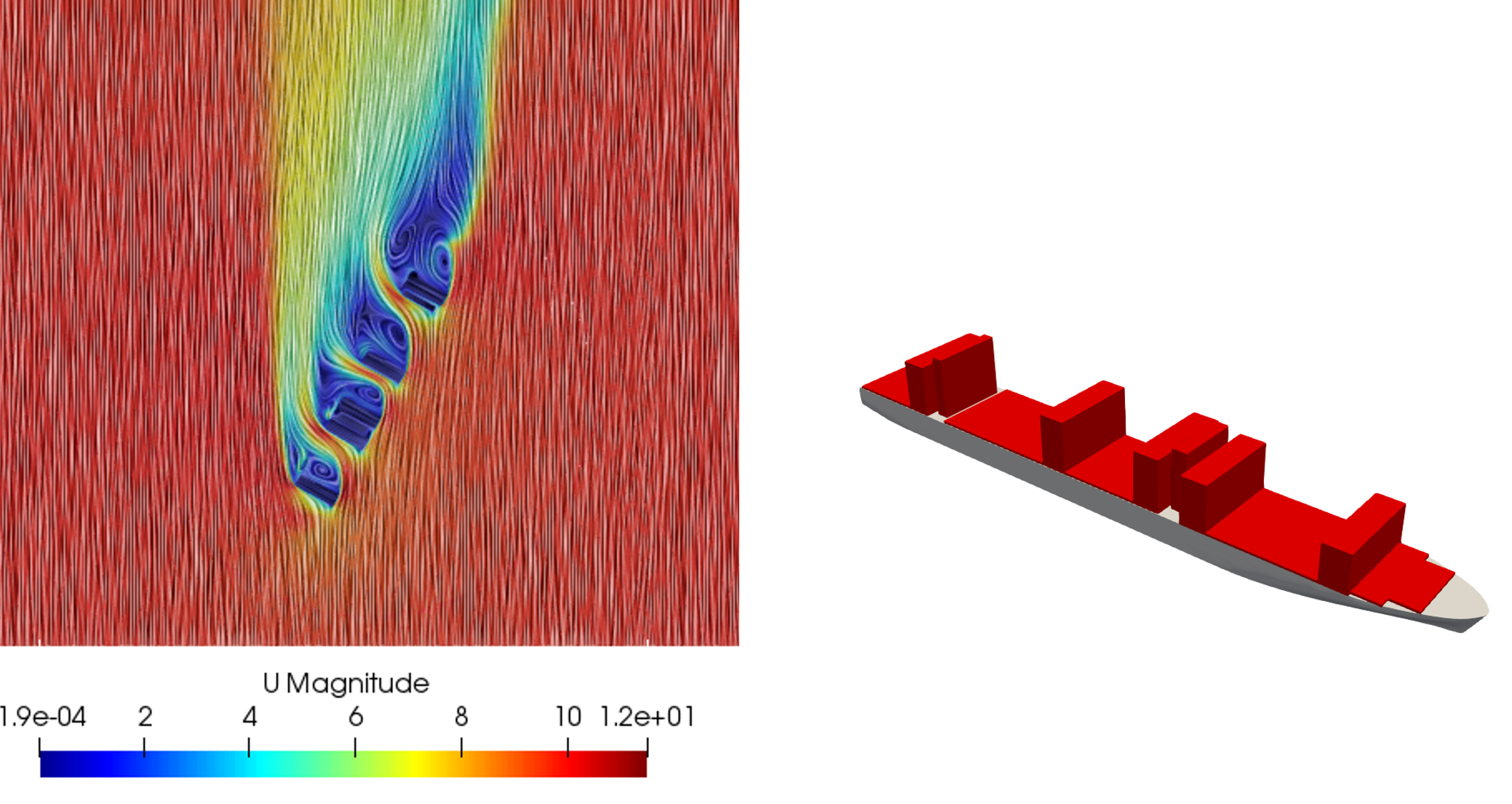}

	\caption{Streamwise velocity contours obtained with sample configurations explored by the training algorithm with the medium fidelity data-source (simplified CFD). The wind velocity $U=10 m/s$ is directed from bottom to top, the angle of attack is $\phi=23^o$. }
	\label{conf_cfd2}
\end{figure*}

\begin{figure*}[t!] 
	\centering
	\includegraphics[scale=0.75]{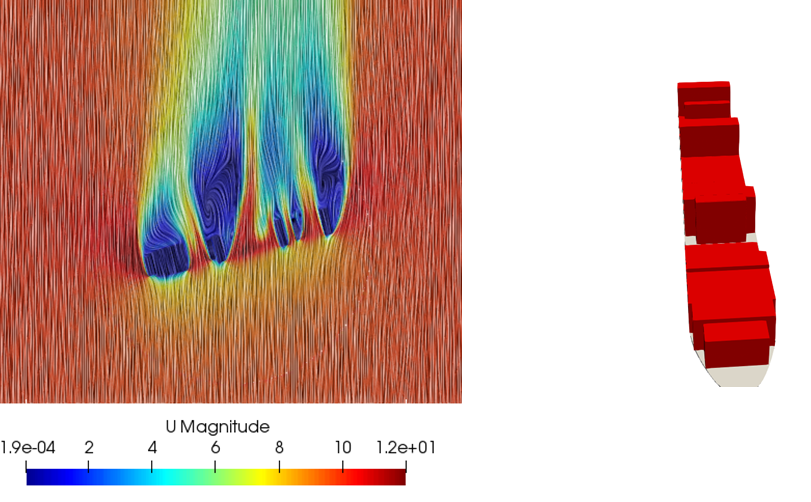}

	\caption{Streamwise velocity contours obtained with sample configurations explored by the training algorithm with the medium fidelity data-source (simplified CFD). The wind velocity $U=10 m/s$ is directed from bottom to top, the angle of attack is $\phi=75^o$. }
	\label{conf_cfd3}
\end{figure*}

\begin{figure*}[t!] 
	\centering
	\includegraphics[scale=0.75]{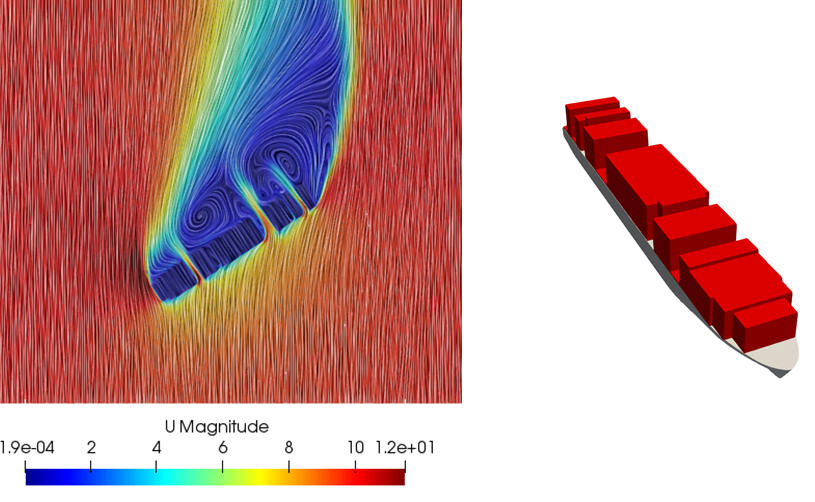}

	\caption{Streamwise velocity contours obtained with sample configurations explored by the training algorithm with the medium fidelity data-source (simplified CFD). The wind velocity $U=10 m/s$ is directed from bottom to top, the angle of attack is $\phi=56^o$. }
	\label{conf_cfd4}
\end{figure*}

\begin{figure*}[t!] 
	\centering
	\includegraphics[scale=0.75]{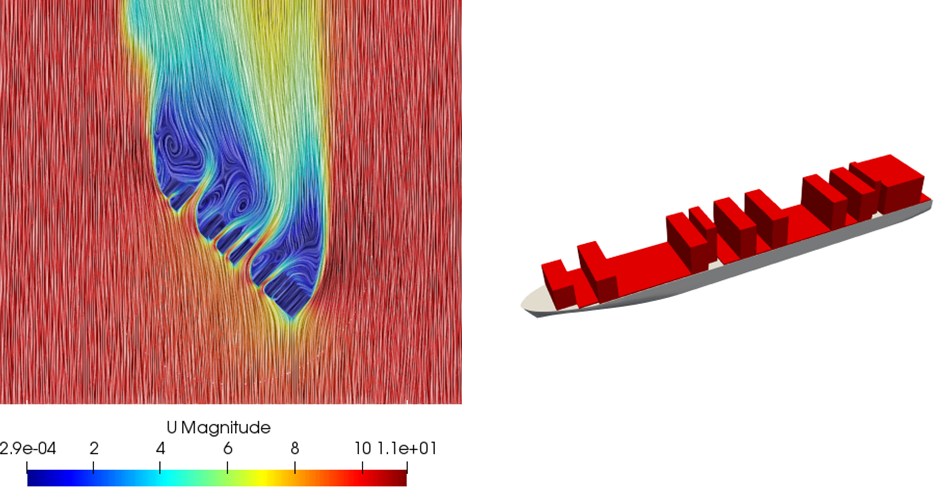}

	\caption{Streamwise velocity contours obtained with sample configurations explored by the training algorithm with the medium fidelity data-source (simplified CFD). The wind velocity $U=10 m/s$ is directed from bottom to top, the angle of attack is $\phi=132^o$. }
	\label{conf_cfd5}
\end{figure*}

An intermediate design space exploration is performed using the medium-fidelity data source (simplified CFD), resulting in approximately 250 MF samples. Examples of the corresponding flow fields are shown in Figs.~\ref{conf_cfd1}--\ref{conf_cfd5}, where contours of the streamwise velocity highlight the variability induced by both the loading configuration (through the extent of the porous regions) and the angle of attack.

\subsection{Performance in multiple and single fidelities (open sea conditions)}

\begin{figure*}[htbp]
    \centering
    \subfloat[][]{\includegraphics[width=0.43\textwidth, trim=3.5cm 5cm 3.5cm 3cm, clip]{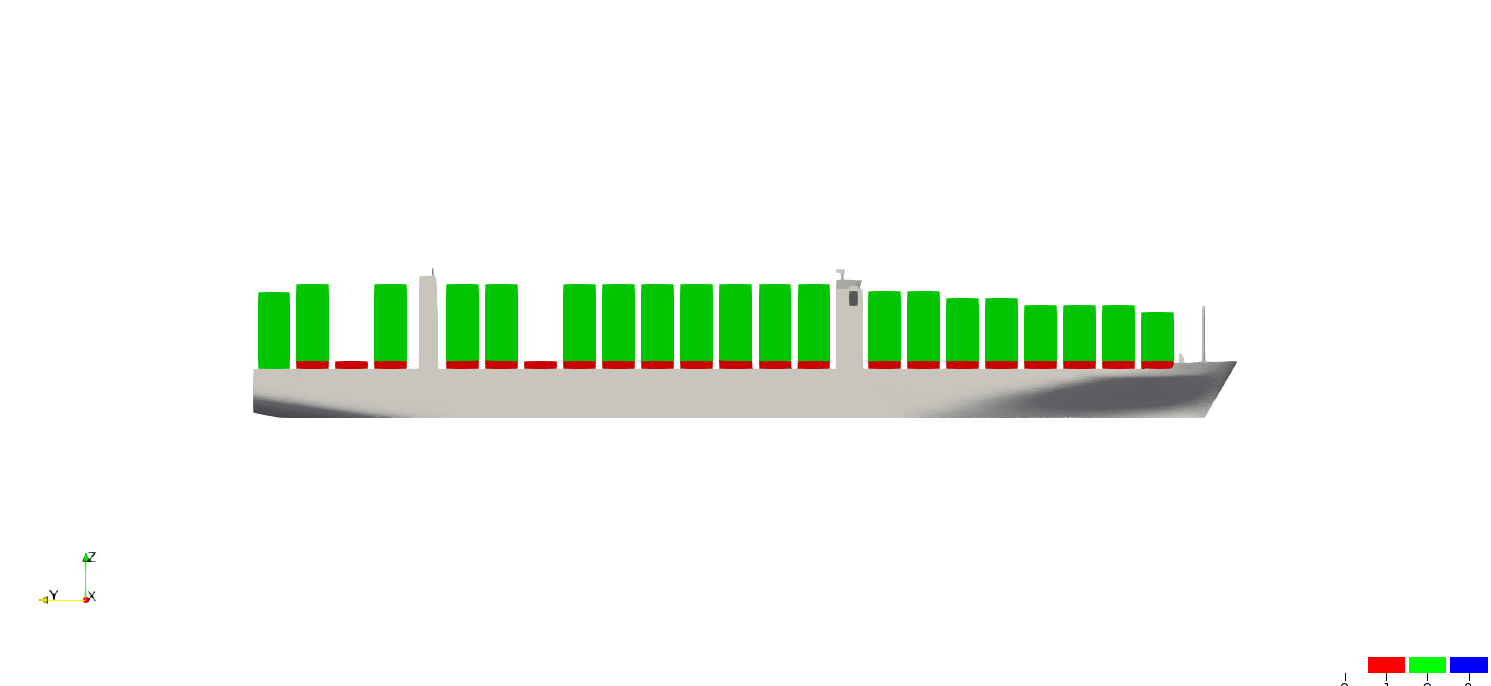}}\!
    \subfloat[][]{\includegraphics[width=0.43\textwidth, trim=3.5cm 5cm 3.5cm 3cm, clip]{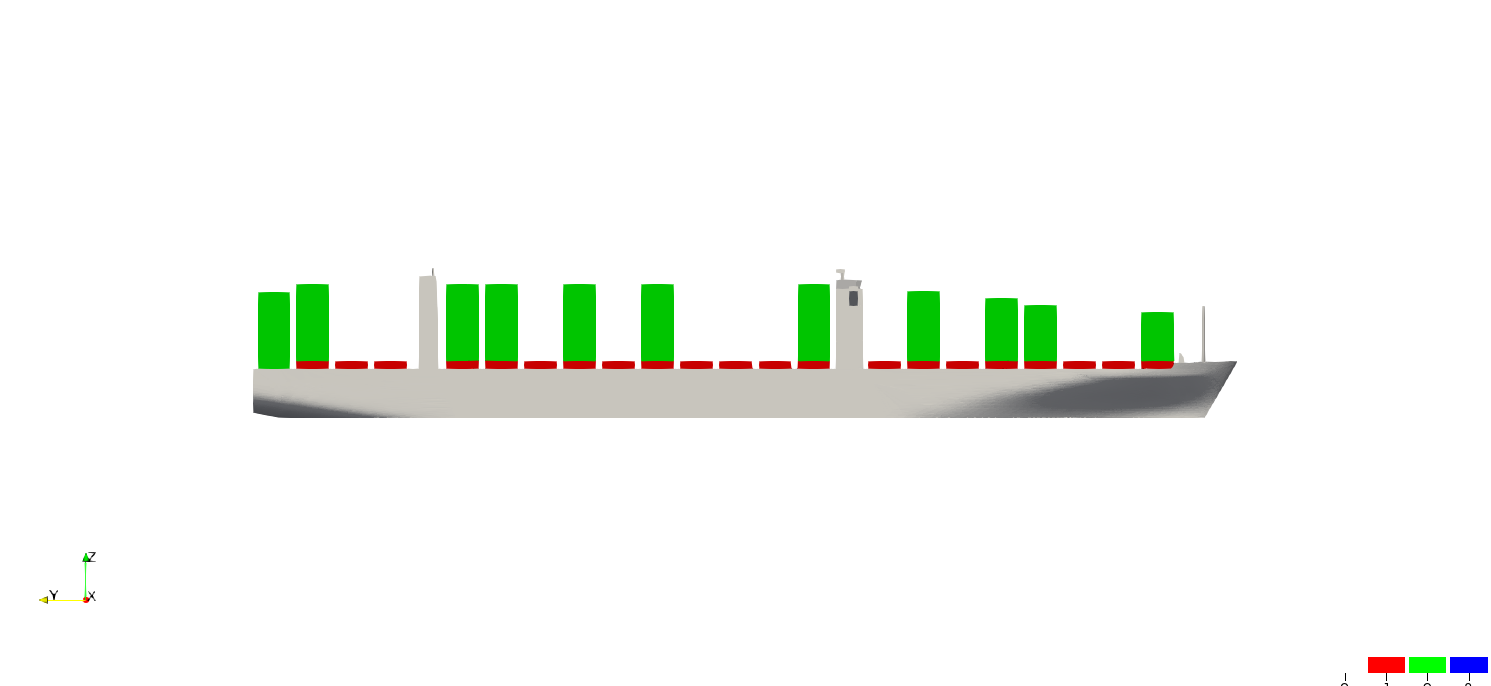}}\\
      \subfloat[][]{\includegraphics[width=0.43\textwidth, trim=3.5cm 5cm 3.5cm 3cm, clip]{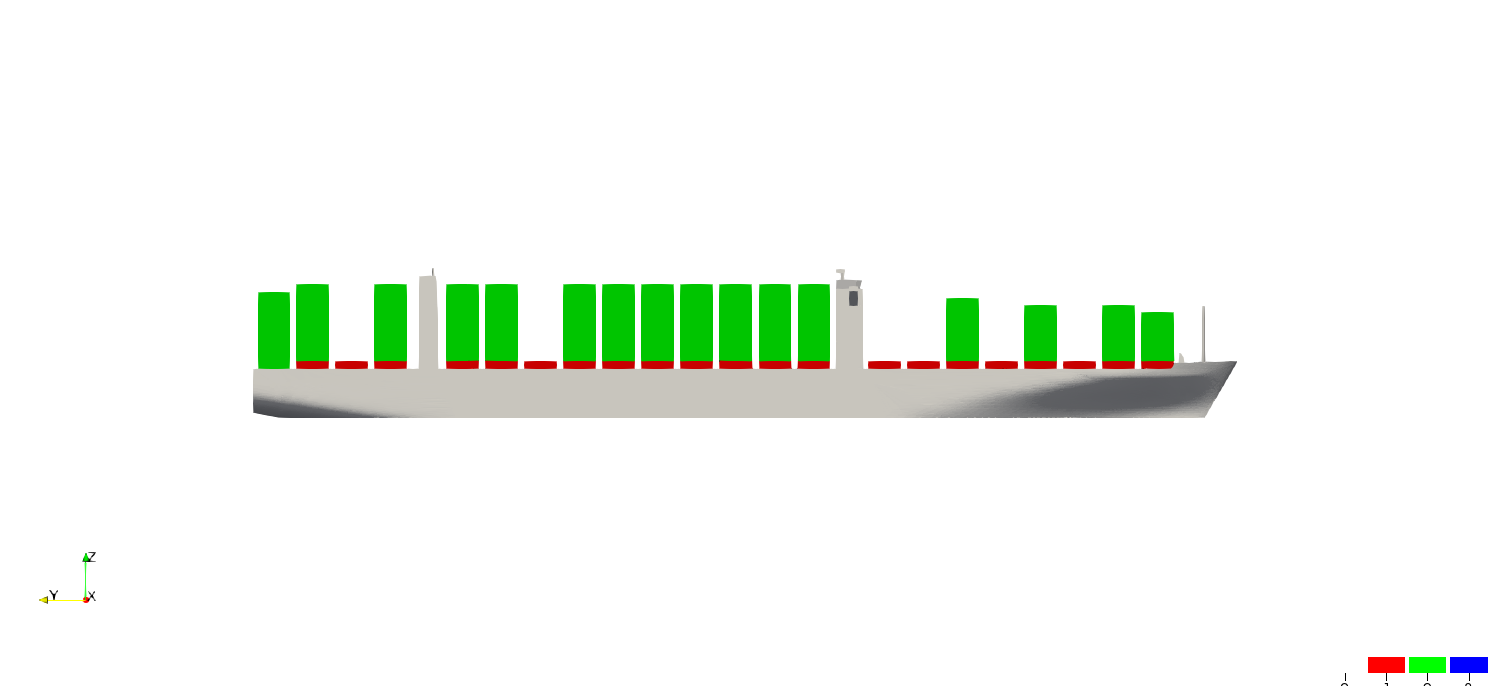}}\!
    \subfloat[][]{\includegraphics[width=0.43\textwidth, trim=3.5cm 5cm 3.5cm 3cm, clip]{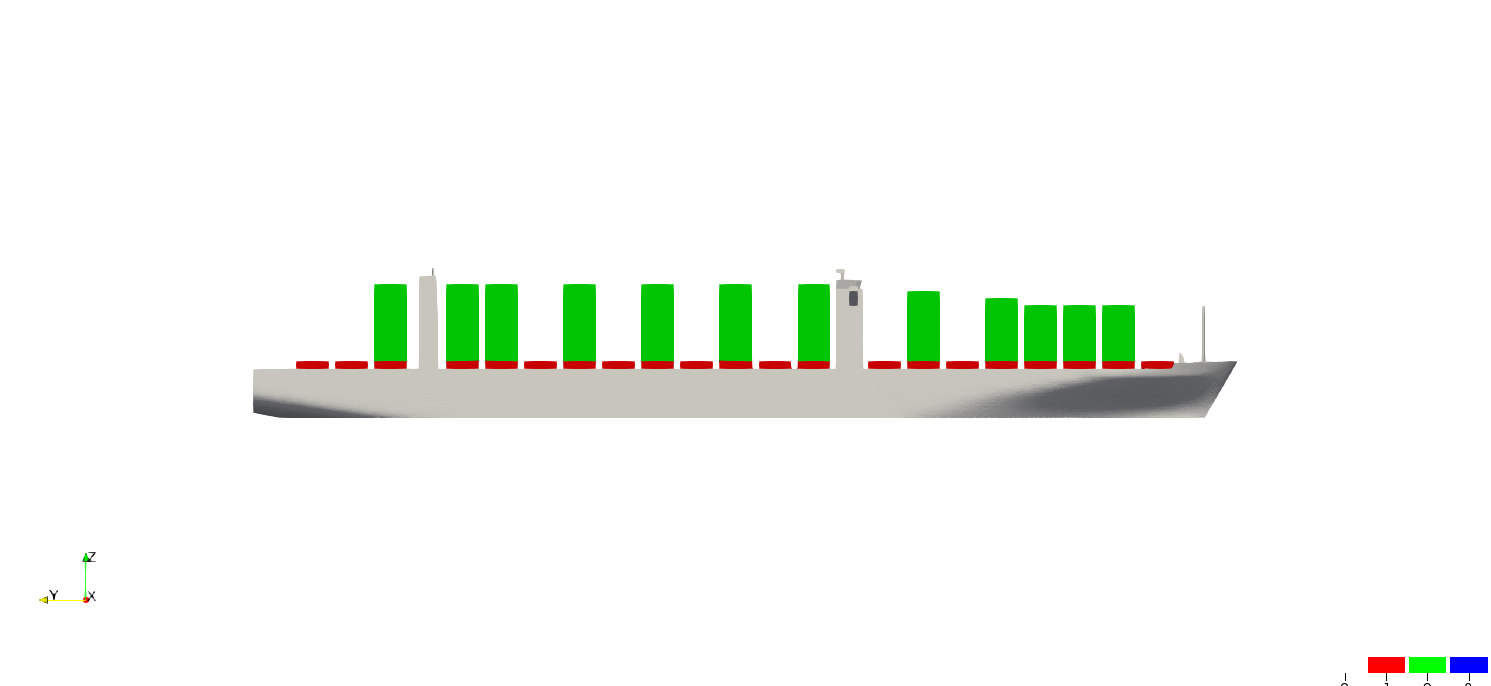}}\\
        \subfloat[][]{\includegraphics[width=0.43\textwidth, trim=3.5cm 5cm 3.5cm 3cm, clip]{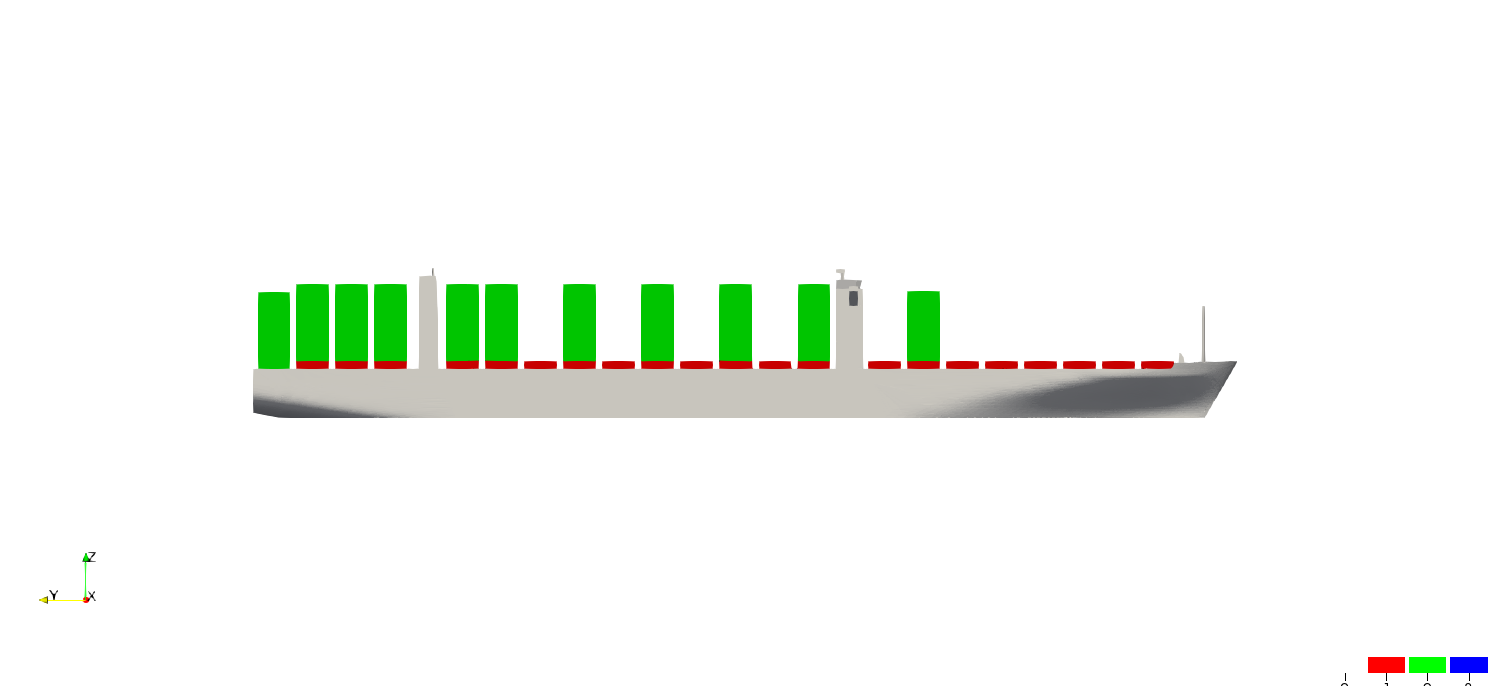}}\!
    \subfloat[][]{\includegraphics[width=0.43\textwidth, trim=3.5cm 5cm 3.5cm 3cm, clip]{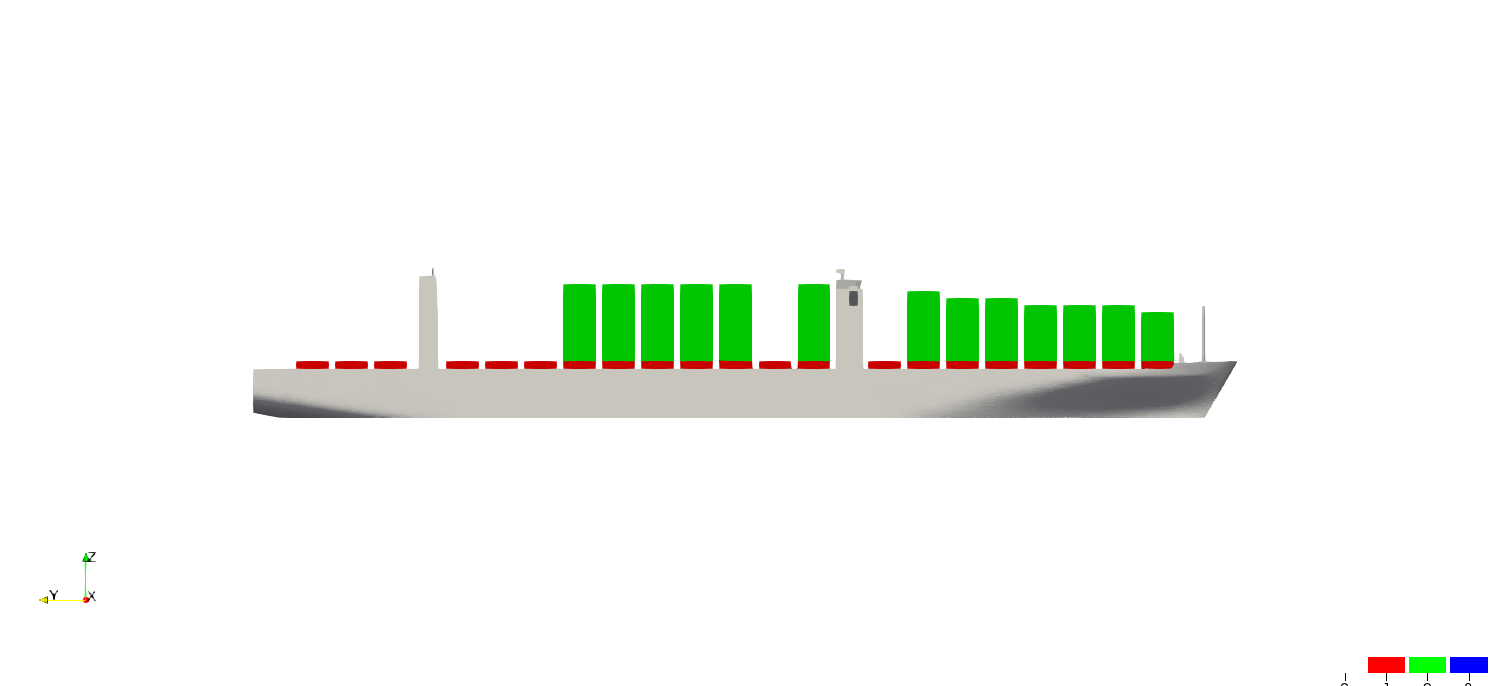}}\\
        \subfloat[][]{\includegraphics[width=0.43\textwidth, trim=3.5cm 5cm 3.5cm 3cm, clip]{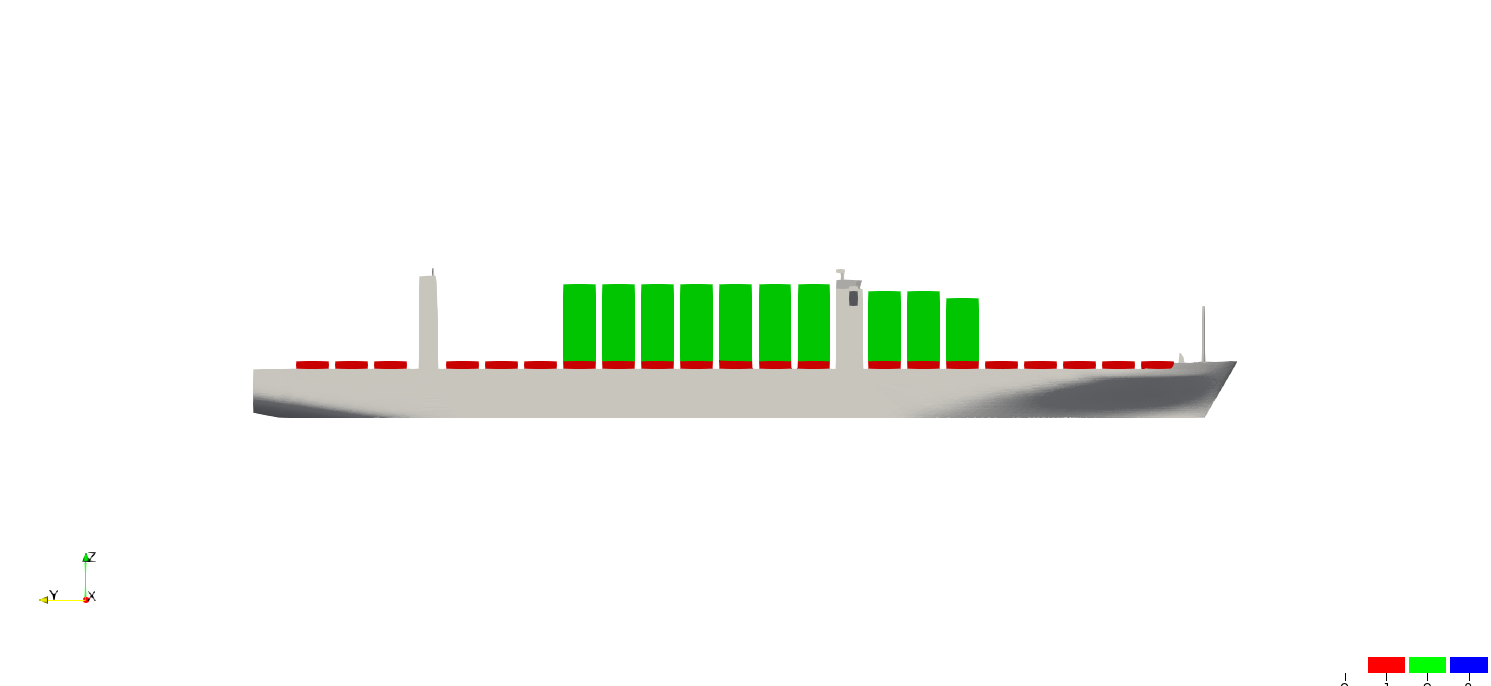}}\!
    \subfloat[][]{\includegraphics[width=0.43\textwidth, trim=3.5cm 5cm 3.5cm 3cm, clip]{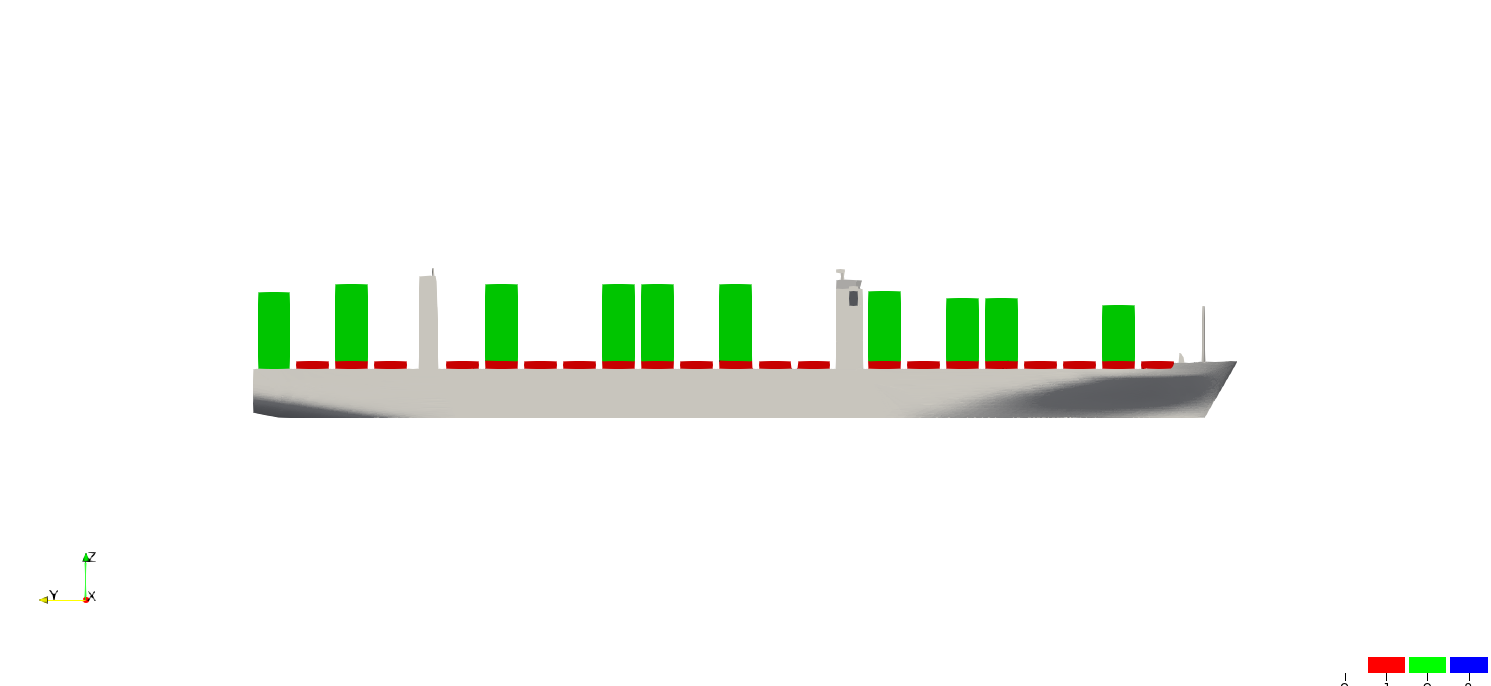}}\\
    \caption{Loading configurations considered for the validation of the surrogate models.}
    \label{validation_confs}
\end{figure*}

The final dataset (bottom of Fig.~\ref{data_dist}) is used for all subsequent analyses and model comparisons. To assess the benefit of the multi-fidelity approach, the predictive performance of the multi-fidelity surrogate model is compared with a single-fidelity counterpart, i.e. a Gaussian Process trained exclusively on high-fidelity (HF) data (detailed CFD in open-sea conditions).

\begin{figure*}[htbp]
\subfloat[$C_X$]{%
  \includegraphics[width=0.4\textwidth]{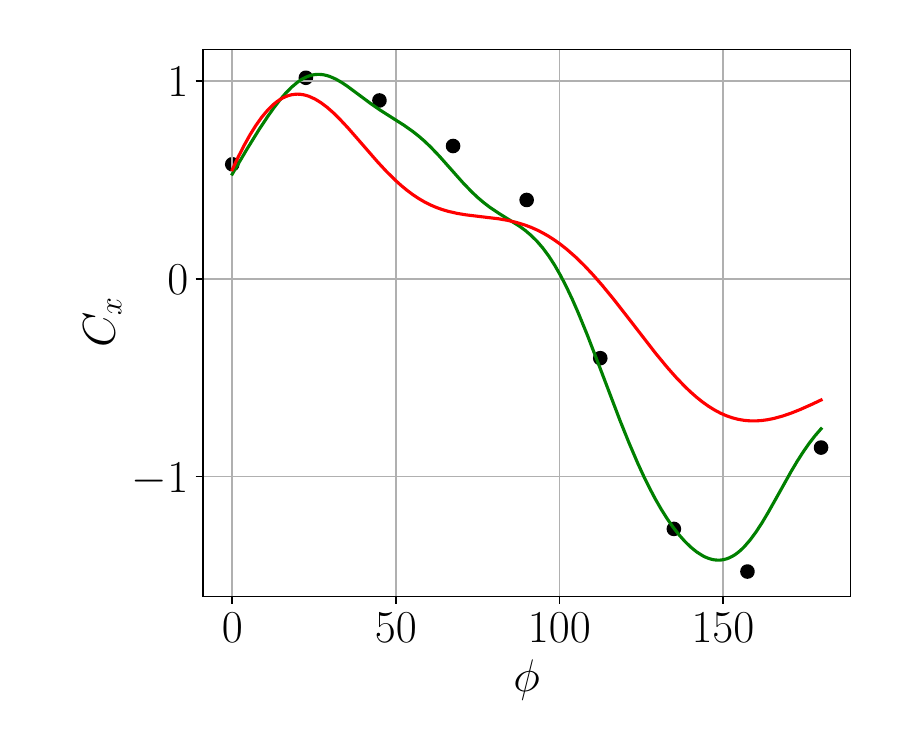}}
\hspace{0.05\textwidth}
\subfloat[$C_Y$]{%
  \includegraphics[width=0.4\textwidth]{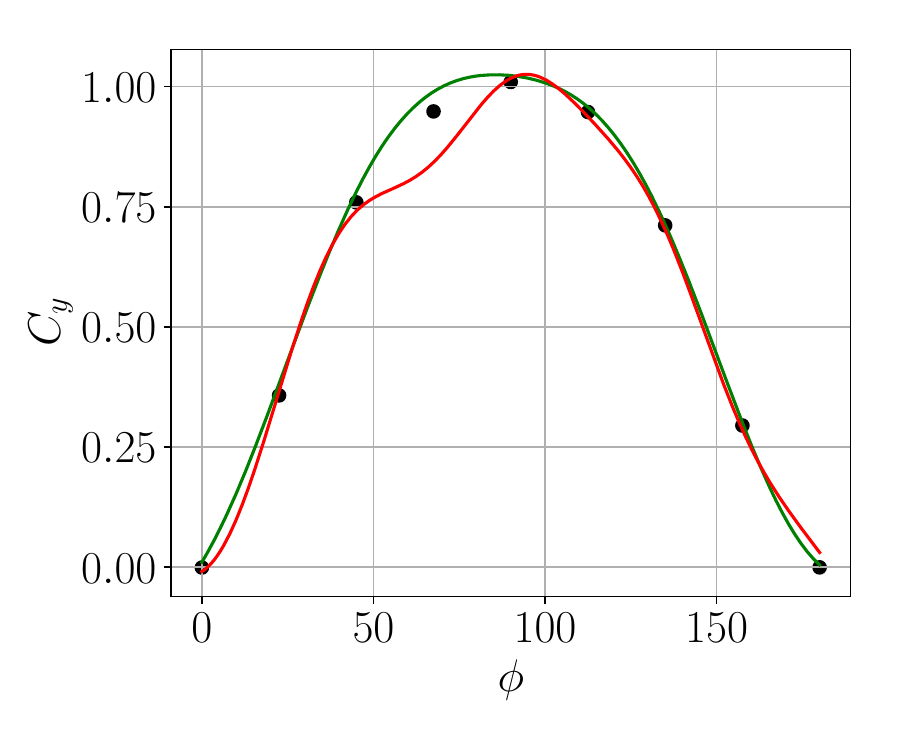}} \\

\subfloat[$C_M$]{%
  \includegraphics[width=0.4\textwidth]{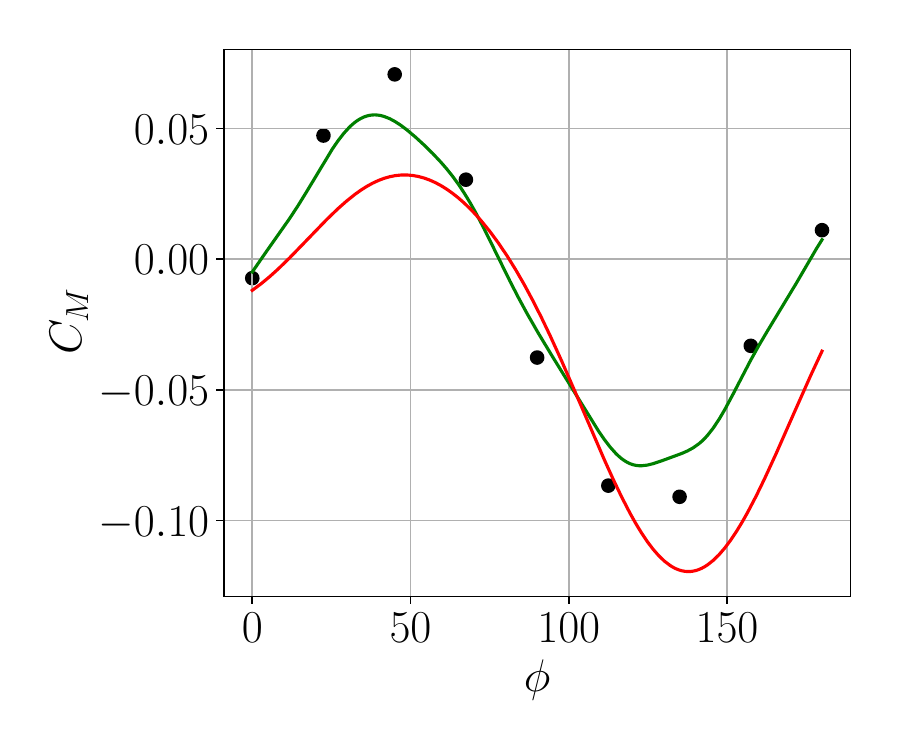}}
\hspace{0.05\textwidth}
\subfloat[Validation loading configuration]{%
\label{tr143}
  \includegraphics[width=0.35\textwidth]{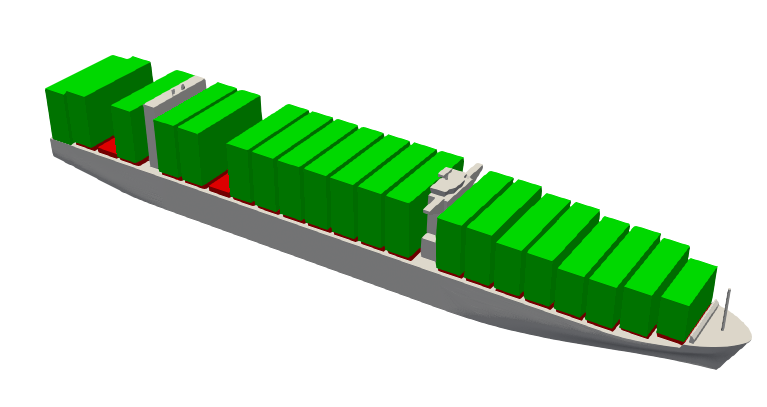}}

\caption{Comparison between the wind load coefficients predicted by the multi-fidelity (green) and single-fidelity (red) surrogate models for the loading configuration depicted in \ref{tr143}. Numerical data from the detailed CFD (open sea) are represented with black dots.}
\label{comparison_OS}
\end{figure*}

\begin{figure*}[htbp]
\subfloat[$C_X$]{%
  \includegraphics[width=0.4\textwidth]{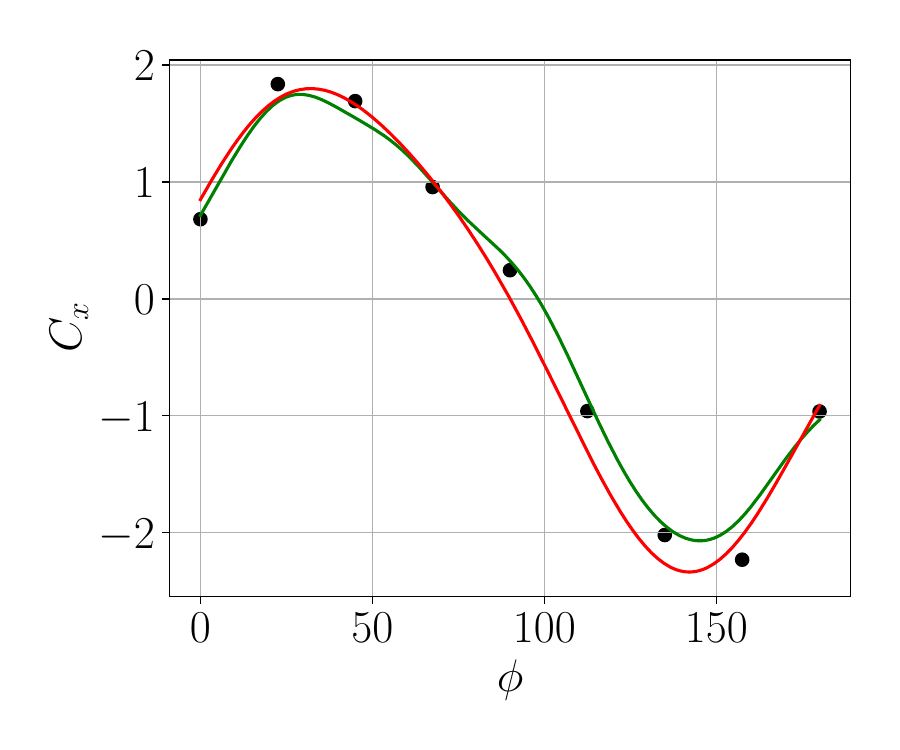}}
\hspace{0.05\textwidth}
\subfloat[$C_Y$]{%
  \includegraphics[width=0.4\textwidth]{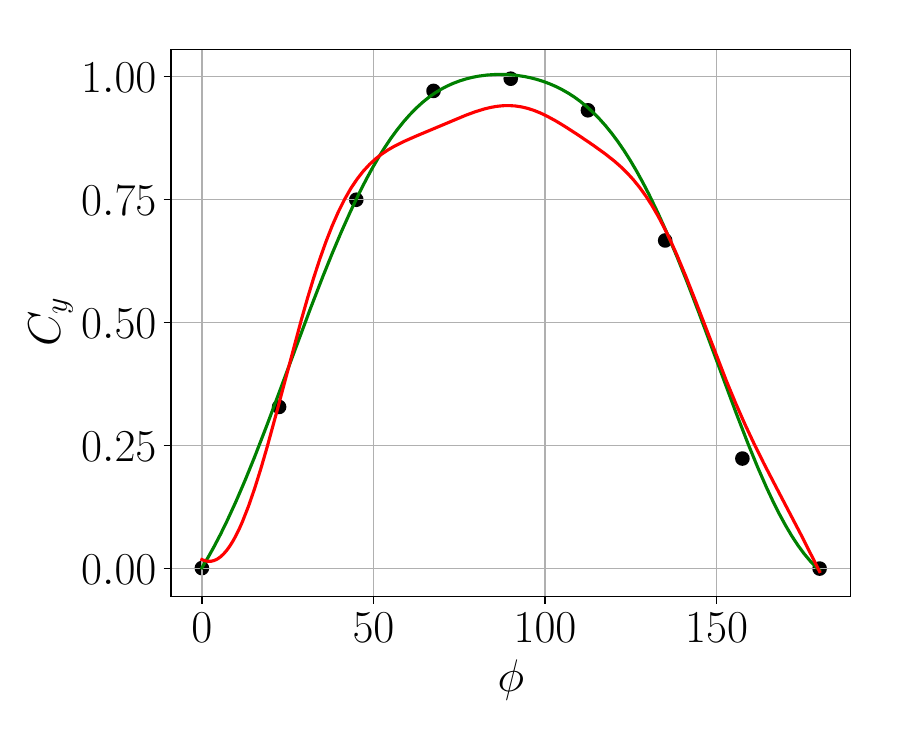}} \\

\subfloat[$C_M$]{%
  \includegraphics[width=0.4\textwidth]{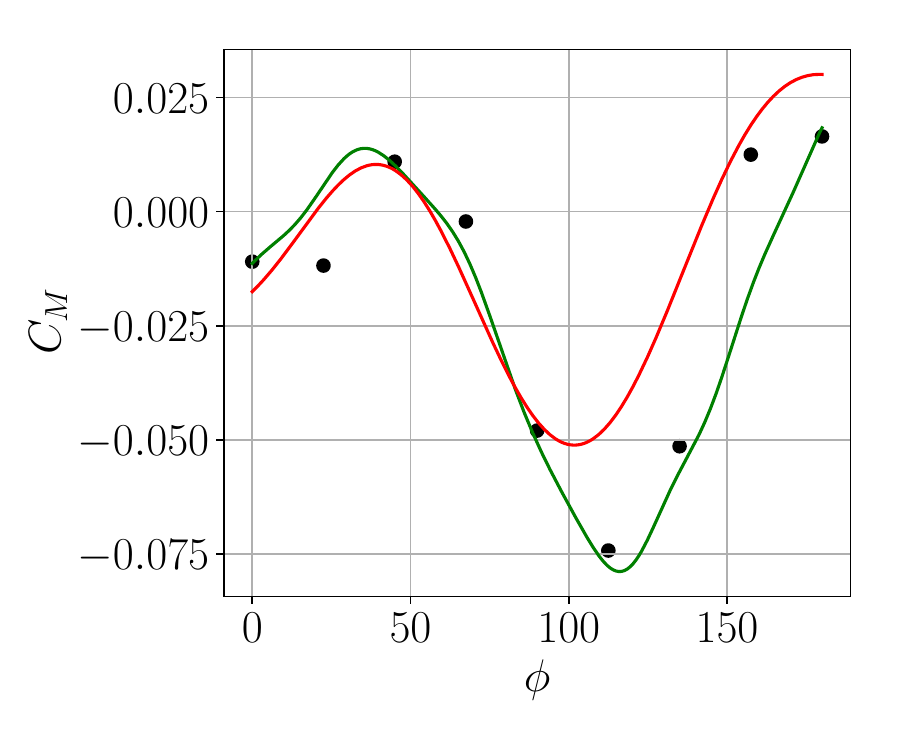}}
\hspace{0.05\textwidth}
\subfloat[Validation loading configuration]{%
\label{tr144}
  \includegraphics[width=0.35\textwidth]{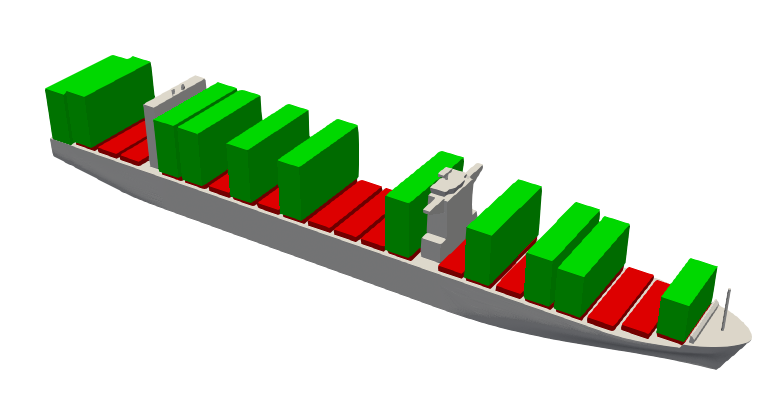}}

\caption{Comparison between the wind load coefficients predicted by the multi-fidelity (green) and single-fidelity (red) surrogate models for the loading configuration depicted in \ref{tr144}. Numerical data from the detailed CFD (open sea) are represented with black dots.}
\label{comparison_OS2}
\end{figure*}

Figures~\ref{comparison_OS} and \ref{comparison_OS2} show the predicted wind-load coefficients for two unseen loading configurations (Figs.~\ref{tr143} and \ref{tr144}). For an equivalent number of high-fidelity samples, the multi-fidelity model consistently provides more accurate predictions than the single-fidelity model. This demonstrates that the lower-fidelity data sources, although individually less accurate (see Fig.~\ref{comparison_wl}), provide valuable information that improves the global reconstruction of the response.

\begin{figure*}[htbp]
\centering
	\subfloat[][]{%
		\label{tr13}
		\includegraphics[width=0.45\textwidth]{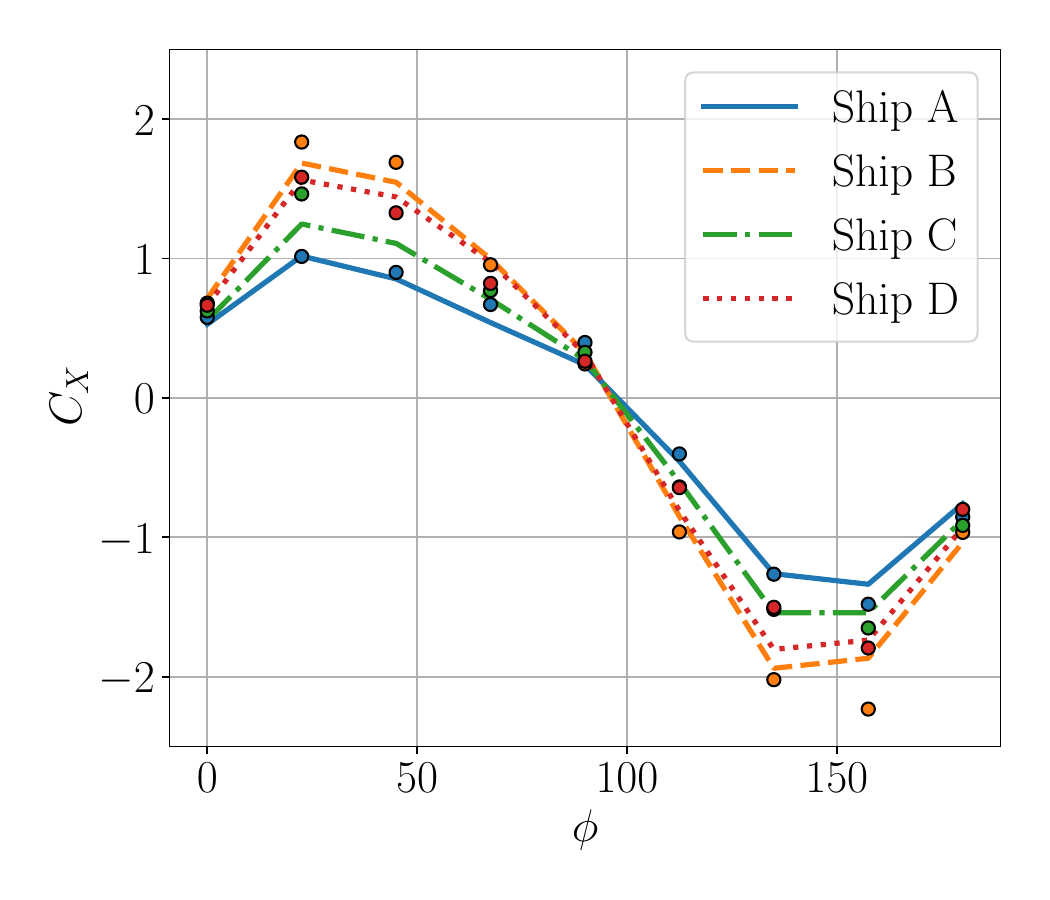}}
\hspace{0.05\textwidth}
\subfloat[][]{\label{CY_OS}
	\includegraphics[width=0.45\textwidth]{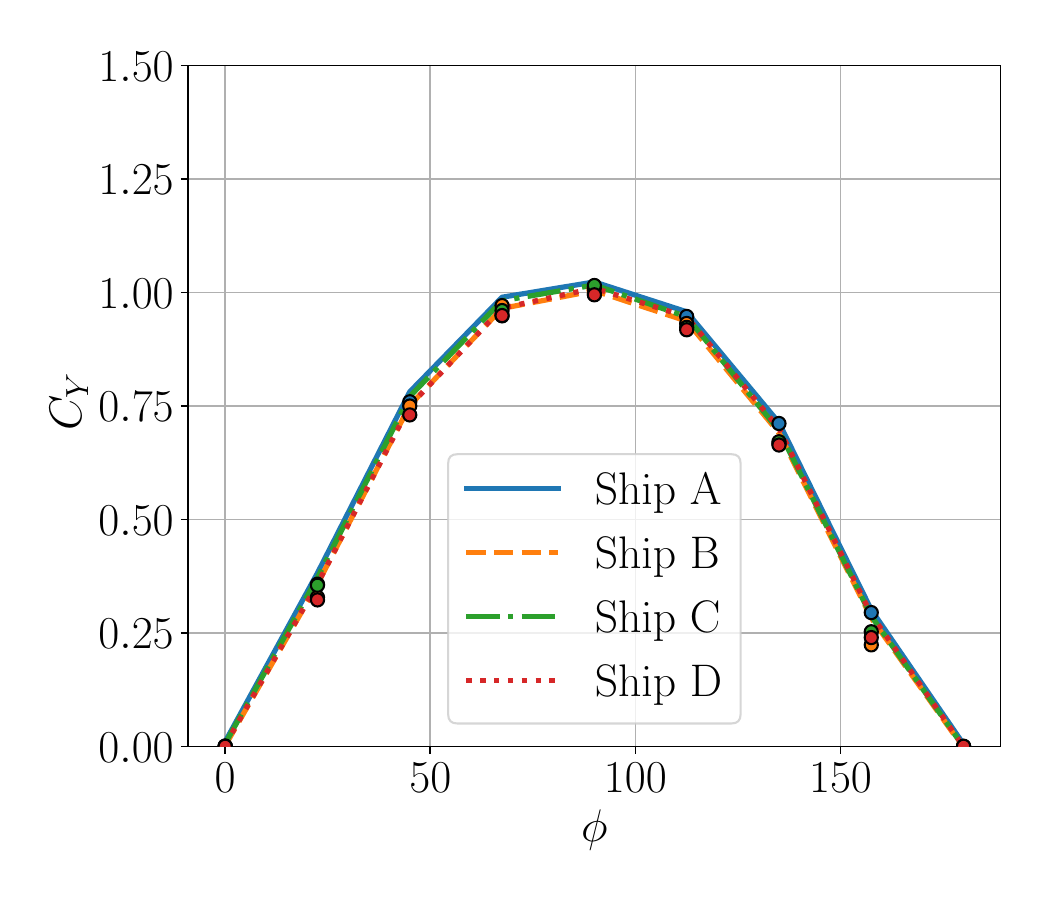}} \\
	\subfloat[][]{%
		\label{tr14}
		\includegraphics[width=0.45\textwidth]{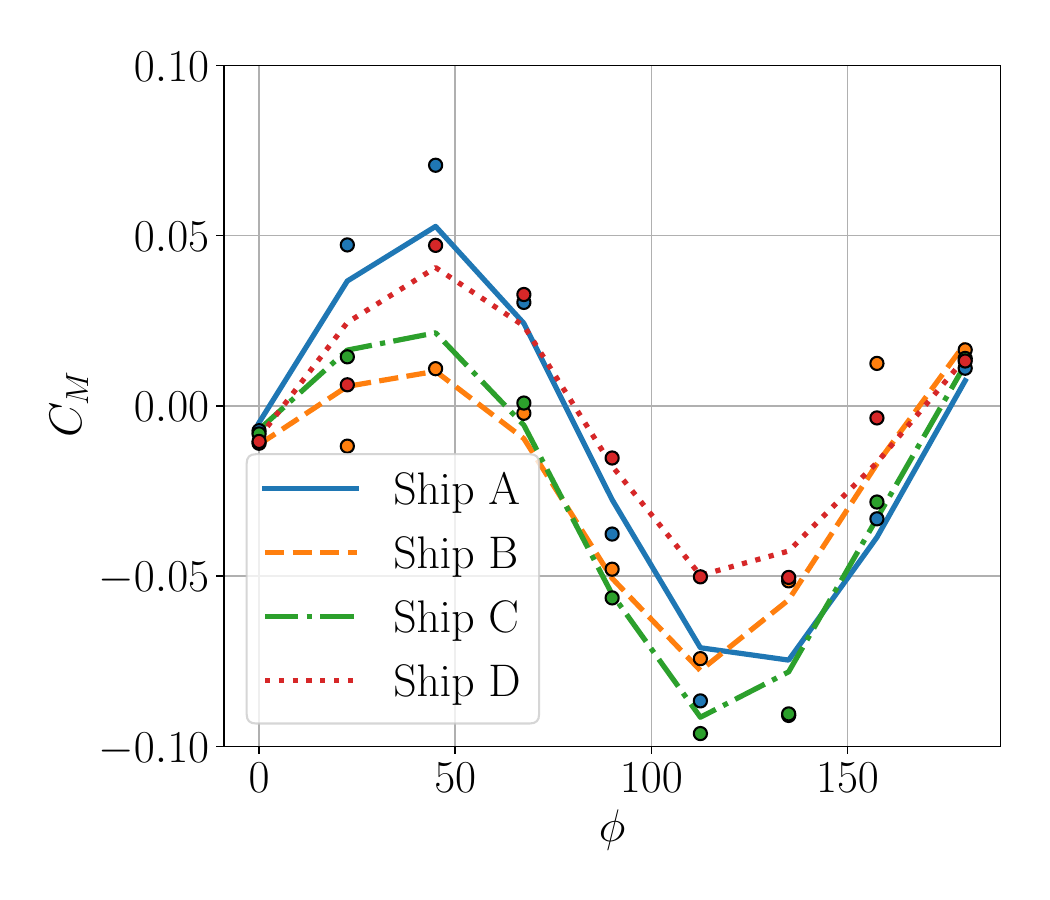}

		}

	\caption{Comparison between predictions of the multi-fidelity surrogate models and simulation data of high-fidelity (detailed CFD) for the load configurations A-D depicted in Figure \ref{validation_confs}. }
	\label{comparison_ships}
\end{figure*}

A more extensive validation is performed for the configurations shown in Fig.~\ref{validation_confs}. The predicted wind-load coefficients for configurations A–D are compared with high-fidelity CFD results in Fig.~\ref{comparison_ships}. The multi-fidelity model captures the increase in the longitudinal coefficient $C_X$ from configurations A and C to B and D, associated with reduced container shielding and a less guided flow, leading to increased drag. While the predictions of $C_Y$ remain accurate across configurations, the yaw moment coefficient $C_M$ exhibits larger discrepancies. Nevertheless, the overall trends and angular dependence are well reproduced.

\begin{figure*}[htbp]
\centering
	\subfloat[][$C_X$]{%
		\label{tr13}
		\includegraphics[width=0.45\textwidth]{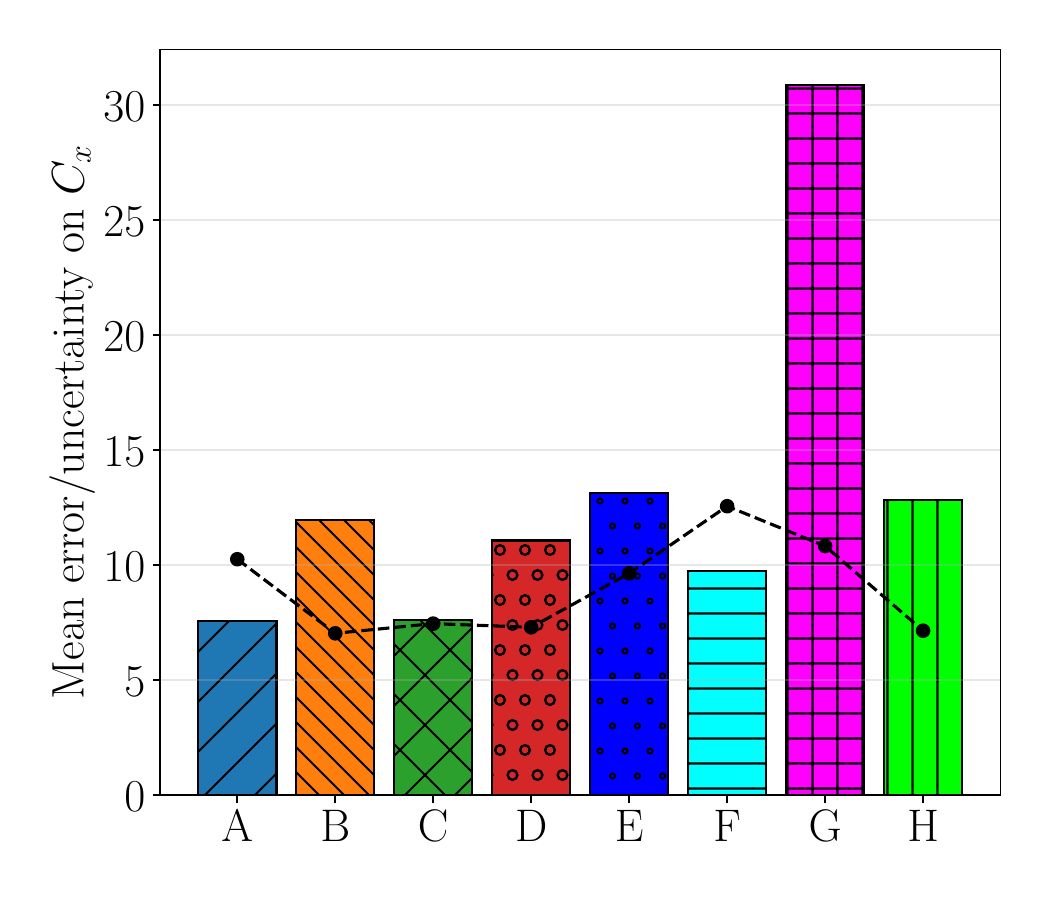}}
\hspace{0.05\textwidth}
\subfloat[][$C_Y$]{%
		\includegraphics[width=0.45\textwidth]{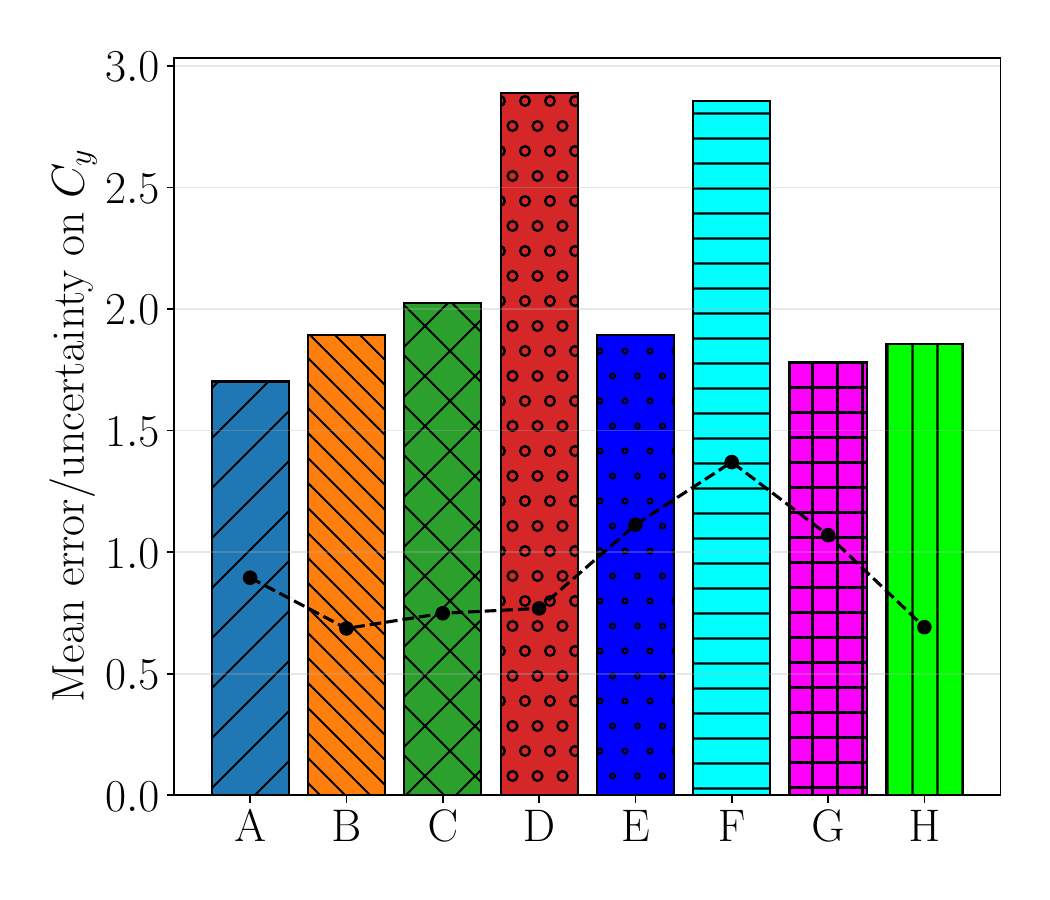}} \\
	\subfloat[][$C_M$]{%
		\label{tr14}
		\includegraphics[width=0.45\textwidth]{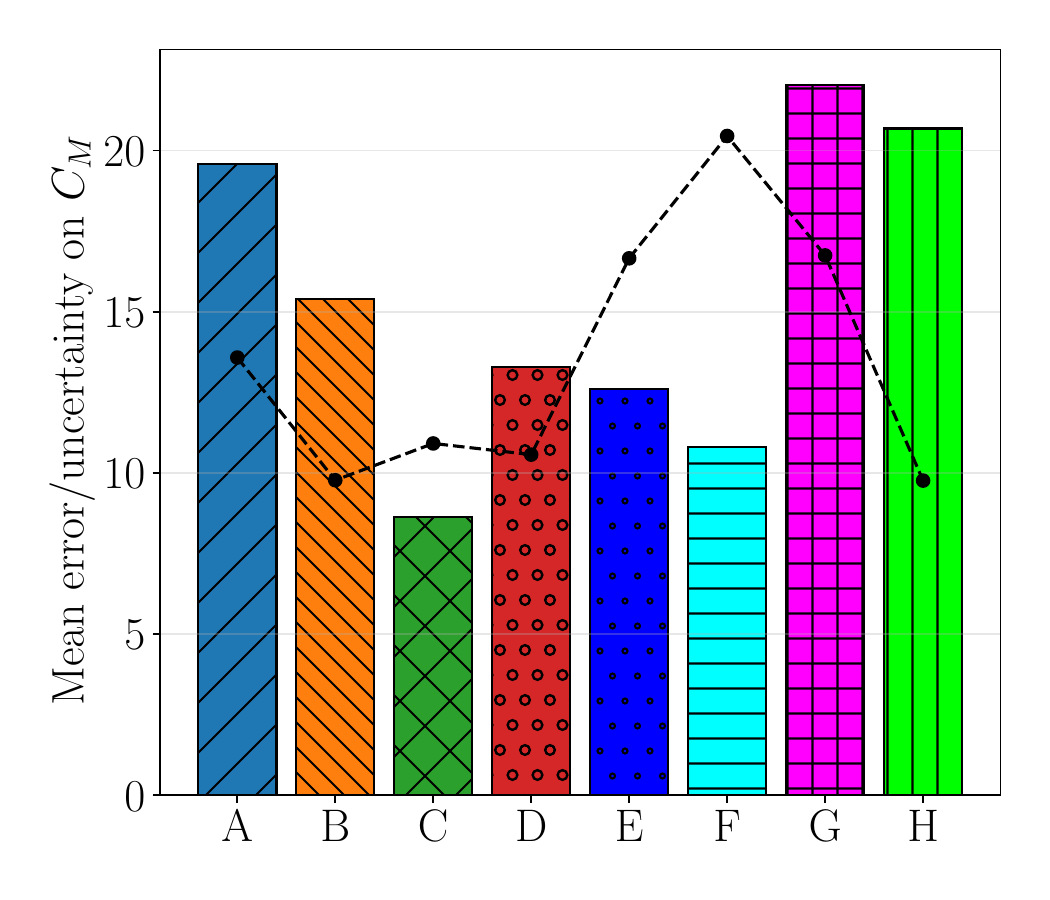}

		}

	\caption{Mean absolute errors (\%) of the multi-fidelity surrogate model with respect to the detailed CFD in open sea for the load configurations depicted in Figure \ref{validation_confs}. The average uncertainties predicted by the multi-fidelity GPs for each configuration are indicated with black dotted lines.}
	\label{comparison_bars}
\end{figure*}

For a quantitative assessment, Fig.~\ref{comparison_bars} reports the mean absolute errors for each loading configuration. The average error in $C_X$ remains below 15\% for most configurations, with a maximum of approximately 30\% for configuration G. The lateral coefficient $C_Y$ is predicted with high accuracy, with errors below 3\% for all configurations. In contrast, $C_M$ exhibits larger errors, reaching approximately 20\% for configurations A, G and H.

The same figure also reports the average epistemic uncertainty predicted by the multi-fidelity surrogate model (Eq.~\eqref{unc_multi_fidelity}), shown as black dashed lines. The predicted uncertainties are of the same order of magnitude as the observed errors, indicating that the model provides a consistent estimate of its own confidence and that the true values are generally expected to lie within, or close to, the predicted uncertainty bounds.

\subsection{Results for containers and tanks environments}

Additional multi-fidelity surrogate models are constructed to predict wind loads in the presence of surrounding structures, namely container and tank environments (Fig.~\ref{fig:loadings_envs}). In these cases, four fidelity levels are considered in the autoregressive framework (Fig.~\ref{MF_algorithm2}): detailed CFD including the environment (highest fidelity), detailed CFD in open-sea conditions, simplified CFD, and the empirical correlation. This hierarchical structure enables the training process to leverage the design space exploration previously performed for the open-sea case.

\begin{figure*}[h!]
	\subfloat[$C_X$]{%
		\label{tr13}
		\includegraphics[width=0.4\textwidth]{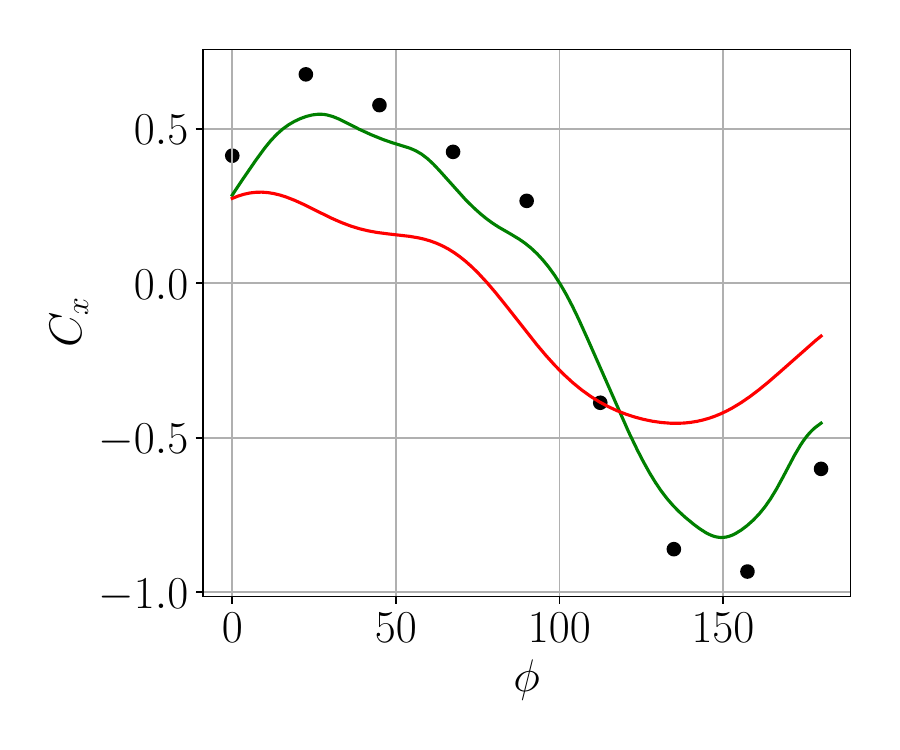}}
	\hspace{0.05\textwidth}
        \subfloat[$C_Y$]{
		\includegraphics[width=0.4\textwidth]{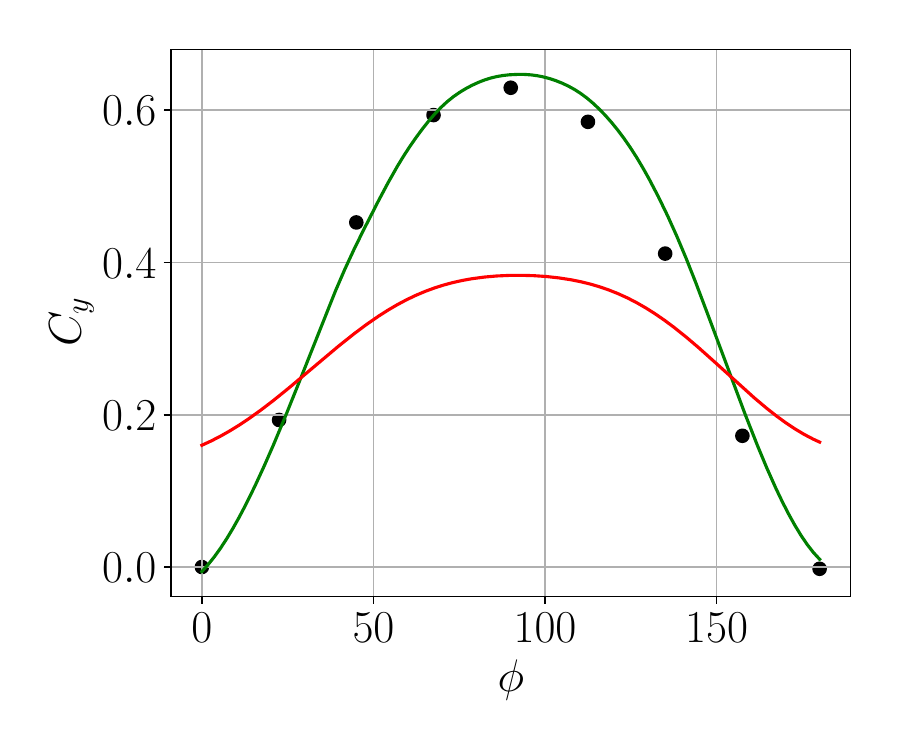}} \\
	\subfloat[$C_M$]{%
		\includegraphics[width=0.4\textwidth]{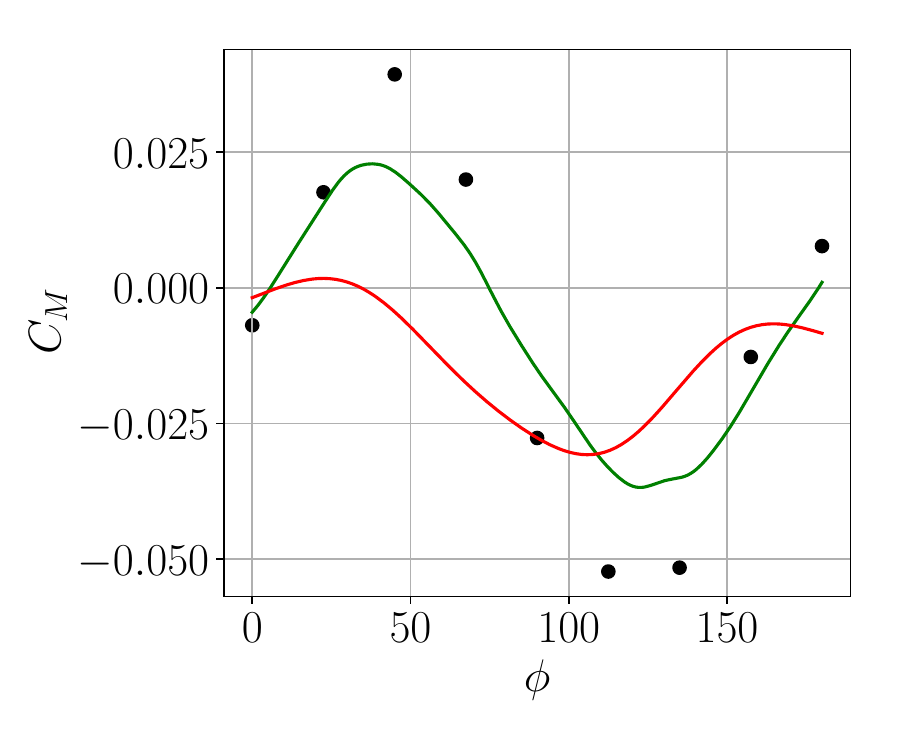}}
    \hspace{0.05\textwidth}
        \subfloat[Validation loading configuration]{\label{val_CE1}
        		\includegraphics[width=0.35\textwidth]{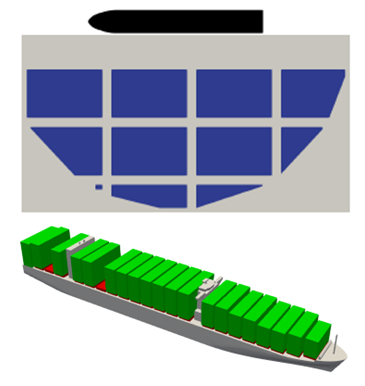}
		}

	\caption{Comparison between the wind load coefficients predicted by the multi-fidelity (green) and single-fidelity (red) surrogate models for the loading configuration depicted in \ref{val_CE1} (right). Numerical data from the detailed CFD (container environment) are represented with black dots.}
	\label{comparison_CE}
\end{figure*}

\begin{figure*}[h!]
	\subfloat[$C_X$]{%
		\label{tr13}
		\includegraphics[width=0.4\textwidth]{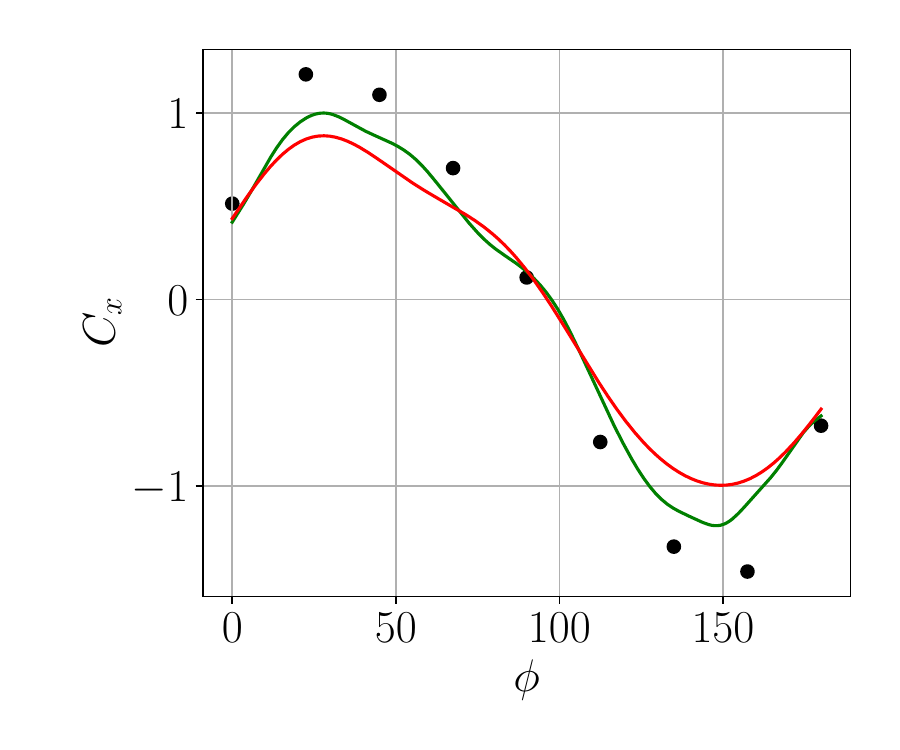}}
	\hspace{0.05\textwidth}
        \subfloat[$C_Y$]{
		\includegraphics[width=0.4\textwidth]{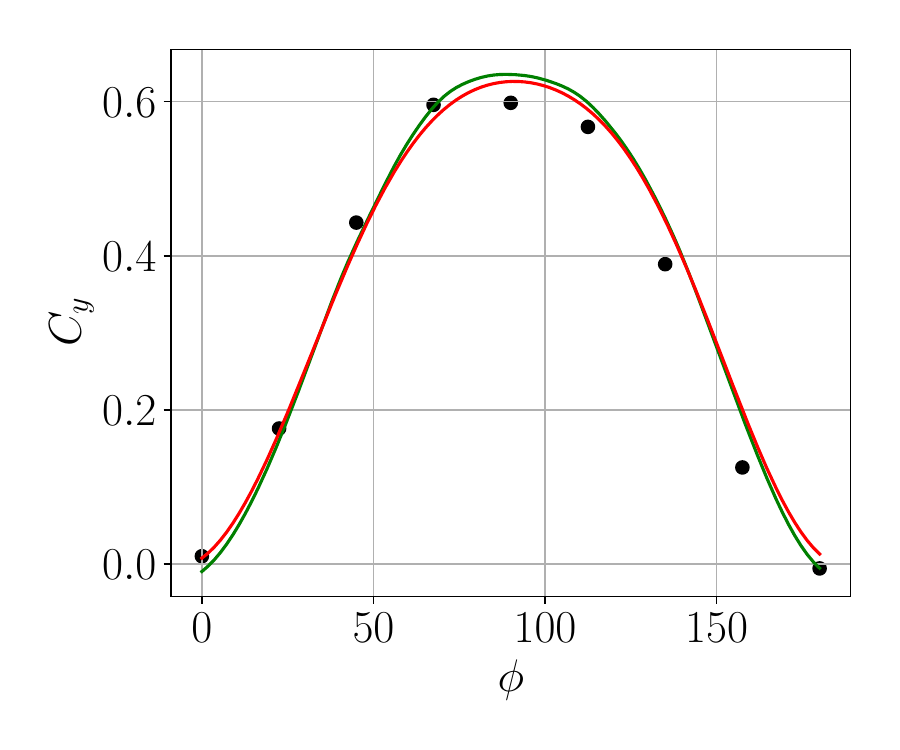}} \\
	\subfloat[$C_M$]{%
		\label{tr14}
		\includegraphics[width=0.4\textwidth]{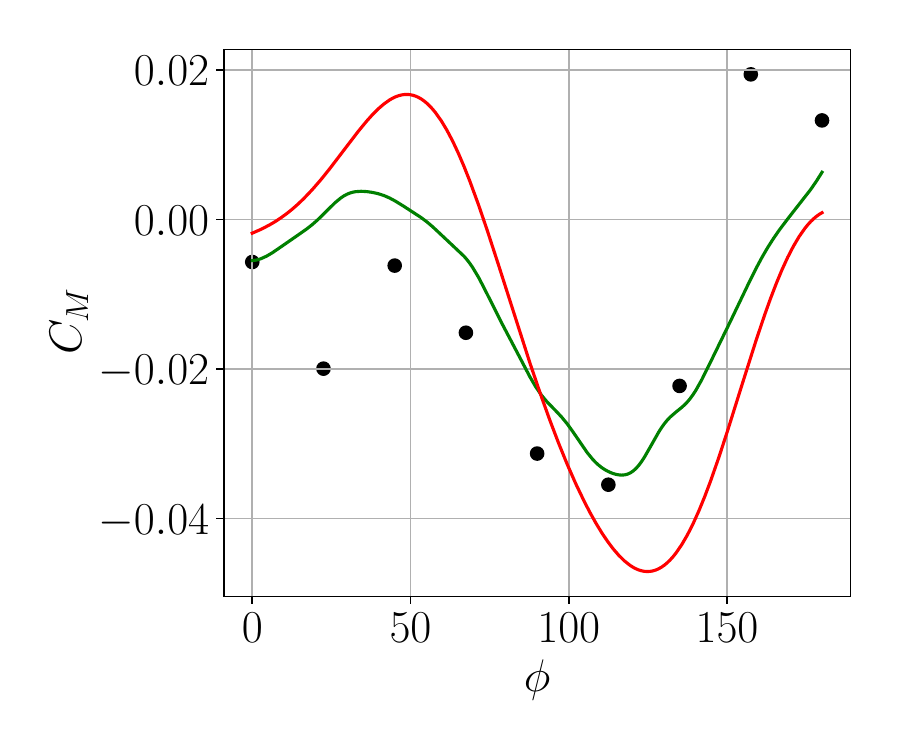}}
  \hspace{0.05\textwidth}
        \subfloat[Validation loading configuration]{\label{val_conf2}
        		\includegraphics[width=0.35\textwidth]{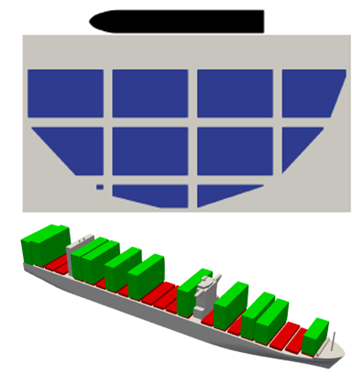}
		}

	\caption{Comparison between the wind load coefficients predicted by the multi-fidelity (green) and single-fidelity (red) surrogate models for the loading configuration depicted in \ref{val_conf2} (right). Numerical data from the detailed CFD (container environment) are represented with black dots.}
	\label{comparison_CE2}
\end{figure*}

\begin{figure*}[h!]
	\subfloat[$C_X$]{%
		\label{tr13}
		\includegraphics[width=0.4\textwidth]{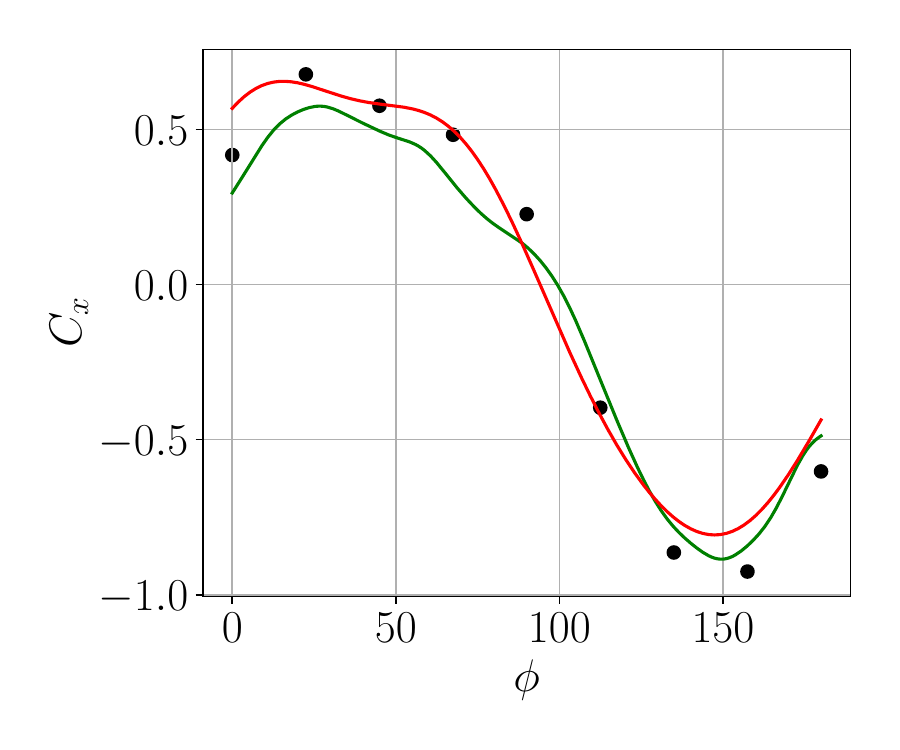}}
		\hspace{0.05\textwidth}
        \subfloat[$C_Y$]{
		\includegraphics[width=0.4\textwidth]{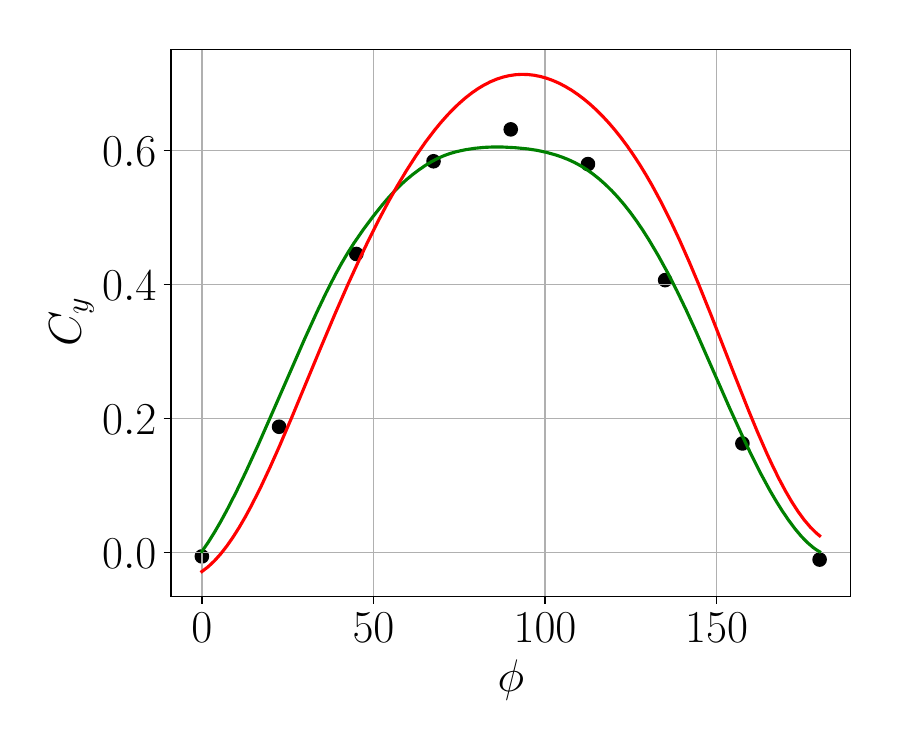}} \\
	\subfloat[$C_M$]{%
		\label{tr14}
		\includegraphics[width=0.4\textwidth]{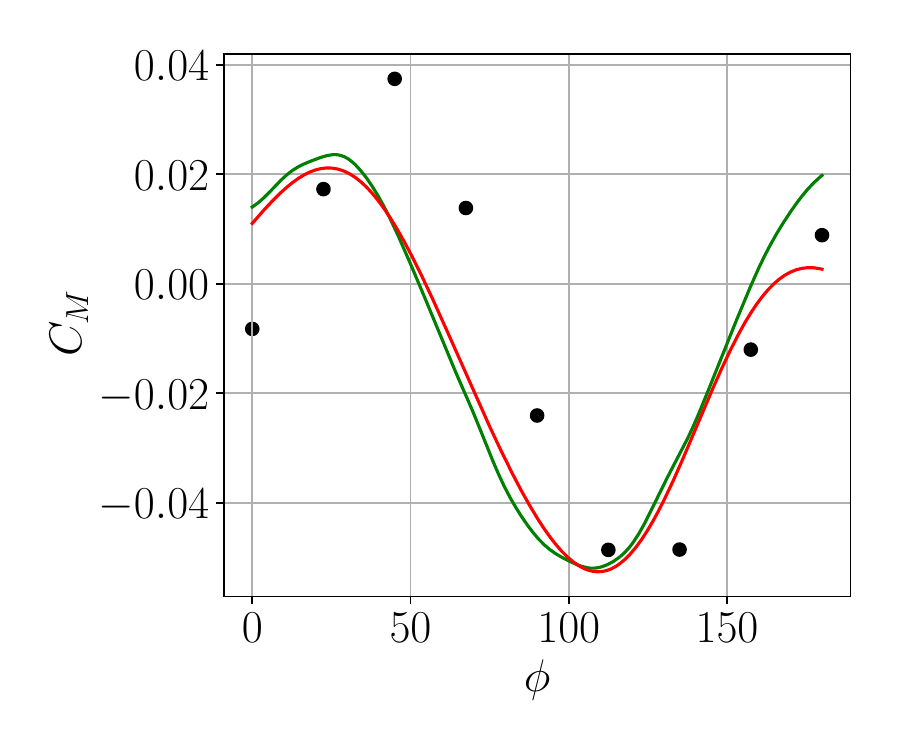}}
     \hspace{0.05\textwidth}
        \subfloat[Validation loading configuration]{\label{val_confTE1}
        		\includegraphics[width=0.35\textwidth]{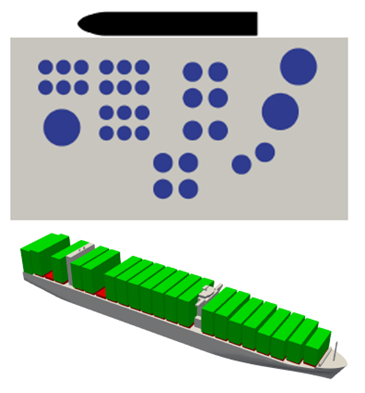}
		}

	\caption{Comparison between the wind load coefficients predicted by the multi-fidelity (green) and single-fidelity (red) surrogate models for the loading configuration depicted in \ref{val_confTE1} (right). Numerical data from the detailed CFD (container environment) are represented with black dots.}
	\label{comparison_TE}
\end{figure*}

\begin{figure*}[h!]
	\subfloat[$C_X$]{%
		\label{tr13}
		\includegraphics[width=0.4\textwidth]{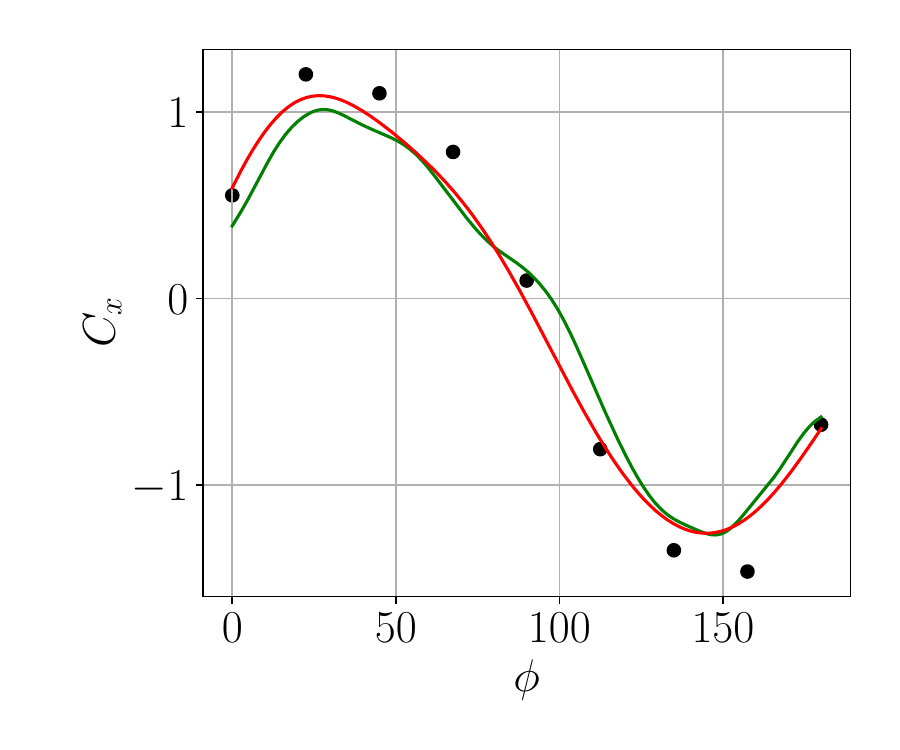}}
	\hspace{0.05\textwidth}
        \subfloat[$C_Y$]{
		\includegraphics[width=0.4\textwidth]{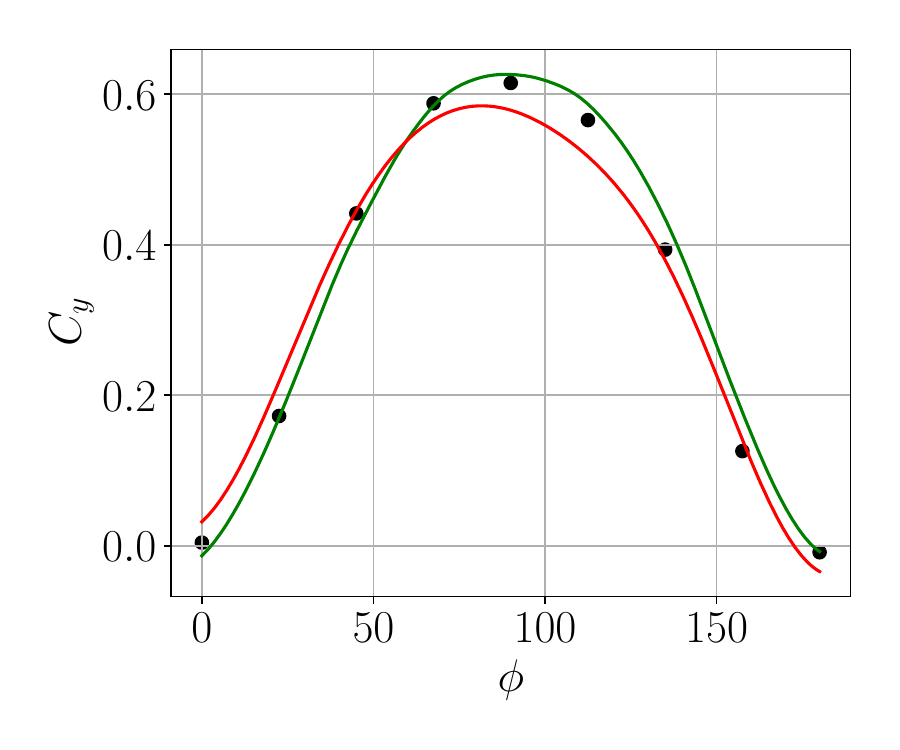}} \\
	\subfloat[$C_M$]{%
		\label{tr14}
		\includegraphics[width=0.4\textwidth]{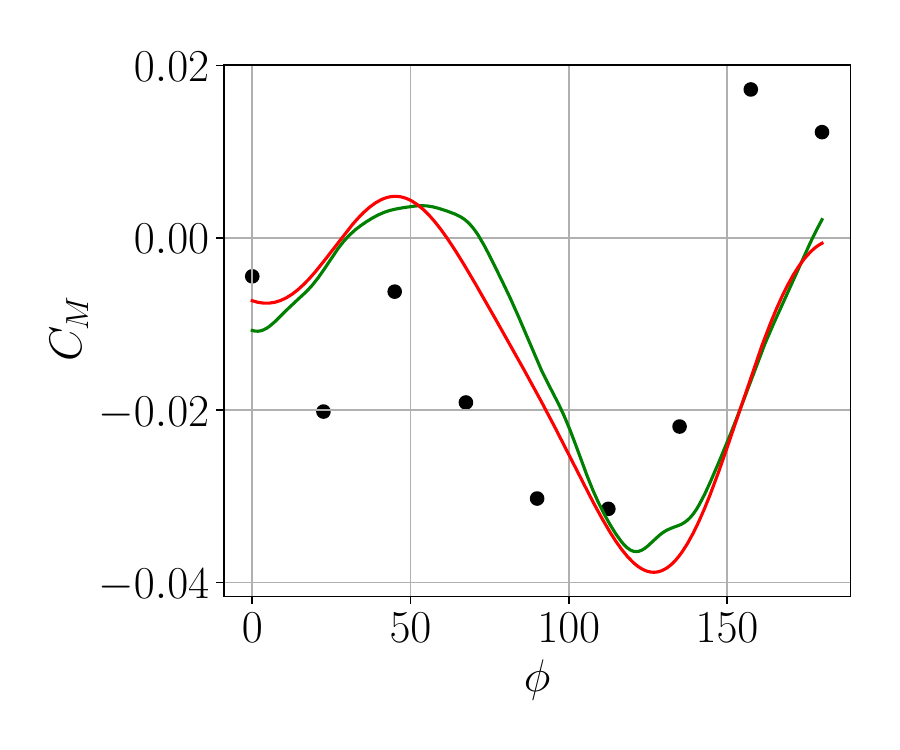}}
    \hspace{0.05\textwidth}
        \subfloat[Validation loading configuration]{\label{val_confTE2}
        		\includegraphics[width=0.35\textwidth]{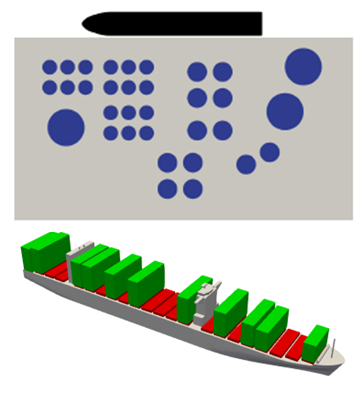}
		}

	\caption{Comparison between the wind load coefficients predicted by the multi-fidelity (green) and single-fidelity (red) surrogate models for the loading configuration depicted in \ref{val_confTE2} (right). Numerical data from the detailed CFD (container environment) are represented with black dots.}
	\label{comparison_TE2}
\end{figure*}

The predictions of the environment-specific surrogate models for configurations A and B are shown in Figs.~\ref{comparison_CE}–\ref{comparison_CE2} (containers environment) and Figs.~\ref{comparison_TE}–\ref{comparison_TE2} (tanks environment). In both cases, the comparison with single-fidelity Gaussian Process models further emphasizes the benefit of the multi-fidelity approach. Despite the limited number of high-fidelity simulations available for each environment, the multi-fidelity models provide accurate predictions of $C_X$, $C_Y$, and $C_M$ by exploiting the information propagated from the lower-fidelity levels.

Larger discrepancies are observed for the yaw moment coefficient $C_M$, particularly in the tank environment, where the flow interactions with the surrounding structures are more complex. This indicates that additional high-fidelity samples would be required to further improve the accuracy in these regimes.

\begin{figure*}[htbp]
\centering
	\subfloat[][$C_X$]{%
		\label{tr13}
		\includegraphics[width=0.45\textwidth]{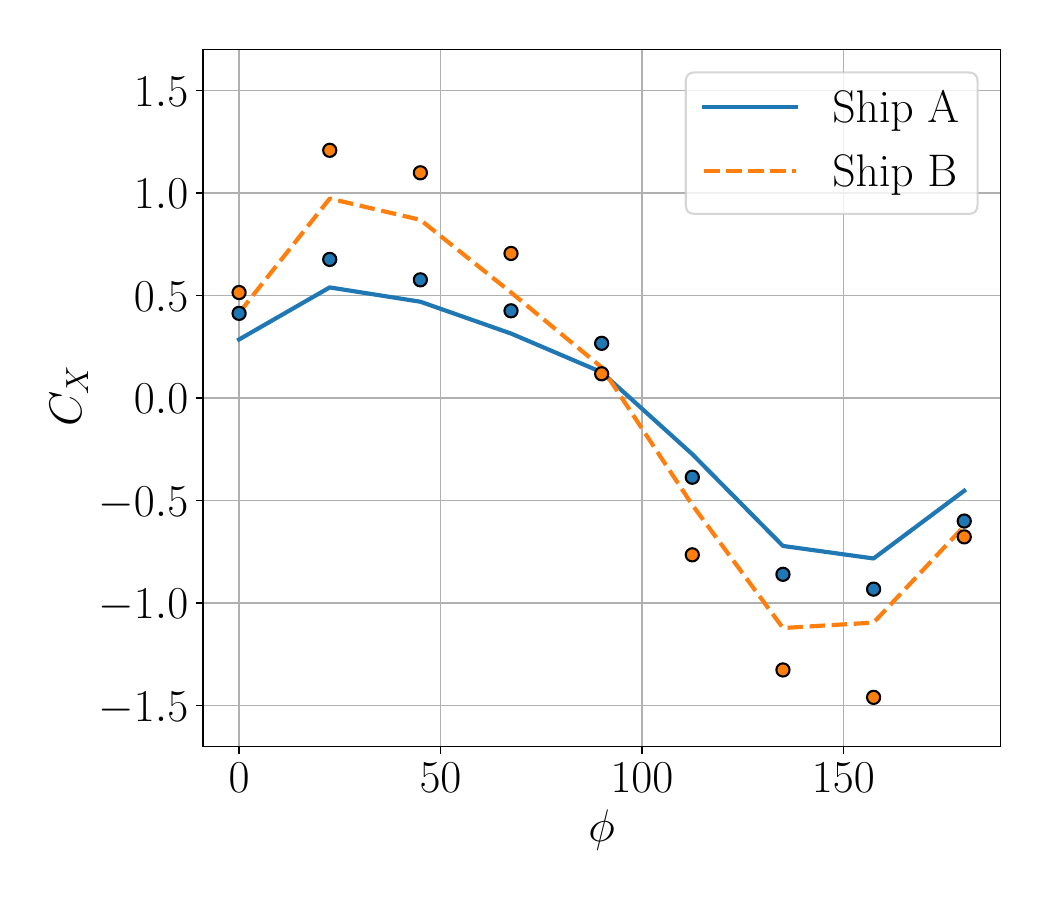}}
\hfill
\subfloat[][$C_Y$]{\label{CY_CE}
		\includegraphics[width=0.45\textwidth]{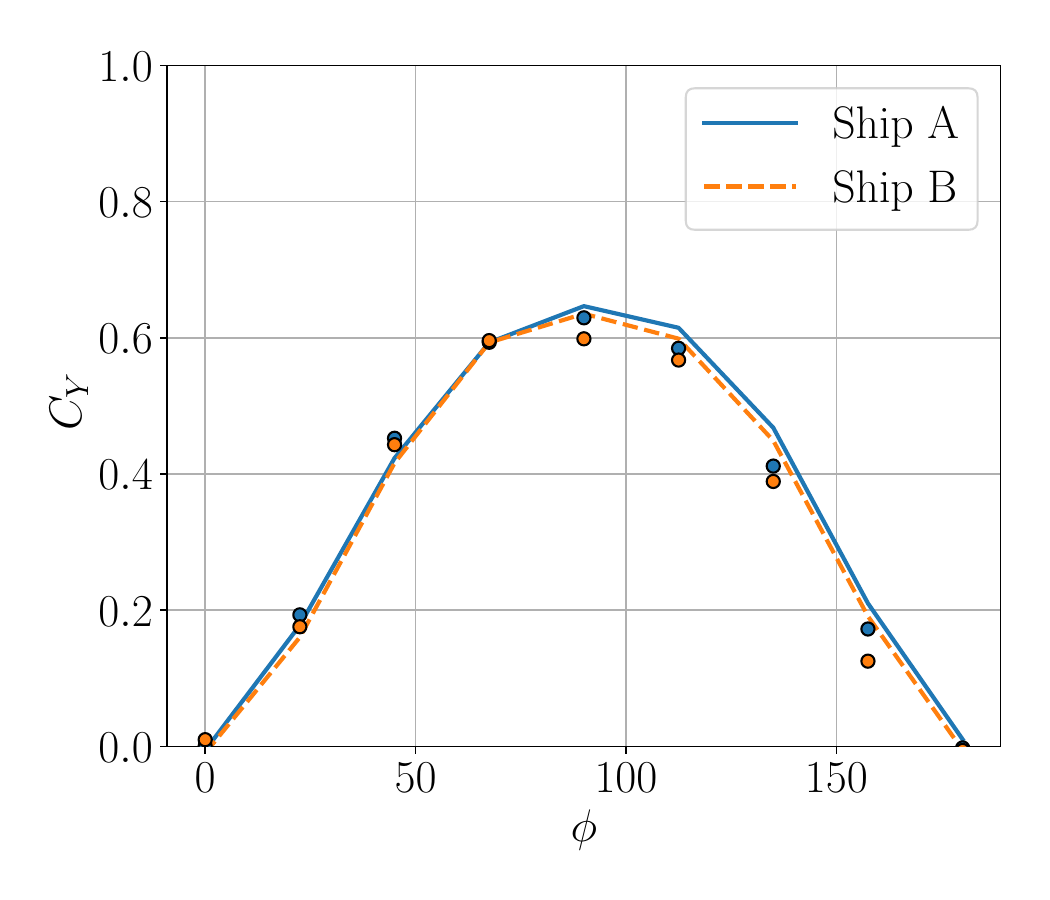}} \\
	\subfloat[][$C_M$]{%
		\label{tr14}
		\includegraphics[width=0.45\textwidth]{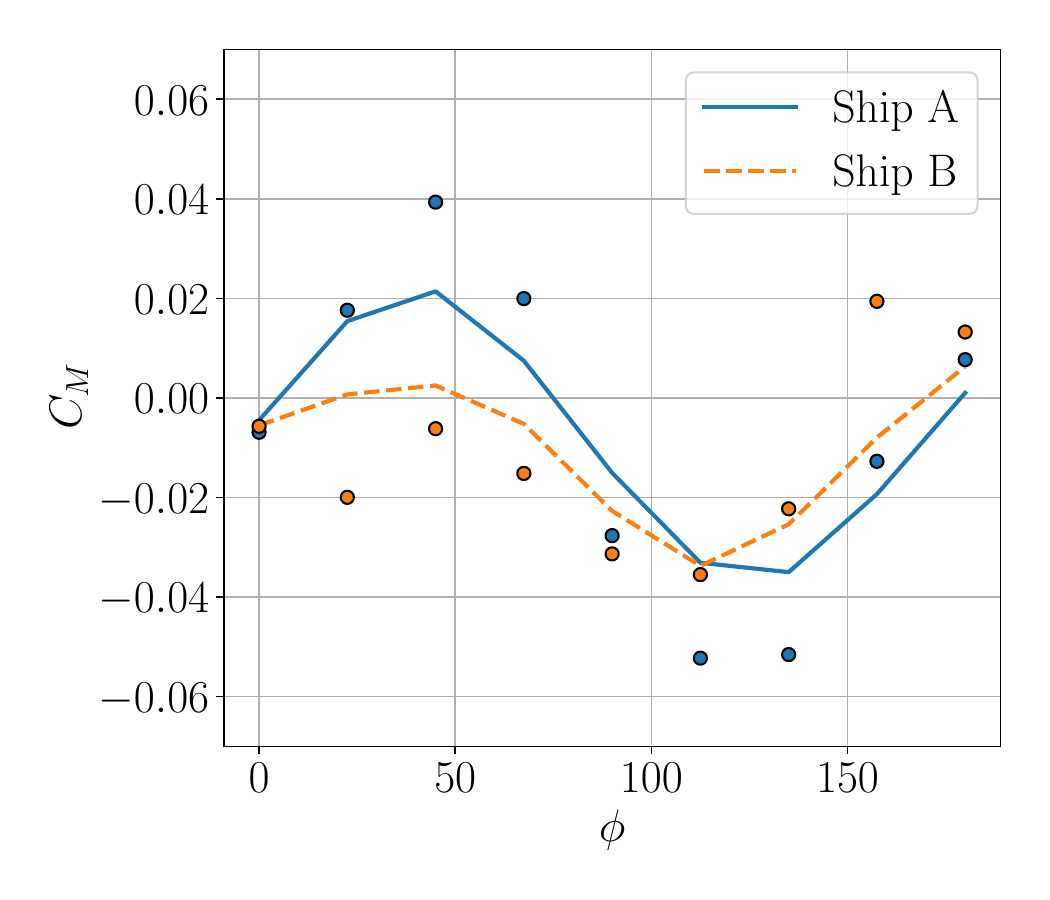}

		}

	\caption{Comparison between predictions and simulation data of high-fidelity (detailed CFD) on the test dataset. Predictions of the multi-fidelity surrogate model and its single-fidelity counterpart are compared.}
	\label{comparison_2CE}
\end{figure*}

\begin{figure*}[htbp]
\centering
	\subfloat[][]{%
		\label{tr13}
		\includegraphics[width=0.45\textwidth]{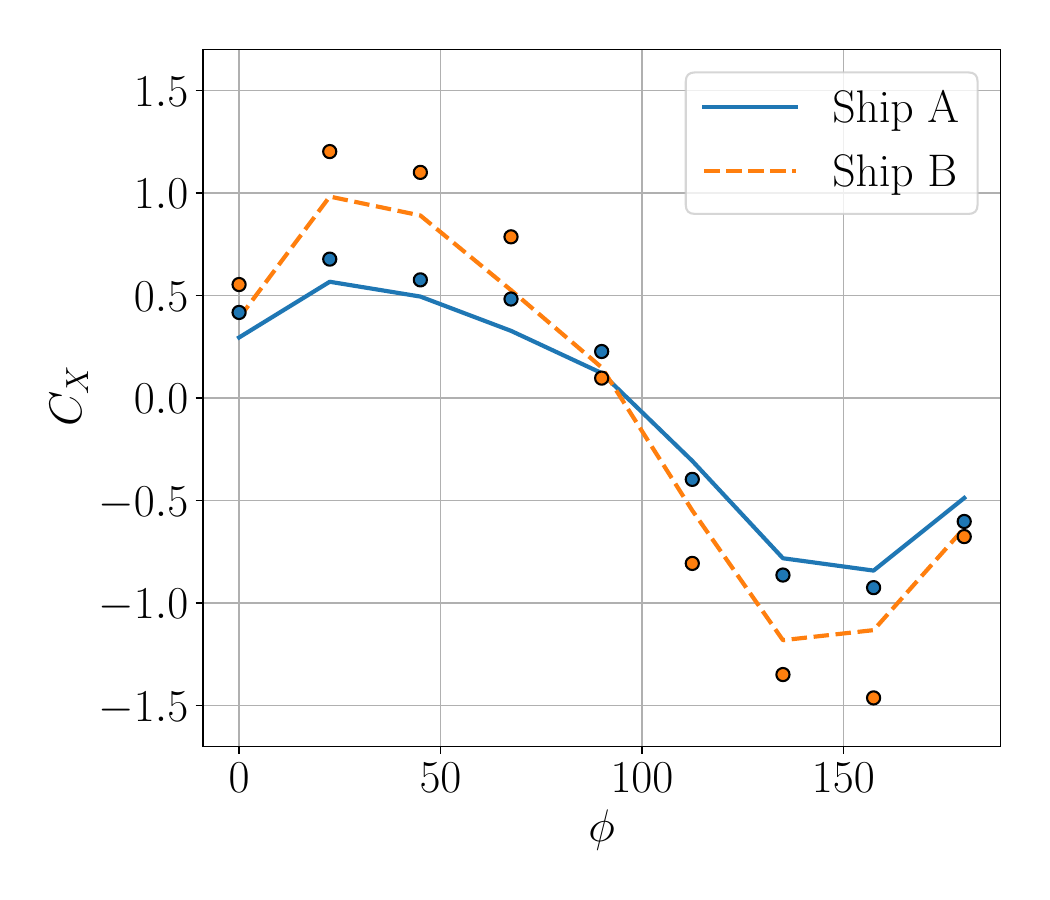}}
\hfill
\subfloat[][]{\label{CY_TE}
		\includegraphics[width=0.45\textwidth]{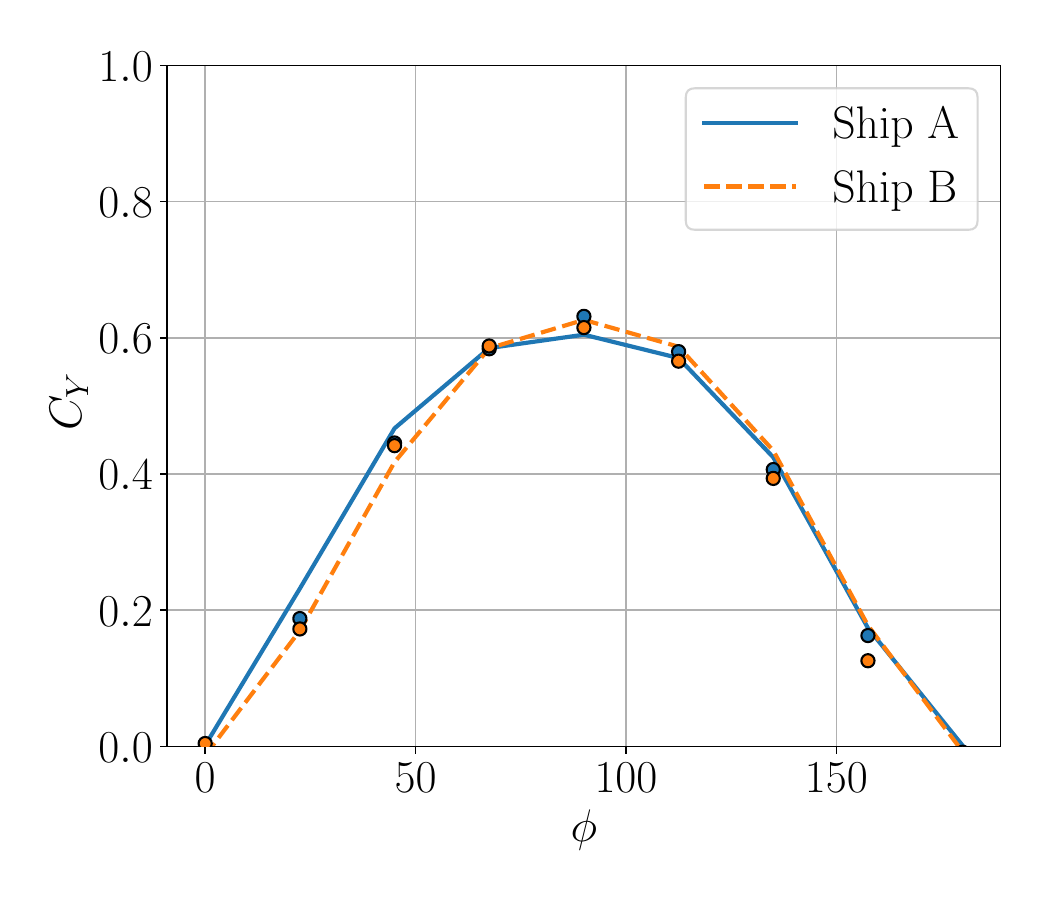}} \\
	\subfloat[][]{%
		\label{tr14}
		\includegraphics[width=0.45\textwidth]{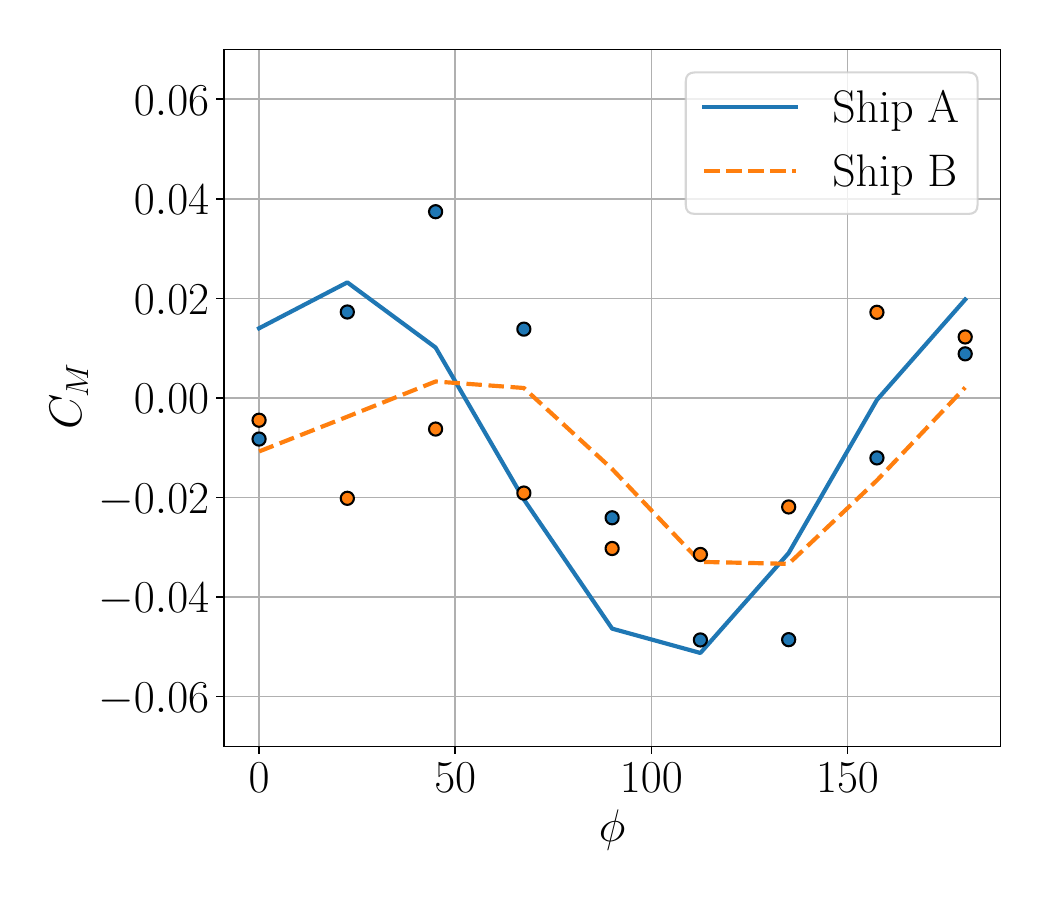}

		}

	\caption{Comparison between predictions and simulation data of high-fidelity (detailed CFD) on the test dataset. Predictions of the multi-fidelity surrogate model and its single-fidelity counterpart are compared.}
	\label{comparison_2TE}
\end{figure*}

A direct comparison between high-fidelity CFD data and multi-fidelity predictions for configurations A and B is provided in Figs.~\ref{comparison_2CE} and \ref{comparison_2TE}. As in the open-sea case, the surrogate models correctly capture the dependence of the wind loads on the loading configuration. In particular, the comparison of the lateral coefficient $C_Y$ across environments (Figs.~\ref{CY_CE}, \ref{CY_TE}, and \ref{CY_OS}) shows that the multi-fidelity models reproduce the sheltering effects induced by upstream structures, leading to a reduction of the lateral loads compared to open-sea conditions.

\subsection{Sensitivity analysis}

\begin{figure*}[h!]
	\centering
	\subfloat[]{%
		\label{tr13}
		\includegraphics[scale=0.4]{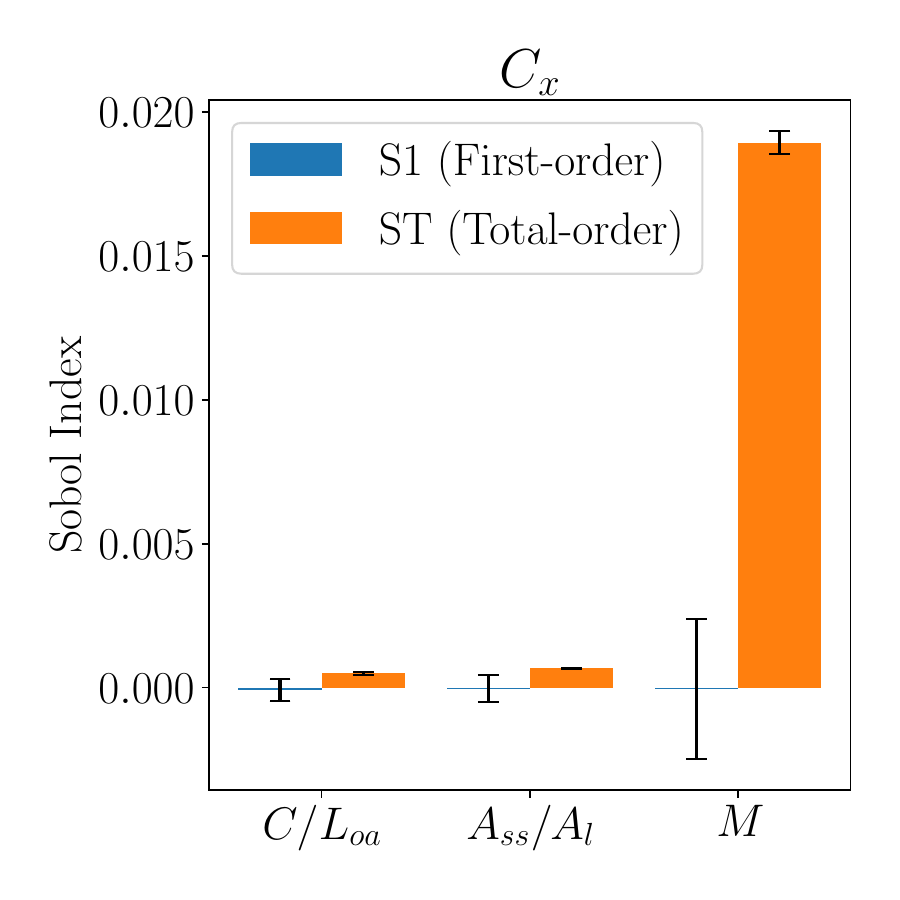}}
        \subfloat[]{
		\hspace{0.04\linewidth}
		\includegraphics[scale=0.4]{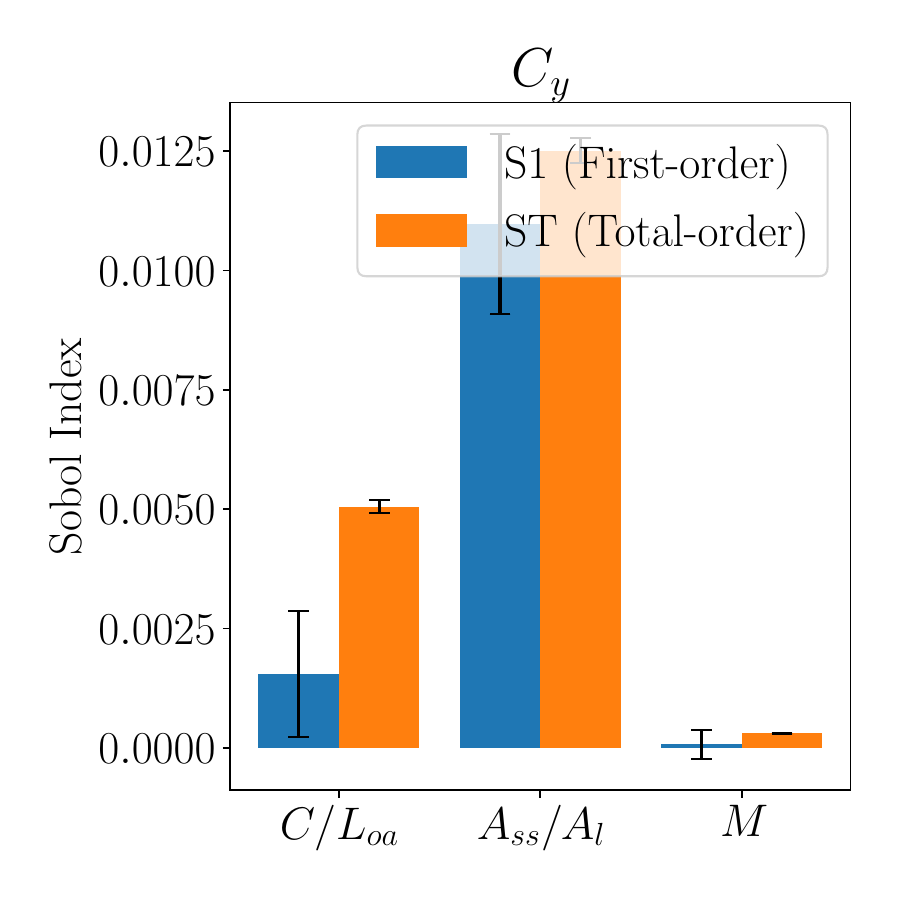}} \\
	\subfloat[]{%
		\label{tr14}
		\includegraphics[scale=0.4]{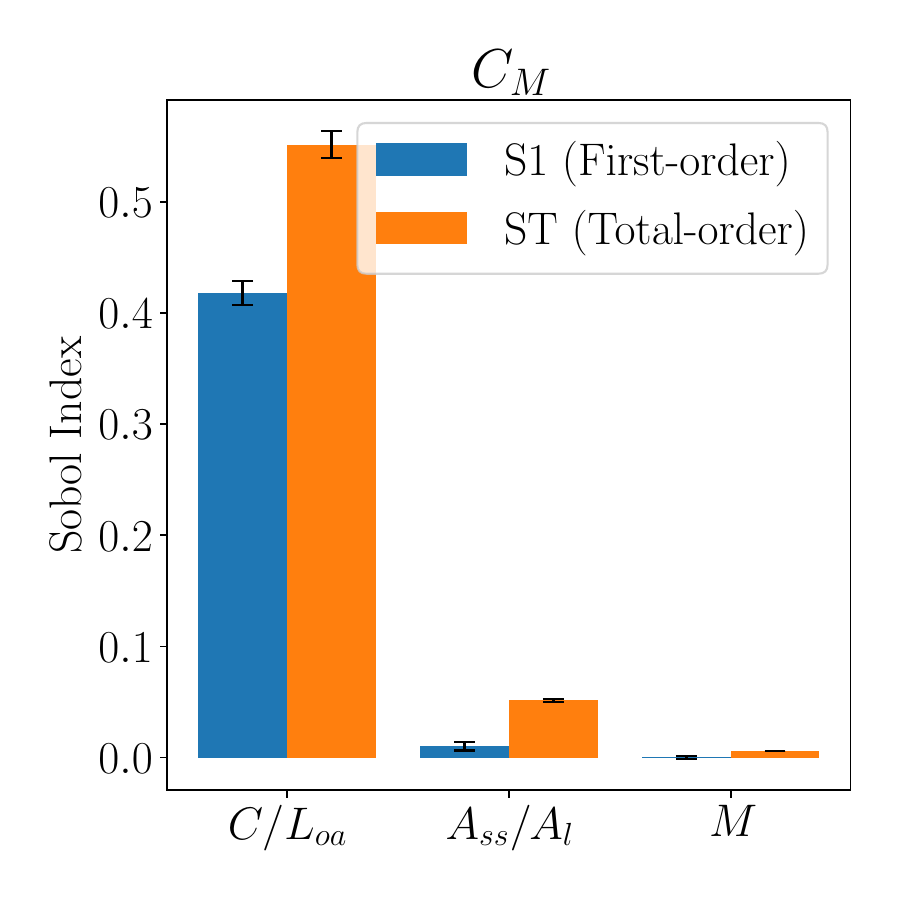}
		}
    
	\caption{Sobol sensitivity indices of the three wind load coefficients with respect to the main geometrical parameters of the ship.}
	\label{sobol_plot}
\end{figure*}

The Sobol sensitivity indices introduced in Section~\ref{sobol_indices} are computed in the reduced input space $\mathbf{z}$ by evaluating the surrogate models over a large set of samples and estimating the corresponding variance-based statistics. The resulting indices for the coefficients $C_X$, $C_Y$, and $C_M$ are reported in Fig.~\ref{sobol_plot}.

The analysis reveals that the dominant geometrical parameters differ significantly across the three wind-load coefficients. The longitudinal force coefficient $C_X$ is primarily governed by the number of container groups, which controls the formation of recirculation zones and low-pressure regions within the container stack. In contrast, the lateral force coefficient $C_Y$ is most sensitive to the superstructure area, although its overall variability with respect to the geometric parameters remains limited. 

The yaw moment coefficient $C_M$ is predominantly influenced by the centroid of the lateral projected area, reflecting the strong dependence of the moment arm on the spatial distribution of the aerodynamic loads. The magnitude of both first-order and total Sobol indices indicates that $C_M$ exhibits the strongest sensitivity to geometric variations, making it inherently more challenging to predict accurately. This observation is consistent with the larger discrepancies observed in the surrogate predictions for $C_M$, particularly in the presence of complex environments (Figs.~\ref{comparison_2CE} and \ref{comparison_2TE}).

\begin{figure*}[htbp!] 
	\centering
	\includegraphics[scale=0.4]{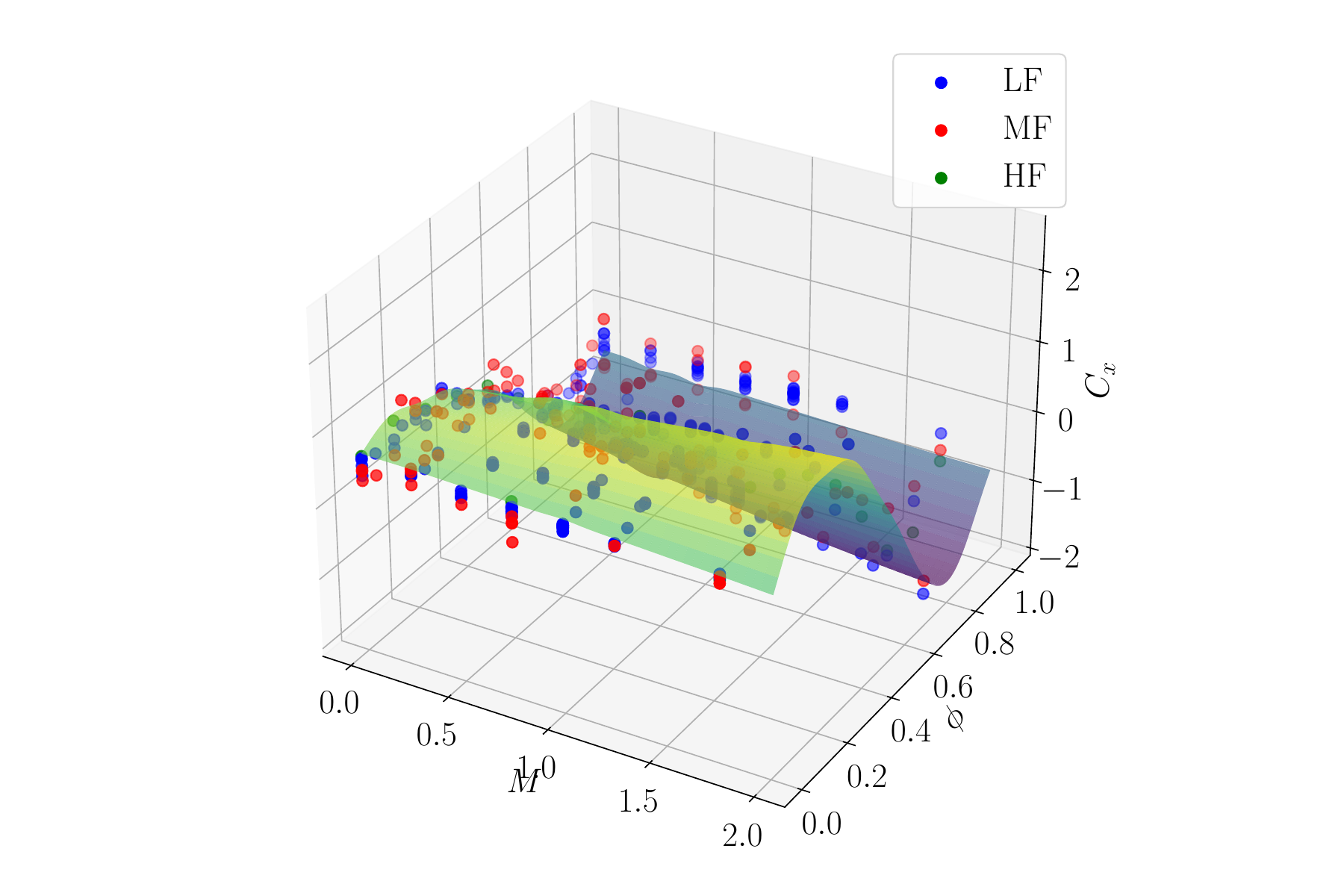}
	\caption{Response surface of the multi-fidelity surrogate model showing the variability of the longitudinal wind force coefficient with $\phi$ and $M$ (the other, less sensitive parameters are set to their average values).}
	\label{surf_Cx}
\end{figure*}

\begin{figure*}[htbp!]
	\centering
	\includegraphics[scale=0.4]{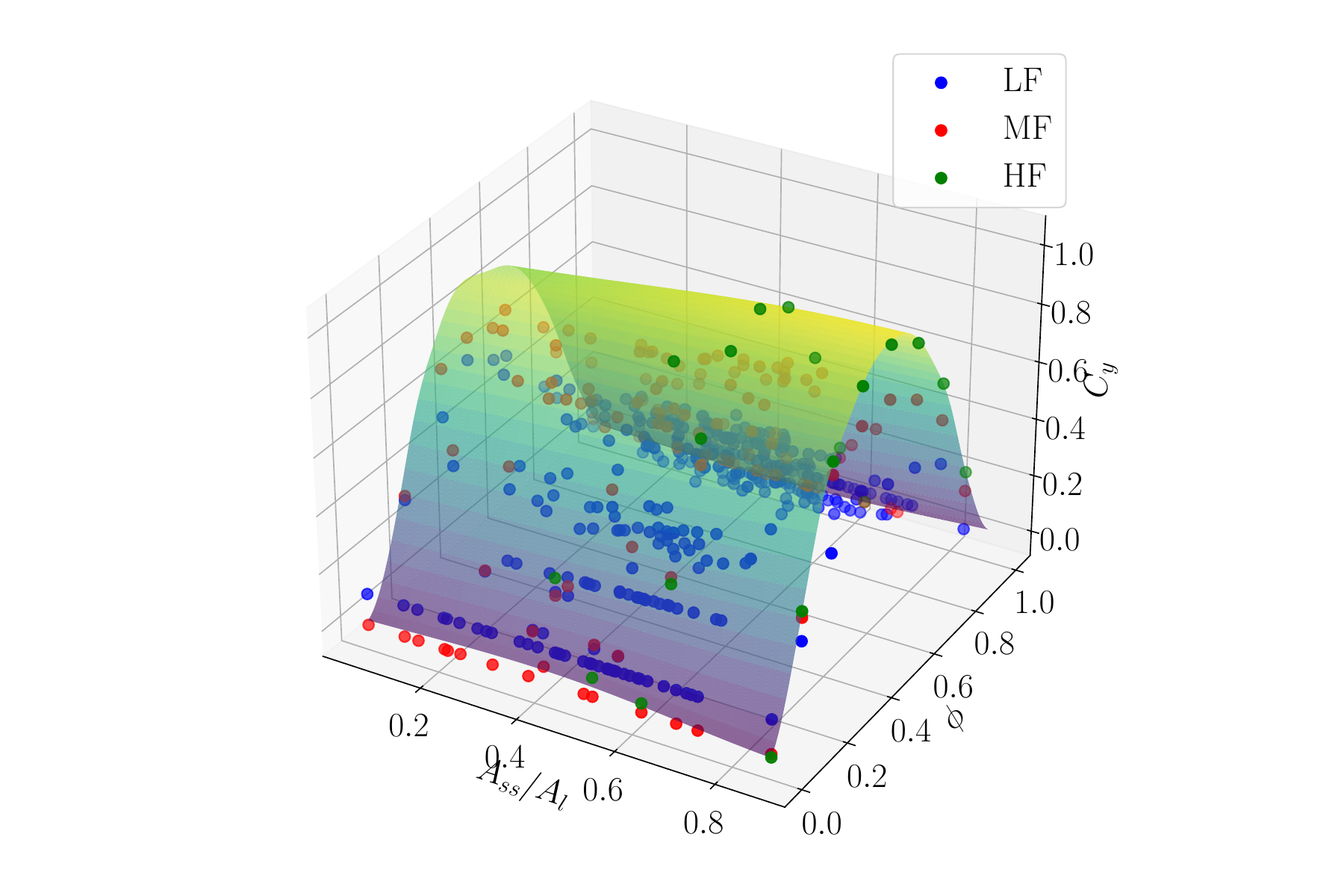}
	\caption{Response surface of the multi-fidelity surrogate model showing the variability of the lateral wind force coefficient with $\phi$ and $A_{ss}/A_l$ (the other, less sensitive parameters are set to their average values).}
	\label{surf_Cy}
\end{figure*}

\begin{figure*}[htbp!]
	\centering
	\includegraphics[scale=0.4]{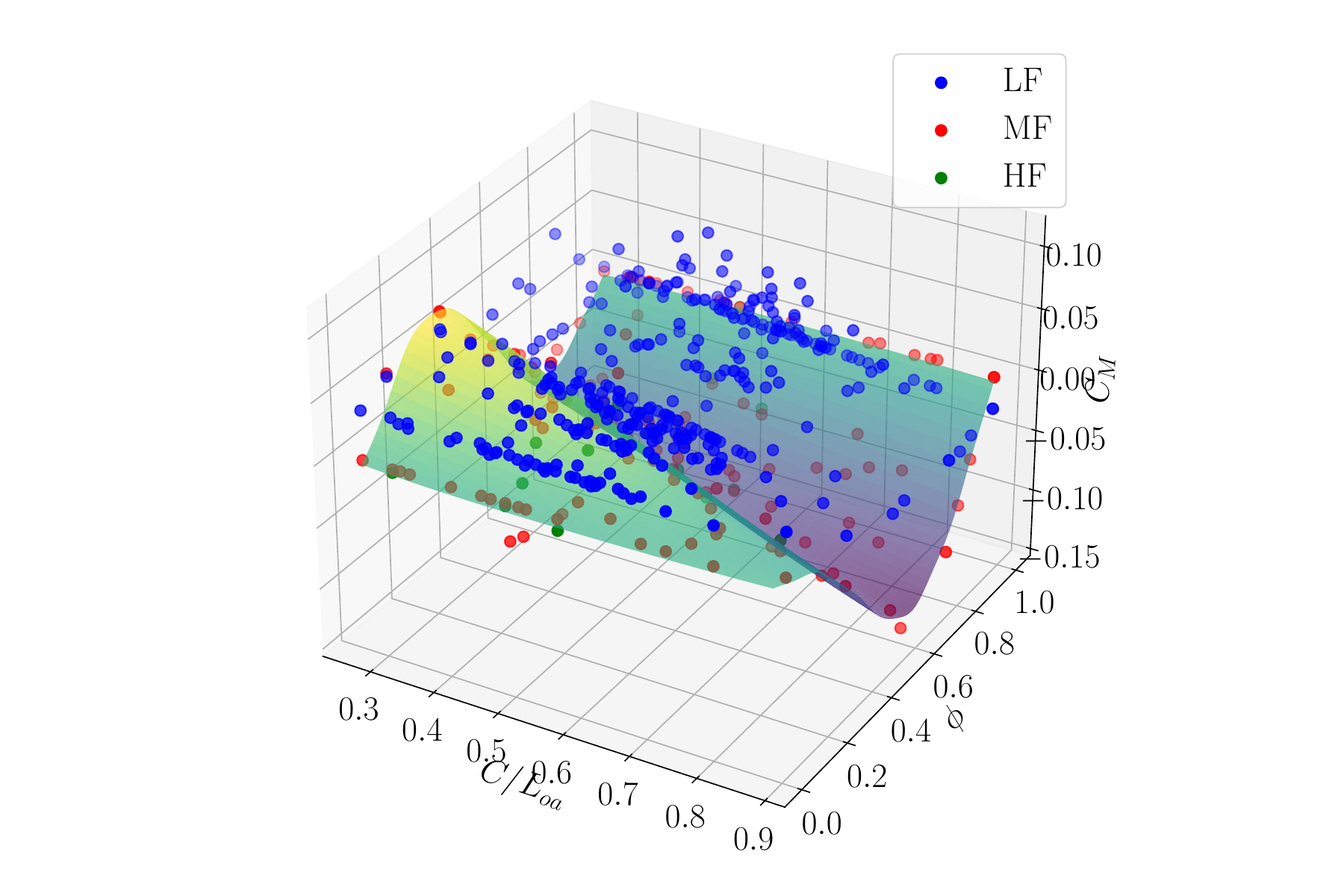}
	\caption{Response surface of the multi-fidelity surrogate model showing the variability of the yaw moment coefficient with $\phi$ and $C/L_{oa}$ (the other, less sensitive parameters are set to their average values).}
	\label{surf_Cm}
\end{figure*}

The influence of the most relevant parameters is further illustrated by the response surfaces shown in Figs.~\ref{surf_Cx}--\ref{surf_Cm}, obtained by varying the dominant parameter while fixing the remaining ones at their mean values. The longitudinal coefficient $C_X$ (Fig.~\ref{surf_Cx}) increases with the number of gaps in the container arrangement, consistently with the enhanced flow penetration through the stack. The lateral coefficient $C_Y$ (Fig.~\ref{surf_Cy}) shows a mild increase with the superstructure area, confirming its relatively weak sensitivity. The yaw moment coefficient $C_M$ (Fig.~\ref{surf_Cm}) exhibits a quasi-sinusoidal dependence on the angle of attack, with the phase and amplitude modulated by the position of the centroid, highlighting the coupling between geometry and flow orientation.

\section{Conclusions and outlook}\label{concl}

This work introduced a multi-fidelity surrogate modelling framework for the prediction of wind loads on container ships, combining heterogeneous data sources within a recursive Gaussian Process formulation. The proposed approach enables an efficient exploitation of low- and medium-fidelity information to significantly reduce the reliance on expensive high-fidelity CFD simulations, while maintaining accurate predictions of aerodynamic loads.

A key element of the methodology is the combination of dimension reduction and adaptive sampling. The active subspace analysis was used to identify the dominant directions in the parameter space, leading to a reduced and physically interpretable set of input variables. This reduced representation was then leveraged within a sequential multi-fidelity training strategy, enabling an efficient exploration of the design space and the construction of a structured database spanning multiple fidelity levels.

The results demonstrate that the multi-fidelity surrogate consistently outperforms single-fidelity models trained on high-fidelity data only, highlighting the benefit of exploiting cross-fidelity correlations. In open-sea conditions, the surrogate achieves satisfactory accuracy, particularly for the longitudinal and lateral force coefficients, with average errors of approximately 15\% for $C_X$, 3\% for $C_Y$, and 20\% for $C_M$. The extension to environment-specific configurations further shows that the proposed framework is able to capture the influence of harbour structures, despite the limited availability of high-fidelity data at these levels.

The sensitivity analysis provides additional insight into the physical drivers of the wind loads, identifying distinct dominant parameters for each coefficient and confirming the higher complexity associated with the prediction of the yaw moment. These findings not only support the interpretation of the surrogate model performance, but also offer guidance for future model refinement and control-oriented applications.

Overall, the proposed framework demonstrates that multi-fidelity surrogate modelling is a viable and effective approach for the prediction of wind loads on large container ships, enabling both improved accuracy and substantial reductions in computational cost. Compared to traditional empirical correlations, the method provides increased flexibility and the capability to account for complex environmental effects.

Future work will focus on extending the methodology to a broader class of ship geometries and more realistic harbour configurations, requiring an enriched parameterization and potentially deeper fidelity hierarchies. In addition, the integration of the surrogate models within operational tools, such as mooring and manoeuvring simulations, will be pursued to assess their performance in real-time and control-oriented applications.

\section*{Acknweledgements} 
\label{sec:acknowledgements}
 This work was made possible by the financial support of VLAIO for the Everblue project (HBC.2022.0663). M.A. Mendez is supported by the European Research Council (ERC) under the European Union’s Horizon Europe research and innovation programme, through a Starting Grant (grant agreement No. 101165479 — RE-TWIST). The views and opinions expressed are those of the authors only and do not necessarily reflect those of the European Union or the European Research Council. The authors acknowledge the Vlaams Supercomputer Centrum (VSC) for the computational resources provided to support this work under project EVERBLUE (2024-093). Finally, the authors would like to thank the Ships and Maritime Technology Division of Ghent University for the ship and environment conceptual design.


\bibliographystyle{elsarticle-num}
\bibliography{references}
\end{document}